\documentclass{article}
\usepackage{microtype}
\usepackage{lipsum} 
\usepackage{graphicx}
\usepackage{caption}
\usepackage{subfigure}
\usepackage{booktabs} 
\usepackage{hyperref}

\usepackage[accepted]{icml2024_conference}

\usepackage{amsmath}
\usepackage{amssymb}
\usepackage{mathtools}
\usepackage{amsthm}
\usepackage{stfloats}

\usepackage{xcolor}

\usepackage[capitalize,noabbrev]{cleveref}

\crefname{subappendix}{\IfAppendix{Sec.}{appendix}}{\IfAppendix{Secs.}{appendices}s}

\protected\def\ignorethis#1\endignorethis{}
\let\endignorethis\relax

\theoremstyle{plain}

\theoremstyle{definition}

\theoremstyle{remark}

\usepackage[textsize=tiny]{todonotes}

\usepackage{amsmath,amsfonts,bm}

\def\Figref#1{Figure~\ref{#1}}

\def\Secref#1{Section~\ref{#1}}

\def\eqref#1{equation~\ref{#1}}

\def\1{\bm{1}}

\def\eps{{\epsilon}}

\DeclareMathAlphabet{\mathsfit}{\encodingdefault}{\sfdefault}{m}{sl}
\SetMathAlphabet{\mathsfit}{bold}{\encodingdefault}{\sfdefault}{bx}{n}

\DeclareMathOperator*{\argmin}{arg\,min}

\usepackage{url}
\usepackage{algorithm}
\usepackage{algorithmic}
\usepackage{float}
\usepackage[export]{adjustbox}
\usepackage{pifont}
\usepackage{enumitem}
\setlist[itemize]{nosep, leftmargin=*}
\newcommand{\cmark}{\color{teal}{\ding{51}}}
\newcommand{\xmark}{\color{purple}{\ding{55}}}

\newcommand{\bepsilon}{{\boldsymbol{\epsilon}}}

\icmltitlerunning{An Image is Worth Multiple Words: Discovering Object Level Concepts using Multi-Concept Prompt Learning}

\begin{document}
\addtocontents{toc}{\protect\setcounter{tocdepth}{0}}

\twocolumn[
\icmltitle{An Image is Worth Multiple Words: Discovering Object Level Concepts using Multi-Concept Prompt Learning}

\begin{icmlauthorlist}
\icmlauthor{Chen Jin}{az}
\icmlauthor{Ryutaro Tanno}{deepmind}
\icmlauthor{Amrutha Saseendran}{az}
\icmlauthor{Tom Diethe}{az}
\icmlauthor{Philip Teare}{az}
\end{icmlauthorlist}

\icmlaffiliation{az}{Centre for AI, DS\&AI, AstraZeneca, UK}
\icmlaffiliation{deepmind}{Google DeepMind, UK}

\icmlcorrespondingauthor{Chen Jin}{chen.jin@astrazeneca.com}
\icmlcorrespondingauthor{Philip Teare}{philip.teare@astrazeneca.com}

\begin{@twocolumnfalse}
  {
    \centering
    \includegraphics[width=\textwidth]{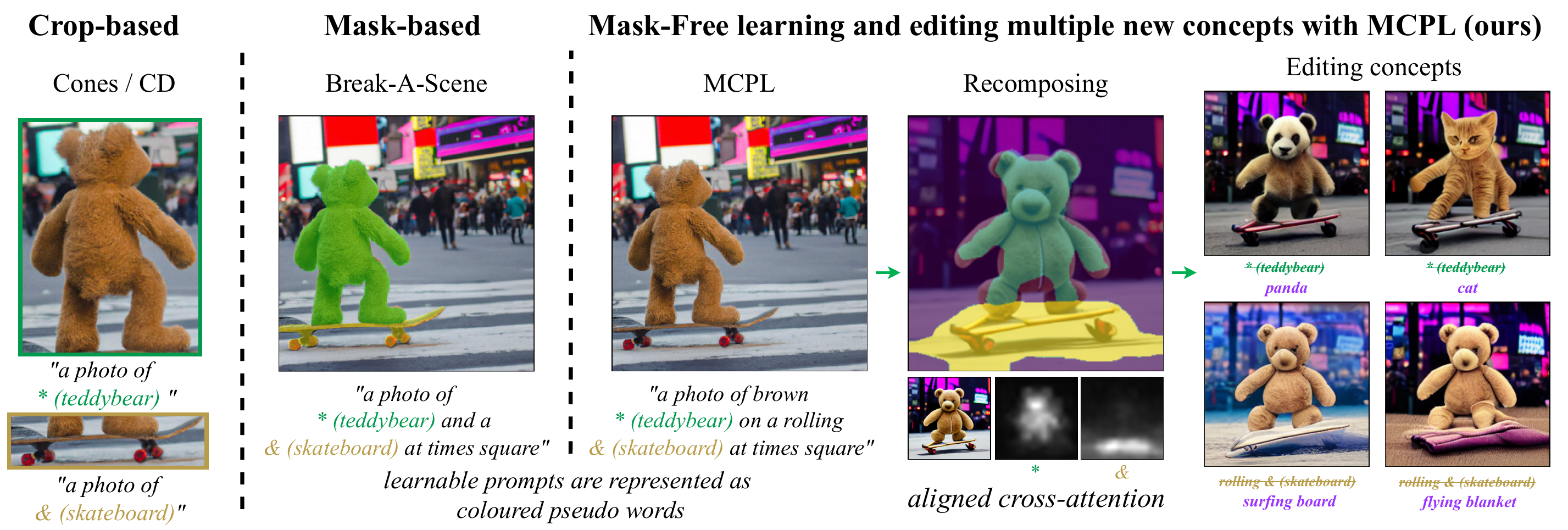}
    \captionof{figure}{\textbf{Language driven multi-concepts learning and applications.} Custom Diffusion (CD) and Cones learn concepts from crops of objects, while Break-A-Scene uses masks. In contrast, our method learns object-level concepts using image-sentence pairs, aligning the cross-attention of each learnable prompt with a semantically meaningful region, and enabling mask-free local editing. The project page, code, and data are available at \href{https://astrazeneca.github.io/mcpl.github.io}{https://astrazeneca.github.io/mcpl.github.io}.
    }
    \label{fig:teaser}
  }
\end{@twocolumnfalse}

\icmlkeywords{Machine Learning, ICML}

\vskip 0.3in
]

\printAffiliationsAndNotice{}  

\begin{abstract}
Textual Inversion, a prompt learning method, learns a singular text embedding for a new ``word'' to represent image style and appearance, allowing it to be integrated into natural language sentences to generate novel synthesised images. 
However, identifying multiple unknown object-level concepts within one scene remains a complex challenge.
While recent methods have resorted to cropping or masking individual images to learn multiple concepts, these techniques require image annotations which can be scarce or unavailable.
To address this challenge, we introduce \textit{Multi-Concept Prompt Learning (MCPL)}, where multiple unknown ``words'' are simultaneously learned from a single sentence-image pair, without any imagery annotations. 
To enhance the accuracy of word-concept correlation and refine attention mask boundaries, we propose three regularisation techniques: 
\textit{Attention Masking}, \textit{Prompts Contrastive Loss}, and \textit{Bind Adjective}.
Extensive quantitative comparisons with both real-world categories and biomedical images demonstrate that our method can learn new semantically disentangled concepts. 
Our approach emphasises learning solely from textual embeddings, using less than 10\% of the storage space compared to others. 

\end{abstract}

\vspace{-6mm}
\section{Introduction}
\label{introduction}

Language-driven vision concept discovery is a human-machine interaction process in which the human describes an image, leaving out multiple unfamiliar concepts. The machine then learns to link each new concept with a corresponding learnable prompt (pseudo words in \Figref{fig:teaser}) from the sentence-image pair.
Such capacity would accelerate the scientific knowledge discovery process either from experimental observations or mining existing textbooks. It also facilitates hypothesis generation through local image editing without concrete knowledge of the new vision concept.

Recent research (\cite{gal2022textual, ruiz2022dreambooth}) shows that the appearance and style of an image can be encapsulated as a cohesive concept via a learned prompt (``word'') optimised in the frozen embedding space of a pre-trained text-to-image diffusion model. 
To bring the learning down to multiple objects in a scene, Custom Diffusion \citep{kumari2023multi} and Cones \citep{liu2023cones}, make use of the crops of objects, while Break-A-Scene \citep{avrahami2023break} uses masks, as shown in \Figref{fig:teaser}. These approaches optimise the integration performance of multiple learned concepts via fine-tuning and storing the diffusion model weights.
Yet, these approaches are less suitable for discovering unknown semantic concepts from historical data at scale when: 
1) annotations are absent (e.g. medical log); 2) concepts are unknown (e.g. discovering new biomarker); 3) minimizing storage per concept is crucial (see Table~\ref{tab:competing_methods}).

\begin{table}[h]
\vspace{-3mm}
    \centering
    \caption{\textbf{Competing methods.} Our method is the first to suggest a solution for discovering new visual concepts using text descriptions (ideally assisted by one adjective per concept). We focus on token learning for minimal storage cost (every 1$\sim$4 token in the table).
    Other methods like BAS perform fine-tuning and storing diffusion model weights for optimal integration performance. 
    }
    \begin{adjustbox}{width=1\columnwidth}
        \begin{tabular}{lccccc}
            \toprule          
            \textbf{Method} & Multi- & Single & Auxiliary & Token & Storage \\
            & concept & image & input & only & cost \\           
            \midrule
            Textual Inversion   & \xmark & \xmark & - & \cmark & $<$0.1MB \\
            Dreambooth          & \xmark & \xmark & - & \xmark & 3.3GB \\
            Custom Diffusion    & \cmark & \xmark &  Crop & \xmark & 72MB \\
            Cones               & \cmark & \cmark &  Crop & \xmark & 1$\sim$10MB \\
            Break-A-Scene       & \cmark & \cmark & Mask & \xmark & 4.9GB \\
            \midrule
            MCPL (Ours)         & \cmark & \cmark & Text & \cmark & $<$0.1MB \\            
            \bottomrule
        \end{tabular}
    \end{adjustbox}
    \label{tab:competing_methods}
\end{table}

In this work, we explore learning object-level concepts using only natural language descriptions and only updating and storing the textual embedding (token).
We start with a motivational study that confirms, without updating DM parameters, while applying masking or cropping yields distinct embeddings, object-level learning and editing relying solely on linguistic descriptions remains challenging. 
Motivated by this finding, we introduce \textit{Multi-Concept Prompt Learning (MCPL)} \Figref{fig:method_overview} (Top) for \textbf{mask-free text-guided learning of multiple prompts from one scene}.

However, without further assumptions on the embedding relationships, jointly learning multiple prompts is problematic. The model may disregard the semantic associations and instead prioritise optimising multiple embedding vectors for optimal image-level reconstruction.  To enhance the accuracy of prompt-object level correlation, we propose the following regularisation techniques: 1) To ensure a concentrated correlation between each prompt-concept pair, we propose \textit{Attention Masking (AttnMask)}, restricting prompt learning to relevant regions defined by a cross-attention-guided mask. 2) Recognising that multiple objects within a scene are semantically distinct, we introduce \textit{Prompts Contrastive Loss (PromptCL)} to facilitate the disentanglement of prompt embeddings associated with multiple concepts. 3) To further enable accurate control of each learned embedding, we bind each learnable prompt with a related descriptive adjective word, referred to as \textit{Bind adj.}, that we empirically observe has a strong regional correlation. The middle and bottom row of \Figref{fig:method_overview} illustrates the proposed regularisation techniques. Our evaluation shows that our framework improves precision in learning object-level concepts and facilitates explainable hypothesis generation via local editing. This is achieved without requiring explicit image annotations, as exemplified in \Figref{fig:teaser} (ours).

In this work we implement our proposed method based on Textual Inversion by \cite{gal2022textual}, which only learns and stores textual embeddings, and is \textit{complementary to the more expensive generation-focused approaches that update DM parameters} (see Table~\ref{tab:competing_methods}).
To our knowledge, our technique is the first to learn multiple object-level concepts without using a crop or mask. 
To evaluate this new task, 
we generate and collect in-distribution natural images and out-of-distribution biomedical images, each featuring 2 to 5 concepts along with object-level masks. This results in a dataset comprising 25 concepts and 1,000 sentence-image pairs. 
We run around 3500 GPU hours experiments to compare with competitive methods, with each run taking approximately one hour.
We assess concept disentanglement using t-SNE and evaluate object embedding similarity against masked ground truth in pre-trained BERT, CLIP, DINOv1, and DINOv2 embedding spaces. Our results show that our method can identify multiple concepts in an image and supports the discovery of new concepts using only text descriptions.

\vspace{-2mm}
\section{Related Works}
\label{related_works}

\paragraph{Language-driven vision concept discovery.} 
In many scientific fields, discovery often begins with visual observation and then progresses by exploring the existing knowledge base to pinpoint unfamiliar object-level concepts. These concepts are subsequently defined using new terms, facilitating the development of hypotheses \citep{sep-scientific-discovery}. 
The emergence of artificial intelligence, particularly large pre-trained Vision-Language Models (VLM), has laid the groundwork for automating this discovery process \citep{wang2023scientific}. Language-driven local editing in VLMs shows promise for helping scientists to generate hypotheses and create designs \cite{hertz2022prompt, tumanyan2023plug, patashnik2023localizing}. 
However, a key challenge remains: relying solely on linguistic descriptions, current methods may not always map words to their corresponding object-level concepts precisely.

\vspace{-3mm}
\paragraph{Prompt learning for Diffusion Model.}
In text-guided image synthesis, prompt learning links the appearance and style of an unseen image to a learnable prompt, enabling transfer to new images. This is achieved either by learning and storing textual embeddings, as in Textual Inversion \cite{gal2022textual}, or by optimising the entire diffusion model to reconstruct a given example image, as demonstrated in DreamBooth \cite{ruiz2022dreambooth}.

\paragraph{Multiple concept learning and composing.} 
Recent advancements focus on efficiently composing multiple concepts learned separately from single object images or crops (Custom Diffusion \citep{kumari2023multi}, Cones \citep{liu2023cones}, SVDiff \citep{han2023svdiff}, Perfusion \citep{tewel2023key}. 
ELITE \citep{wei2023elite} and Break-A-Scene \citep{avrahami2023break} adopt masks for improved object-level concept learning, with Break-A-Scene specifically aiming for multi-concept learning and integrating within single images, aligning closely with our objectives.
Our approach differs from Break-A-Scene in two key aspects: 1) we aim to \textit{eliminate the need for labour-intensive image annotations}, and 2) we explore the limits of multi-concept learning \textit{without updating or storing the DM parameters}. 
Inspiration Tree (IT) \citep{vinker2023concept} shares a similar spirit as our work. MCPL excels in instruction following discovery, adept at identifying concepts distinct in both visual and linguistic aspects, whereas IT specializes in unveiling abstract and subtle concepts independently of direct human guidance.

\section{Methods}
\label{methods}

In this section, we outline the preliminaries in \Secref{sec: preliminaries} and present a motivational study in \Secref{sec: empirical}. 
These tests validate the presence of object-level embeddings in the pretrained textual embedding space, highlighting the challenges in learning multiple concepts without image annotations. Inspired by these results, we introduce the \textit{Multi-Concept Prompt Learning (MCPL)} in \Secref{sec: mcpl}. To address the multi-object optimisation challenge in tandem with a single image-level reconstruction goal, we propose several regularisation techniques in \Secref{sec: regularisation}.

\subsection{Preliminaries}
\label{sec: preliminaries}

\textbf{Text-guided diffusion models} are probabilistic generative models that approximate the data distribution (specifically, images in our work) by progressively denoising Gaussian random noise, conditioned on text embeddings. 
Specifically, we are interested in a denoising network $\eps_\theta$ being pre-trained such that, given an initial Gaussian random noise map $\bepsilon \sim \mathcal{N}(\mathbf{0}, \textbf{I})$, conditioned to text embeddings $v$, generates an image $\tilde{x}=\eps_\theta(\bepsilon, v)$ closely resembling a given example image $x$. 
Here, $v = c_\phi(p)$, where $c_\phi$ is a pre-trained text encoder with parameters $\phi$ and $p$ is the text. During training, $\phi$ and $\theta$ are jointly optimised to denoise a noised image embedding $z_t$ to minimise the loss:
%
\begin{equation}
\label{eq:DM_loss}
    L_{DM} = L_{DM}(x,\tilde{x}) := E_{z,v,\eps,t} \Vert \eps-\eps_\theta(z_t,v) \Vert^{2}.
\end{equation}
Here, $z_t \coloneqq \alpha_t z + \sigma_t \bepsilon$ is the noised version of the initial image embedding $z$ at time $t\sim \text{Uniform}(1,T)$, $\alpha_t, \sigma_t$ are noise scheduler terms, and $z = \mathcal{E}(x)$, where $\mathcal{E}$ is the encoder of a pretrained autoencoder $D\left(\mathcal{E}(x)\right)\approx x$, following Latent Diffusion Models (LDMs) \citep{rombach2022high} for computational efficiency.
During inference, the pre-trained model iteratively eliminates noise from a new random noise map to generate a new image.

\textbf{Textual Inversion} \cite{gal2022textual} is aimed at identifying the text embedding $v^*$ for a new prompt $p^*$ with pre-trained $\{\eps_\theta, c_\phi\}$. Given a few (3-5) example images representing a specific subject or concept, the method optimises $v^*$ in the frozen latent space of text encoder $c_\phi$. The objective is to generate images via the denoising network $\eps_\theta$ that closely resembles the example images (after decoding) when conditioned on $v^*$. The optimisation is guided by the diffusion model loss defined in \eqref{eq:DM_loss}, updating only $v^*$ while keeping $c_\phi$ and $\epsilon_\theta$ frozen.
During training, the generation is conditioned on prompts combining randomly selected text templates $y$ (e.g., ``A photo of", ``A sketch of" from CLIP \citep{radford2021learning}) with the new prompt $p^*$, resulting in phrases like ``A photo of $p^*$" and ``A sketch of $p^*$".

\textbf{Cross-attention layers} play a pivotal role in directing the text-guided diffusion process. Within the denoising network, $\eps_\theta$, at each time step $t$ the textual embedding, $v = c_\phi(p)$, interacts with the image embedding, $z_{t}$, via the cross-attention layer. Here, $Q=f_Q(z_{t})$, $K=f_K(v)$, and $V=f_V(v)$ are acquired using learned linear layers $f_Q, f_K, f_V$. As \cite{hertz2022prompt} highlighted, the per-prompt cross-attention maps, $M=\text{Softmax}(QK^T/\sqrt{d})$, correlate to the similarity between $Q$ and $K$. Therefore the average of the cross-attention maps over all time steps reflects the crucial regions corresponding to each prompt word, as depicted in \Figref{fig:method_overview}. 
In this study, the per-prompt attention map is a key metric for evaluating the prompt-concept correlation. Our results will show that without adequate constraints, the attention maps for newly learned prompts often lack consistent disentanglement and precise prompt-concept correlation.

\subsection{Motivational study}
\label{sec: empirical}

\begin{figure*}[h]
    \centering
    \includegraphics[width=\linewidth]{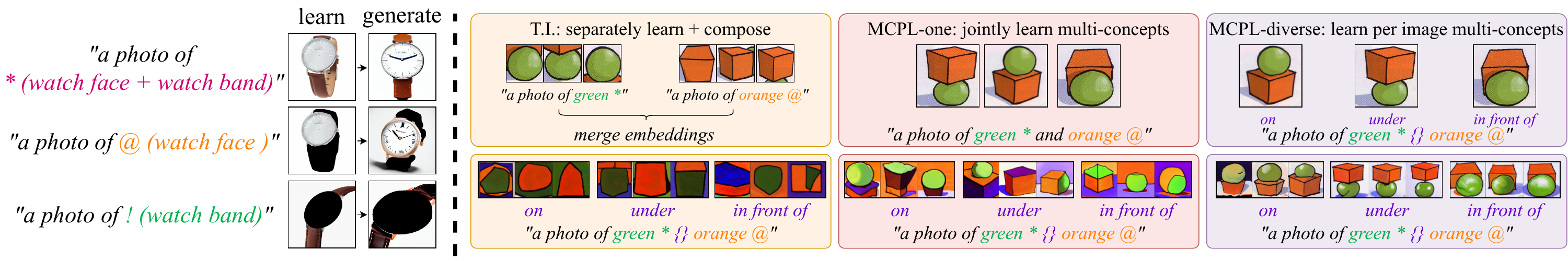}
    \vspace{-4mm}
    \caption{
    \textbf{Motivational study and preliminary MCPL results.} We use \textit{Textual Inversion (T.I.)} to learn concepts from both masked (left-first) or cropped (left-second) images; \textit{MCPL-one}, learning both concepts jointly from the full image with a single string; and \textit{MCPL-diverse} accounting for per-image specific relationships. 
    }
    \label{fig: motivation_combined}
\end{figure*}

To understand the possibility of learning multiple concepts within a frozen textual embedding space, we explored whether \textit{Textual Inversion} can discern semantically distinct concepts from both masked and cropped images, each highlighting a single concept. \Figref{fig: motivation_combined} (two examples on the left) gives a highlight of our result, with a full version in Appendix \ref{sec:full_motivation}, confirms that: 1) multiple unique embeddings can be derived from a single multi-concept image, albeit with human intervention, and 2) despite having well-learned individual concepts, synthesising them into a unified multi-concept scene remains challenging. To address these issues, we introduce the Multi-Concept Prompt Learning (MCPL) framework. MCPL modifies Textual Inversion to enable simultaneous learning of multiple prompts within the same string. 

\subsection{Multi-Concept Prompt Learning (MCPL)}
\label{sec: mcpl}
\begin{figure*}[ht]
    \centering
    \includegraphics[width=\linewidth]{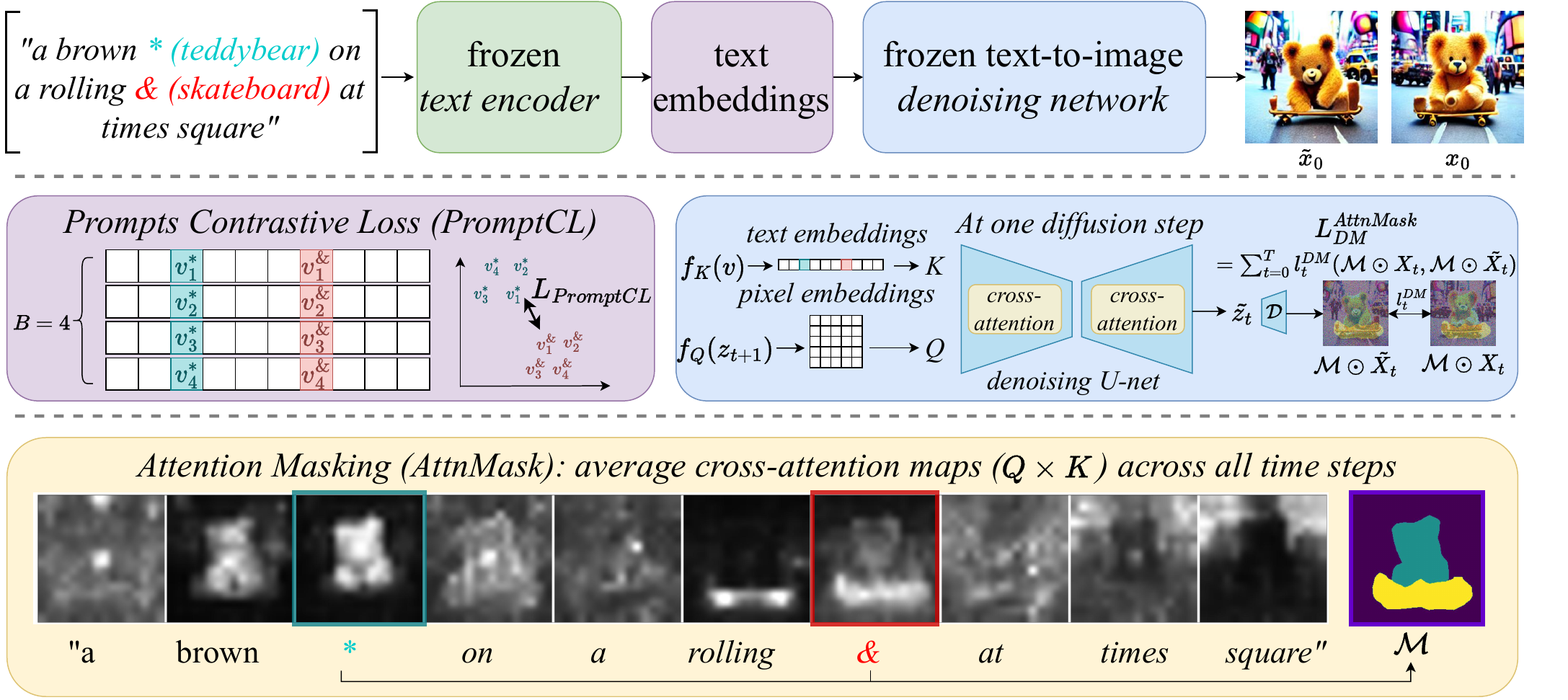}
    \vspace{-3mm}
    \caption{
    \footnotesize
    \textbf{Method overview.} 
    \textit{MCPL} takes a sentence (top-left) and a sample image $x_0$ (top-right) as input, feeding them into a pre-trained text-guided diffusion model comprising a text encoder $c_\phi$ and a denoising network $\epsilon_\theta$. 
    The string's multiple prompts are encoded into a sequence of embeddings which guide the network to generate images $\tilde{x}_0$ close to the target one $x_0$. MCPL focuses on learning multiple learnable prompts (coloured texts), updating only the embeddings $v^*$ and $v^\&$ of the learnable prompts while keeping $c_\phi$ and $\epsilon_\theta$ frozen.
    We introduce \textit{Prompts Contrastive Loss (PromptCL)} to help separate multiple concepts within learnable embeddings. We also apply \textit{Attention Masking (AttnMask)}, using masks based on the average cross-attention of prompts, to refine prompt learning on images. Optionally we associate each learnable prompt with an adjective (e.g., ``brown") to improve control over each learned concept, referred to as \textit{Bind adj.}
    }
    \label{fig:method_overview}
\end{figure*}

For example image(s) $x$, MCPL learn a list of embeddings $\mathcal{V}=[v^*,\ldots,v^\&]$ corresponds to multiple new prompts $\mathcal{P}=[p^*,\ldots,{p}^\&]$ within the descriptive sentence as shown in \Figref{fig:method_overview}. 
The optimisation is guided by the image-level $L_{DM}$ (\eqref{eq:DM_loss}), but now updating $\mathcal{V}$ while keeping $c_\phi$ and $\epsilon_\theta$ frozen. 
The MCPL algorithm is outlined in Algorithm \ref{alg:alg_mcpl}.
Here each sentence $\mathcal{P}$ can be sub-grouped into
    a set of noun words of target concepts $[n^*,\ldots,{n}^\&]$, 
    a set of adjective words $[a^*,\ldots,{a}^\&]$ describing each noun
    and the rest preposition texts $[t^*,\ldots,{t}^\&]$ in the sentence (exclude random neutral texts e.g. ``a photo of''). 
Each sub-group can be designated as either learnable or fixed, depending on the training strategies that we will define in the subsequent section.
For evaluation, we request a human or a machine, such as GPT-4, to describe each image using one adjective and one noun for each target concept.

\begin{algorithm}[H]
\footnotesize
\caption{MCPL (generic form)}
\label{alg:alg_mcpl}
\begin{algorithmic}[1]
\STATE \textbf{Input:} example image(s) $x$, pre-trained $\{c_\theta, \eps_\theta\}$.
\STATE \textbf{Output:} a list of embeddings $\mathcal{V}=[v^*,\ldots,v^\&]$ corresponds to multiple new prompts $\mathcal{P}=[p^*,\ldots,{p}^\&]$.
\STATE initialise $[v^*,\ldots,v^\&] = [c_\theta(p^*),\ldots,c_\theta(p^\&)]$
\STATE \textcolor{blue}{\texttt{\# optimising $\{v^*,\ldots,v^\&\}$ with $L_{DM}$}}
\FOR{$step = 1$ to $S$}
    \scriptsize
    \STATE \textbf{Encode} example image(s) $z = \mathcal{E}(x)$ and randomly sample neutral texts $y$ to make string $[y, p^*,\ldots,p^\&]$
    \STATE \makebox[0pt][l]{\textbf{Compute} $\mathcal{V_y}=[v^y,v^*,\ldots,v^\&] = [c_\theta(p^y),c_\theta(p^*),\ldots,c_\theta(p^\&)]$}
    \FOR{$t = T$ down to $1$}
        \STATE $\mathcal{V} := \argmin_{\mathcal{V}} E_{z,\mathcal{V},\eps,t} \Vert \eps - \eps_\theta(z_t, \mathcal{V_y}) \Vert^{2}$
    \ENDFOR
\ENDFOR
\STATE \textbf{Return} $(\mathcal{P}, \mathcal{V})$
\end{algorithmic}
\end{algorithm}

%
\paragraph{Training strategies and preliminary results.} 
Recognising the complexity of learning multiple embeddings with a single image-generation goal, we propose three training strategies: 1) \textit{MCPL-all}, a naive approach that learns embeddings for all prompts in the string (including adjectives, prepositions and nouns. etc.); 2) \textit{MCPL-one}, which simplifies the objective by learning single prompt (nouns) per concept; 3) \textit{MCPL-diverse}, where different strings are learned per image to observe variances among examples.
The formal definitions of each MCPL training strategy are given in Appendix \ref{sec: algorithm}.
Preliminary evaluations of \textit{MCPL-one} and \textit{MCPL-diverse} methods on the ``ball" and ``box" multi-concept task are shown in \Figref{fig: motivation_combined}. Our findings indicate that \textit{MCPL-one} enhance the joint learning of multiple concepts within the same scene over separate learning. Meanwhile, \textit{MCPL-diverse} goes further by facilitating the learning of intricate relationships between multiple concepts.

\begin{figure}[H]
  \centering
  \includegraphics[width=\linewidth]{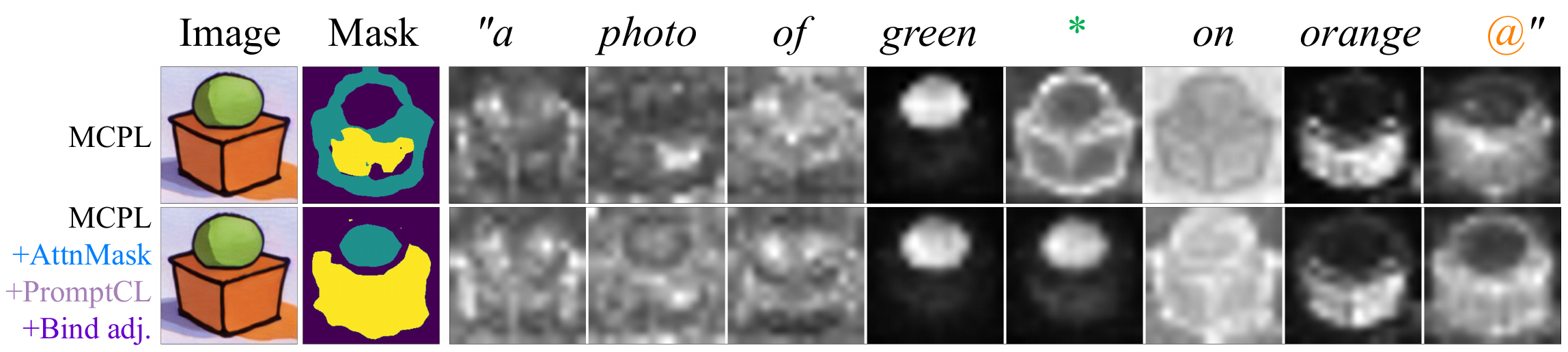}
  \caption{\textbf{Enhancing object-level prompt-concept correlation in MCPL} using the proposed regularisations: \textit{AttnMask}, \textit{PromptCL} and \textit{Bind adj.}. 
  We compare \textit{MCPL-one} applying all regularisation terms against the \textit{MCPL-one}, using a ``\textcolor{teal}{Ball} and \textcolor{orange}{Box}" example. We use the average cross-attention maps and the \textit{AttnMask} to assess the accuracy of correlation. Full ablation results in Appendix \ref{sec: full_ablation}}
    \label{fig:improve_MCPL_sim}
\end{figure}

\paragraph{Limitations of plain MCPL.} We aim to discover new visual concepts using only linguistic descriptions and then enable accurate local editing. It requires accurate object-level prompt-concept correlation, to evaluate, we visualise the average cross-attention maps for each prompt. As depicted in \Figref{fig:improve_MCPL_sim} (top), plain \textit{MCPL} inadequately capture this correlation, especially for the target concept. These results suggest that \textit{naively extending image-level prompt learning techniques \citep{gal2022textual} to object-level multi-concept learning poses optimisation challenges}, notwithstanding the problem reformulation efforts discussed in \Secref{sec: mcpl}. Specifically, optimising multiple object-level prompts based on a single image-level objective proves to be non-trivial. Given the image generation loss \eqref{eq:DM_loss}, prompt embeddings may converge to trivial solutions that prioritize image-level reconstruction at the expense of semantic prompt-object correlations, thereby contradicting our objectives. In the next section, we introduce multiple regularisation terms to overcome this challenge.

\subsection{Regularising the multi-concept prompts learning}
\label{sec: regularisation}

\paragraph{Encouraging focused prompt-concept correlation with Attention Masking (\textit{AttnMask}).}
Previous results show plain \textit{MCPL} may learn prompts focused on irrelevant areas. To correct this, we apply masks to both generated and target images over all the denoising steps (\Figref{fig:method_overview}, middle-right). These masks, derived from the average cross-attention of selected learnable prompts (\Figref{fig:method_overview}, bottom-row), constrain the image generation loss (\eqref{eq:DM_loss}) to focus on pertinent areas, thereby improving prompt-concept correlation.
To calculate the mask, we compute for each selected learnable prompt $p \in \mathcal{P}$ the average attention map over all time steps $\overline{M}^p = 1/T \sum_{t=1}^{T} M_t^p$. We then apply a threshold to produce binary maps for each learnable prompt, where $B(M^p) := \{1 \text{ if } M^p > k, 0 \text{ otherwise} \}$ and $k = 0.5$ throughout all our experiments. For multiple prompt learning objectives, the final mask $\mathcal{M}$ is a union of multiple binary masks of all learnable prompts $\mathcal{M} = \bigcup_{p \in \mathcal{P}} B(M^p)$. We compute the Hadamard product of $\mathcal{M}$ with $x$ and $\tilde{x}$ to derive our masked loss $L^{\textit{AttnMask}}_{DM}$ as \eqref{eq:DM_loss_masked}.
Our \textit{AttnMask} is inspired by \cite{hertz2022prompt}, but a reverse of the same idea, where the \textit{AttnMask} is applied over the pixel-level loss \eqref{eq:DM_loss} to constrain the prompt learning to only related regions.
\begin{equation}
    L^{\textit{AttnMask}}_{DM} = L_{DM} \left(\mathcal{M} \odot x, \mathcal{M} \odot \tilde{x} \right),
    \label{eq:DM_loss_masked}
\end{equation}
%
\paragraph{Encouraging semantically disentangled multi-concepts with Prompts Contrastive Loss (\textit{PromptCL}).}
\textit{AttnMask} focuses the learning of multiple prompts on the joint area of target objects, eliminating the influence of irrelevant regions like the background. However, it doesn't inherently promote separation between the embeddings of different target concepts. Leveraging the mutual exclusivity of multiple objects in an image, we introduce a contrastive loss in the latent space where embeddings are optimised. Specifically, we employ an InfoNCE loss \cite{oord2018representation}, a standard in contrastive and representation learning, to encourage disentanglement between groups of embeddings corresponding to distinct learnable concepts (\Figref{fig:method_overview}, middle-left). 

Concretely, at each learning step as described in Algorithm \ref{alg:alg_mcpl}, a mini-batch $B$ minor augmented (e.g. with random flip) example images are sampled, with $N$ learnable prompts for each image, yields a set of $BN$ embeddings, $\{v_b^n\}_{b=1}^{B},_{n=1}^N$. Then, the similarity between every pair $v_i$ and $v_j$ of the $BN$ samples is computed using cosine similarity: 
\begin{equation}\label{eq:cos_sim}
        sim(v_i,v_j) = v_i^{T}.v_j/||v_i||||v_j||.
\end{equation}
Given our goal is to differentiate the embeddings corresponding to each prompt, we consider the embeddings of the same concept as positive samples while the others as negative. 
Next, the contrastive loss $l^{\eta}_{i,j \in B}$ for a positive pair $v^{\eta}_{i}$ and $v^{\eta}_{j}$ of each concept $\eta \in N$ (two augmented views of the example image) is shown in the \eqref{eq:prompt_cl_single}, where $\tau$ is a temperature parameter following \cite{chen2020simple}.
The contrastive loss is computed for $BN$ views of each of the $N$ learnable concepts. The total contrastive loss $L_{PromptCL}$ is shown in \eqref{eq:prompt_cl_total}. 
\begin{equation}\label{eq:prompt_cl_single}
l^{\eta}_{i,j \in B} = -\log\left (\frac{\exp(sim(v^{\eta}_{i},v^{\eta}_{j}))/\tau} {\sum_{\eta=1}^{N} \sum_{j=1, j \neq {i}}^{B} \exp(sim(v^{\eta}_{i},v^{\eta}_{j})/\tau ) }\right)
\end{equation}
\begin{equation}\label{eq:prompt_cl_total}
        L_{PromptCL}= \frac{1}{N}\frac{1}{B}\sum_{\eta=1}^{N}\sum_{i=1}^{B} { l^{\eta}_{i,j \in B}}
\end{equation}

\paragraph{Enhance prompt-concept correlation by binding learnable prompt with the adjective word (\textit{Bind adj.}).} 
An additional observation from the misaligned results in \Figref{fig:improve_MCPL_sim} (top) reveals that adjective words often correlate strongly with specific regions. This suggests that the pre-trained model is already adept at recognising descriptive concepts like colour or the term "fluffy" (see full results in \Figref{fig:improve_MCPL}).  
To leverage this innate understanding, we propose to optionally associate one adjective word for each learnable prompt as one positive group during the contrastive loss calculation. In particular, consider $M$ adjective words associated with $N$ learnable prompts. Then the positive pair $v^{\eta}_{i}$ and $v^{\eta}_{j}$ of each concept is sampled from $\eta \in MB$ instead of $B$. Therefore the contrastive loss is now computed for $BNM$ views of each of the $N$ learnable concepts. The resulting total contrastive loss $L_{PromptCL}^{adj}$ is detailed in \eqref{eq:prompt_cl_total_adj}. We scale $L_{PromptCL}^{adj}$ with a scaling term $\gamma$ and add with $L^{\textit{AttnMask}}_{DM}$ (\eqref{eq:DM_loss_masked}), for them to have comparable magnitudes, resulting our final loss in \eqref{eq:final_loss}.

\begin{equation}\label{eq:prompt_cl_total_adj}
        L_{PromptCL}^{adj}= \frac{1}{N}\frac{1}{MB}\sum_{\eta=1}^{N}\sum_{i=1}^{MB} { l^{\eta}_{i,j \in B}}
\end{equation}
\begin{equation}
    L = L^{\textit{AttnMask}}_{DM} + \gamma L_{PromptCL}^{adj},
    \label{eq:final_loss}
\end{equation}

\paragraph{Assessing regularisation terms with cross-attention.}
\label{sec: qualitative}
We assess our proposed regularisation terms on improving the accuracy of semantic correlations between prompts and concepts. 
We visualise the cross-attention and segmentation masks, as shown in \Figref{fig:improve_MCPL_sim}.
Our visual results suggest that incorporating all of the proposed regularisation terms enhances concept disentanglement, whereas applying them in isolation yields suboptimal outcomes (refer to full ablation results in Appendix \ref{sec: full_ablation}). Moreover, the results demonstrate that \textit{MCPL-one} is a more effective learning strategy than \textit{MCPL-all}, highlighting the importance of excluding irrelevant prompts to maintain a focused learning objective.

\section{Experiments}
\label{sec: experiments}
\subsection{Experiment and Implementation Details}

\paragraph{Multi-concept dataset.}
We generate in-distribution natural images and collect out-of-distribution biomedical images, each featuring 2 to 5 concepts along with object-level masks. This results in a dataset comprising 25 concepts and 1,000 sentence-image pairs.
For natural images, we generate multi-concept images using prior local editing \citep{patashnik2023localizing} and multi-concept composing method \citep{avrahami2023break}, both support generating or predicting object masks. 
We generate each image using simple prompts, comprising one adjective and one noun for every relevant concept. For biomedical images, we request a human or a machine, such as GPT-4, to similarly describe each image using one adjective and one noun for each pertinent concept.
For more details, a full list of prompts used and examples, please read Appendix \ref{sec:dataset}.
We release the dataset \href{https://github.com/AstraZeneca/MCPL/tree/master/dataset}{here}.

\paragraph{Competing Methods.}
We compare three baseline methods:
1) \textit{Textual Inversion (TI-m)} applied to each masked object serving as our best estimate for the unknown disentangled ``ground truth" object-embedding.
2) \textit{Break-A-Scene (BAS)}, the state-of-the-art (SoTA) mask-based multi-concept learning method, serves as a performance upper bound, though it's not directly comparable.  
3) \textit{MCPL-all} as our naive adaptation of the \textit{Textual Inversion} method to achieve the multi-concepts learning goal.
For our method, we compare two training strategies of our method: MCPL-all and MCPL-one. For each, we examine three variations to scrutinise the impact of the regularisation terms discussed in Section \ref{sec: regularisation}. All MCPL learnings are performed on unmasked images without updating DM parameters.
To evaluate the robustness and concept disentanglement capability of each learning method, we repeatedly learn each multi-concept pair around 10 times, randomly sampling four images each time. 
We generated a total of 560 masked objects for each MCPL variant and 320 for the BAS baseline.

However, it's important to note that preparing \textit{the BAS as the object-level embedding upper bound is costly}, as it requires an additional pre-trained segmentation model, and occasionally human-in-the-loop, to obtain masks during both the learning and evaluation phases (see Appendix \ref{sec: bas_exp_setup} for details). In contrast, our method utilises its own \textit{AttnMask} to \textit{generate masked concepts during image generation with no extra cost}.

\paragraph{Implementation details.} 
We use the same prompts collected during the data preparation, substituting nouns as learnable prompts, which are merged with CLIP prompts from \Secref{sec: preliminaries}. This process creates phrases such as \textit{``A photo of brown * on a rolling @ at times square"}.
Please find full implementation details in Appendix \ref{sec: implementation_details}.

\subsection{Quantitative Evaluations}
\label{sec: quantitative}

\begin{figure}[h]
    \centering
    \includegraphics[width=\linewidth]{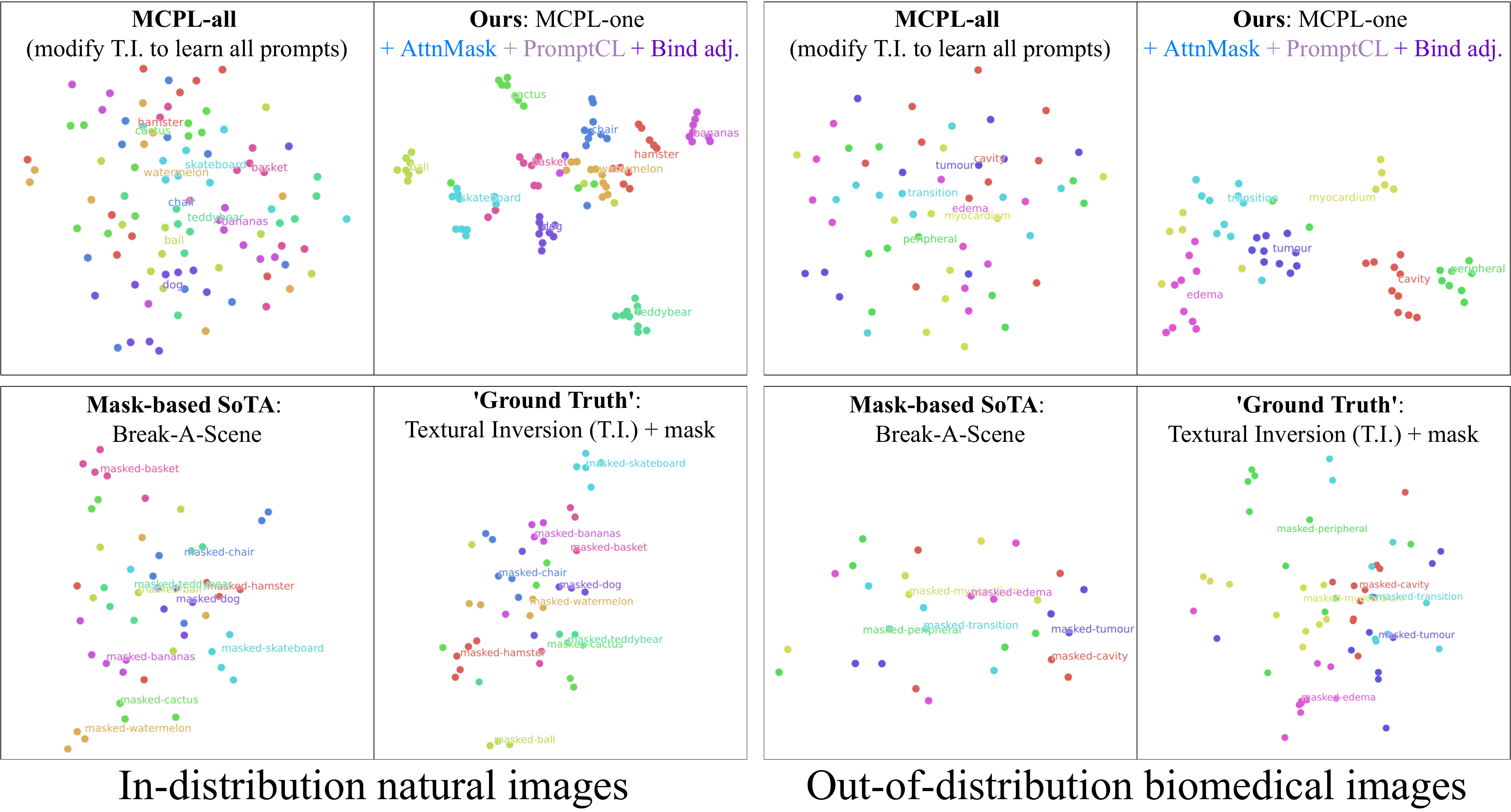}
    \vspace{-6mm}
    \caption{\textbf{The t-SNE projection of the learned embeddings}. Our method can effectively distinguish all learned concepts (about 10 embeddings each concept) compared to Textual Inversion (MCPL-all), the SoTA mask-based learning method, Break-A-Scene, and the masked ``ground truth" (see full results in Appendix \ref{sec: full_tsne}).}
    \label{fig:t_sne}
\end{figure}

\paragraph{Investigate the concepts disentanglement with t-SNE.} 
We aim to learn semantically distinct classes hence our experimental design, therefore we anticipate a clustering effect where objects of the same class naturally cluster together.
To assess clustering, we begin by calculating and visualising the t-SNE projection of the learned features \cite{van2008visualizing}. The results, depicted in \Figref{fig:t_sne}, encompass both natural and biomedical datasets. They illustrate that our \textit{MCPL-one} combined with all regularisation terms can effectively distinguish all learned concepts compared to all baselines.
The learned embeddings from both the mask-based ``ground truth" and BAS are less distinct than ours, due to their absence of a disentanglement goal like MCPL's \textit{PromptCL} loss.

\paragraph{Embedding similarity relative to the ``ground truth".}
To assess the preservation of per-concept semantic and textual details, we calculate both prompt and image fidelity. This evaluation follows prior research by \cite{gal2022textual} and  \cite{ruiz2022dreambooth}, but differently, we perform the calculations at the object level. 
Prompt fidelity is determined by measuring the average pairwise cosine similarity between the embeddings learned from the estimated ``ground truth" and the generated masked images, in the pre-trained embedding space of BERT \citep{devlin2018bert}.
Image fidelity refers to the average pairwise cosine similarity between masked "ground truth" images and generated masked objects within four pre-trained embedding spaces of CLIP \cite{radford2021learning}, DINOv1 \citep{caron2021emerging} and DINOv2 \citep{oquab2023dinov2}, all based on the ViT-S.

\begin{figure}[htb]
  \centering
  \includegraphics[width=1\linewidth]{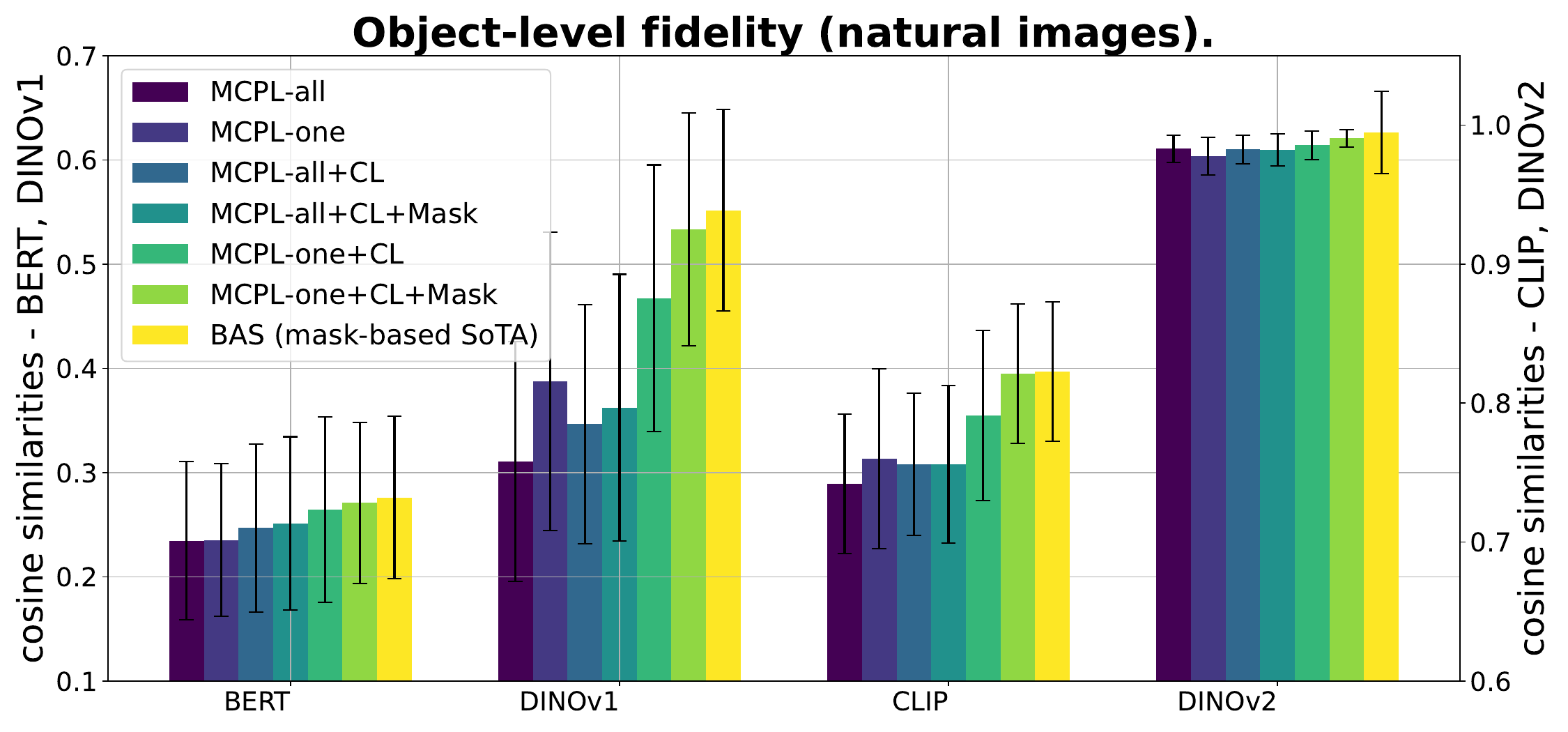}
  \\
  \includegraphics[width=1\linewidth]{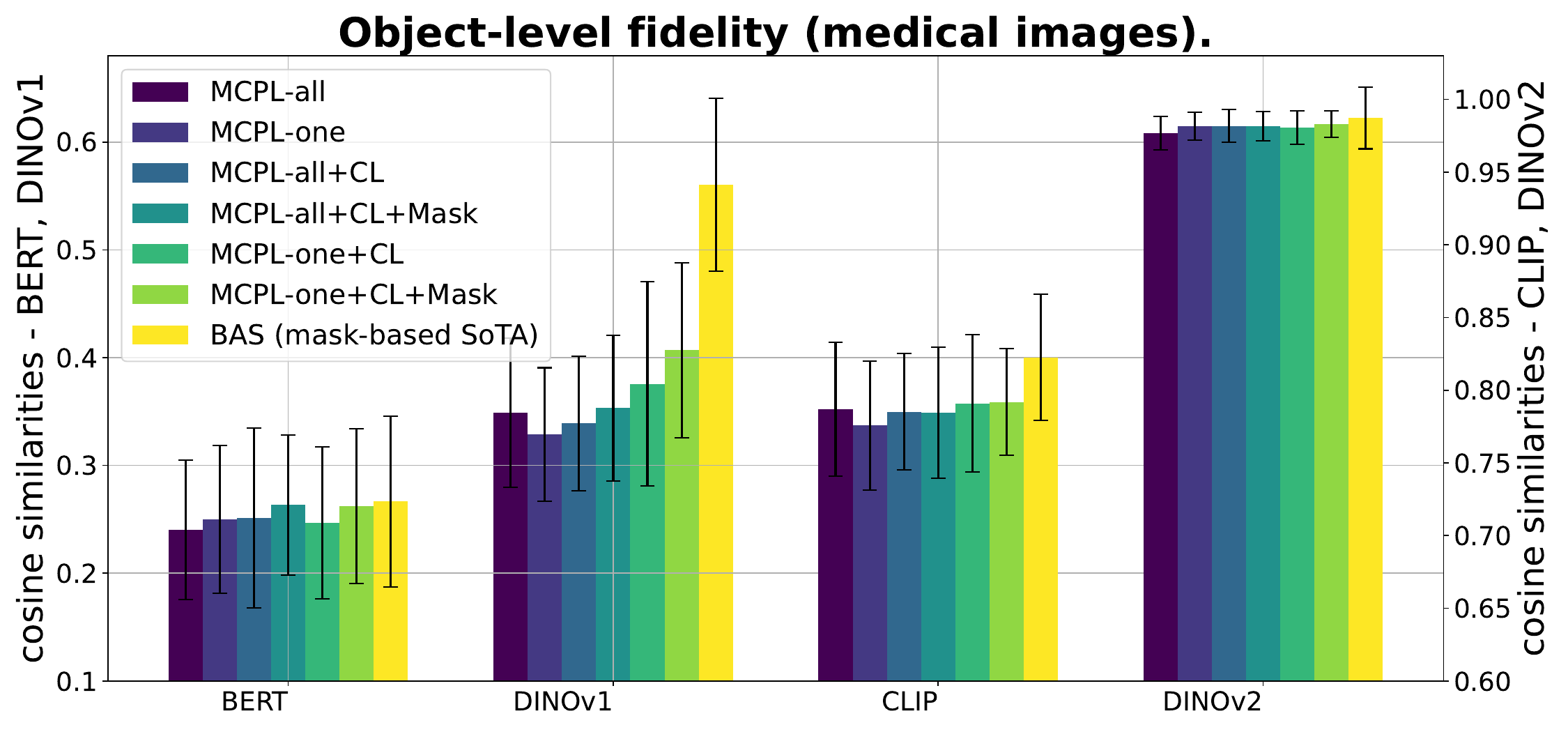}
  \vspace{-6mm}
  \caption{\footnotesize{\textbf{Embedding similarity in learned object-level concepts compared to masked ``ground truth" (two concepts per image).} We compare Textual Inversion (MCPL-all) and the SoTA mask-based learning method, BAS, against our regularised versions. The analysis is conducted in both pre-trained text (BERT) and image encoder spaces (CLIP, DINOv1, and DINOv2), with each bar representing an \textit{average of 40k} pairwise cosine similarities.}}
    \label{fig:embedding_similarity}
\end{figure}

\begin{figure*}[ht]
    \centering
    \includegraphics[width=1\linewidth]{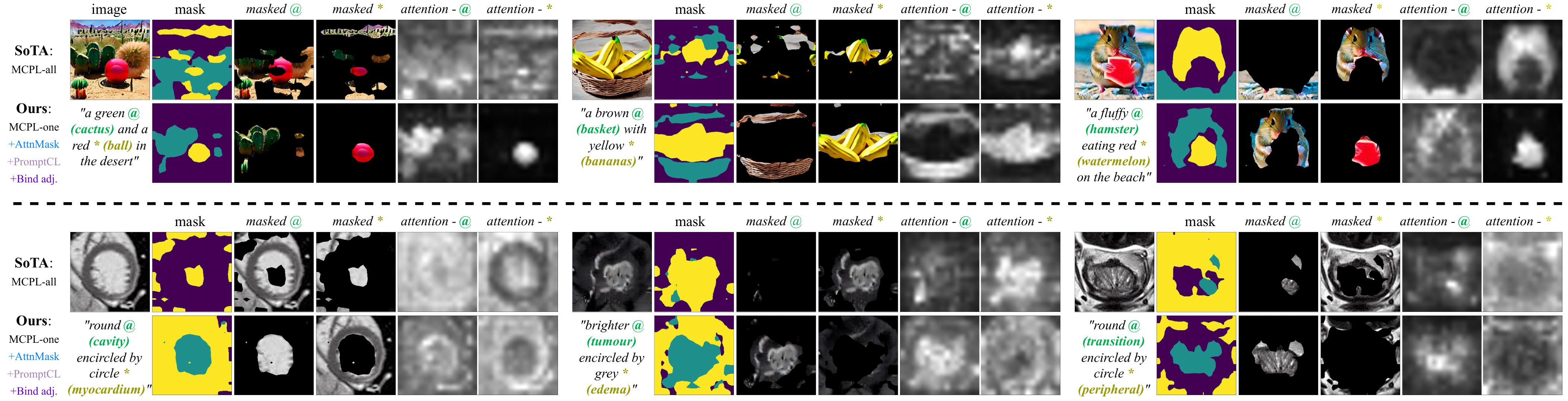}
    \vspace{-6mm}
    \caption{\textbf{Visualisation of generated concepts with the MCPL-all (top) and our fully regularised method (bottom).} Masks are derived from cross-attentions. Full ablation results are presented in the Appendix \ref{sec: full_ablation}}
    \label{fig:visualise_gen_masks}
\end{figure*}

The results in \Figref{fig:embedding_similarity} show our method combined with all the proposed regularisation terms can improve both prompt and image fidelity consistently.
Our fully regularised version (\textit{MCPL-one+CL+Mask}) achieved competitive performance compared to the SoTA mask-based method (BAS) on the natural dataset. In the OOD medical dataset, BAS outperformed our method significantly in the DINOv1 embedding space, although the performance was comparable in other spaces. 
The discrepancy stems from our method's attention masks yielding less accurate object masks compared to \textit{BAS whose masks are obtained through a specialised human-in-the-loop segmentation protocol}, as detailed in Appendix \ref{sec: bas_exp_setup}. This difference is depicted in Figures \ref{fig:quantitative_dataset_highlight}, \ref{fig:concepts_345_dataset} and \ref{fig:visualise_gen_masks}.

\begin{figure}[h]
  \centering
  \includegraphics[width=\linewidth]{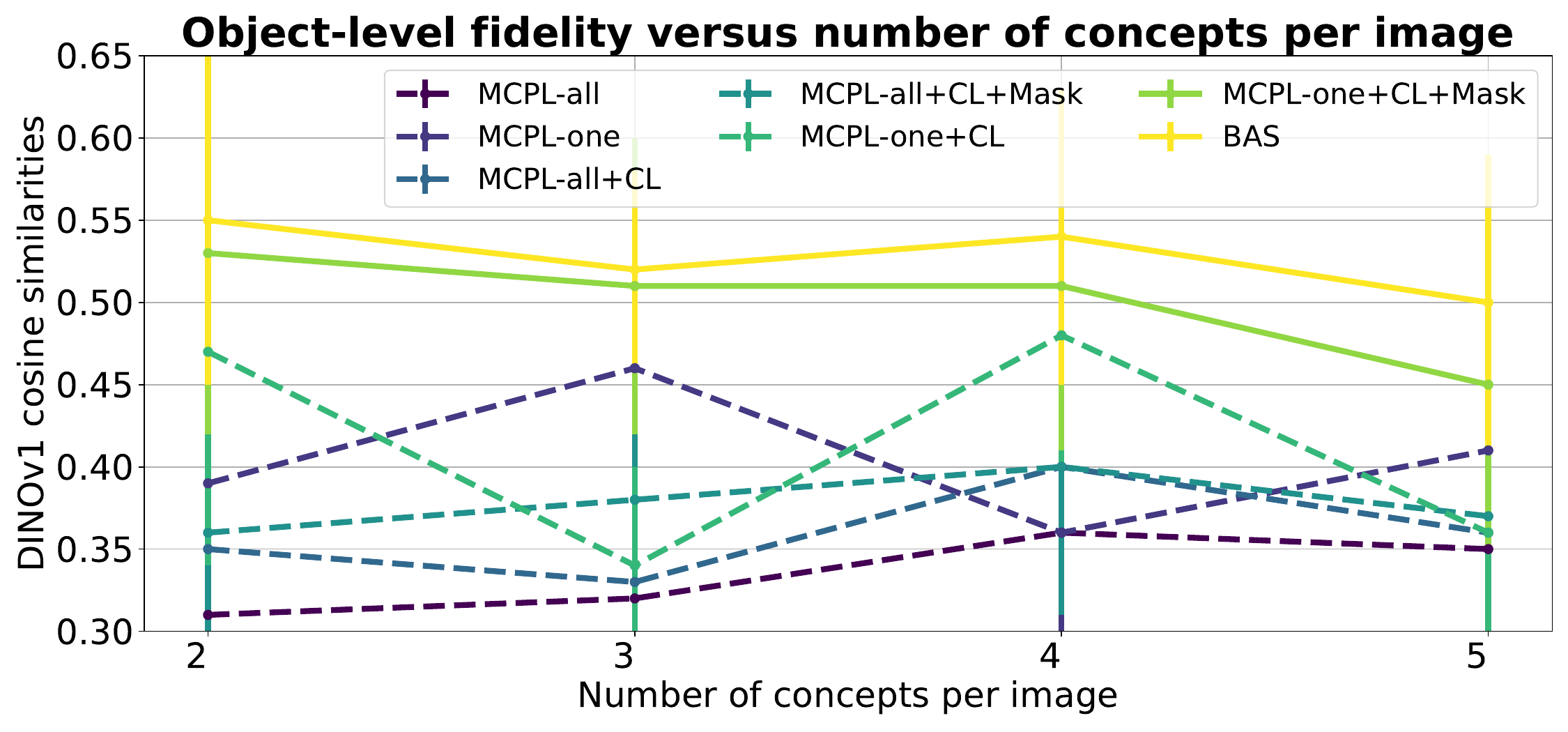}
  \vspace{-8mm}
  \caption{\footnotesize{\textbf{Evaluate the learning as the number of concepts per image increases.} Here each data point represents an \textit{average of 20$\sim$40k} pairwise cosine similarities measured by DINOv1.}}
    \label{fig:embedding_similarity_345}
\end{figure}

\paragraph{Learning more than two concepts.}
To validate our method's robustness, we expanded the evaluation to learning tasks with more than two concepts per image, specifically natural images containing 3, 4, and 5 concepts. 
We group the learned embeddings by the number of concepts per image to evaluate their impact on learning efficiency. 
The results in \Figref{fig:embedding_similarity_345} reveal that: 1) learning efficiency diminishes with the increase in the number of concepts per image, a trend also evident in the mask-based BAS approach; 2) although our fully regularised version continues to outperform under-regularised versions, the performance gap to BAS widens, highlighting the heightened challenge of mask-free multi-concept learning in more complex scenes. 
We also collect real image datasets for the same evaluation and obtain consistent conclusions, with full details in Appendix \ref{sec:real_images}.

\begin{figure}[ht]
  \centering
  \includegraphics[width=\linewidth]{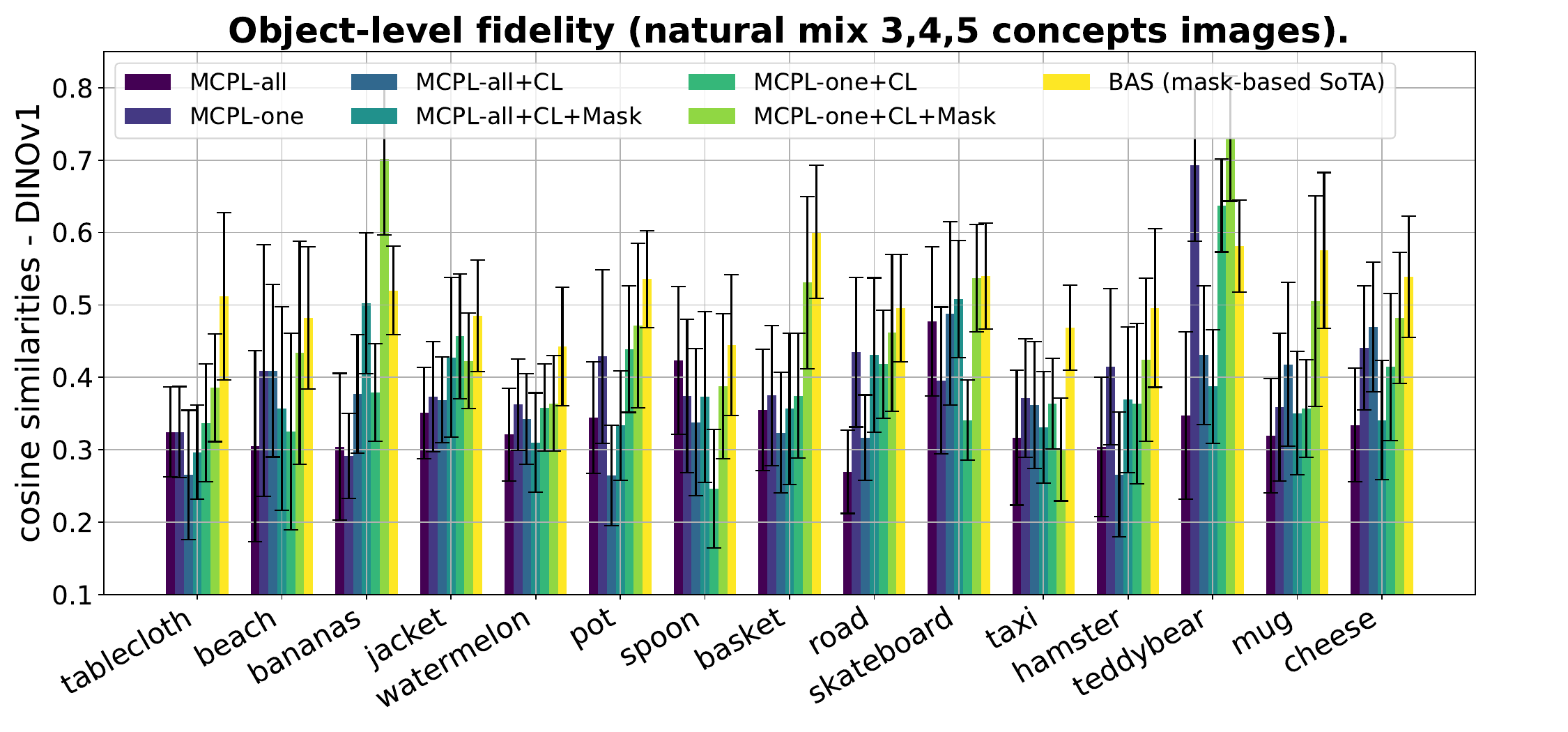}
  \vspace{-8mm}
  \caption{\footnotesize{\textbf{Object-level embedding similarities with DINOv1 (mix of 3 to 5 concepts per image).} Each bar representing an \textit{average of 160k} pairwise cosine similarities. A comprehensive object-level comparison is available in the Appendix (\Secref{sec: emb_similarity_all_objects}).}}
    \label{fig:object_similarity_345}
\end{figure}

\paragraph{Object-level evaluations.}
As our main goal is to accurately learn object-level embeddings, we perform an analysis of embedding similarity at the object level as highlighted in \Figref{fig:object_similarity_345}. 
Notably, our method sometimes surpasses the mask-based method, BAS, at the object level. \textit{This underscores the potential of our mask-free, language-driven approach to learning multiple concepts from a single image.}


\paragraph{User studies.}
Lastly, we collect 41 anonymized user studies to evaluate prompt-following during image editing. Each study encompassed three types of tasks: text alignment, prompt following, and semantic correspondence, totalling 30 quizzes. The user studies further justify our method, with full results in Appendix \ref{sec: user}.
 
\subsection{Qualitative Evaluation}
\label{sec: applications}

\begin{figure}[h]
    \centering
    \includegraphics[width=1\linewidth]{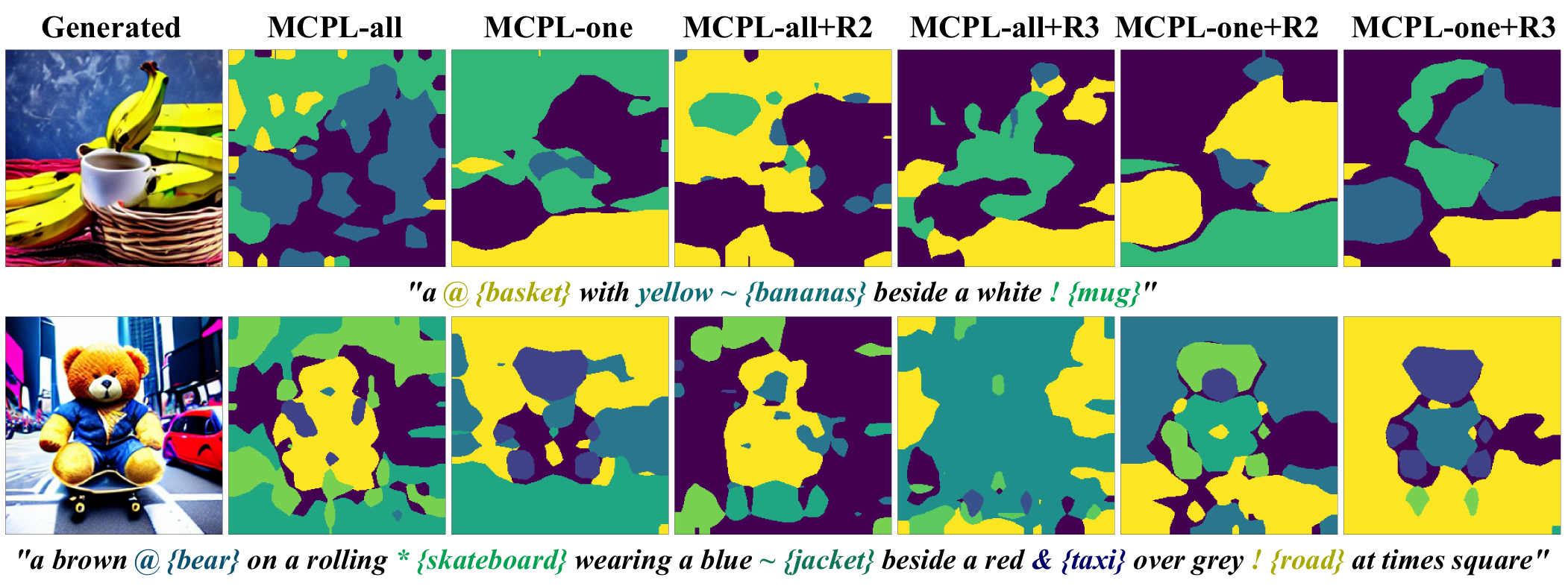}
    \vspace{-6mm}
    \caption{\textbf{Visualisation of generated concepts with the ``SoTA" and our method (3 or 5 concepts).} Masks are derived from cross-attentions. Full ablation results are presented in the Appendix \ref{sec: full_ablation}}
    \label{fig:comcepts_345_segment_highlight}
\end{figure}

\begin{figure}[h]
    \centering
    \includegraphics[width=1\linewidth]{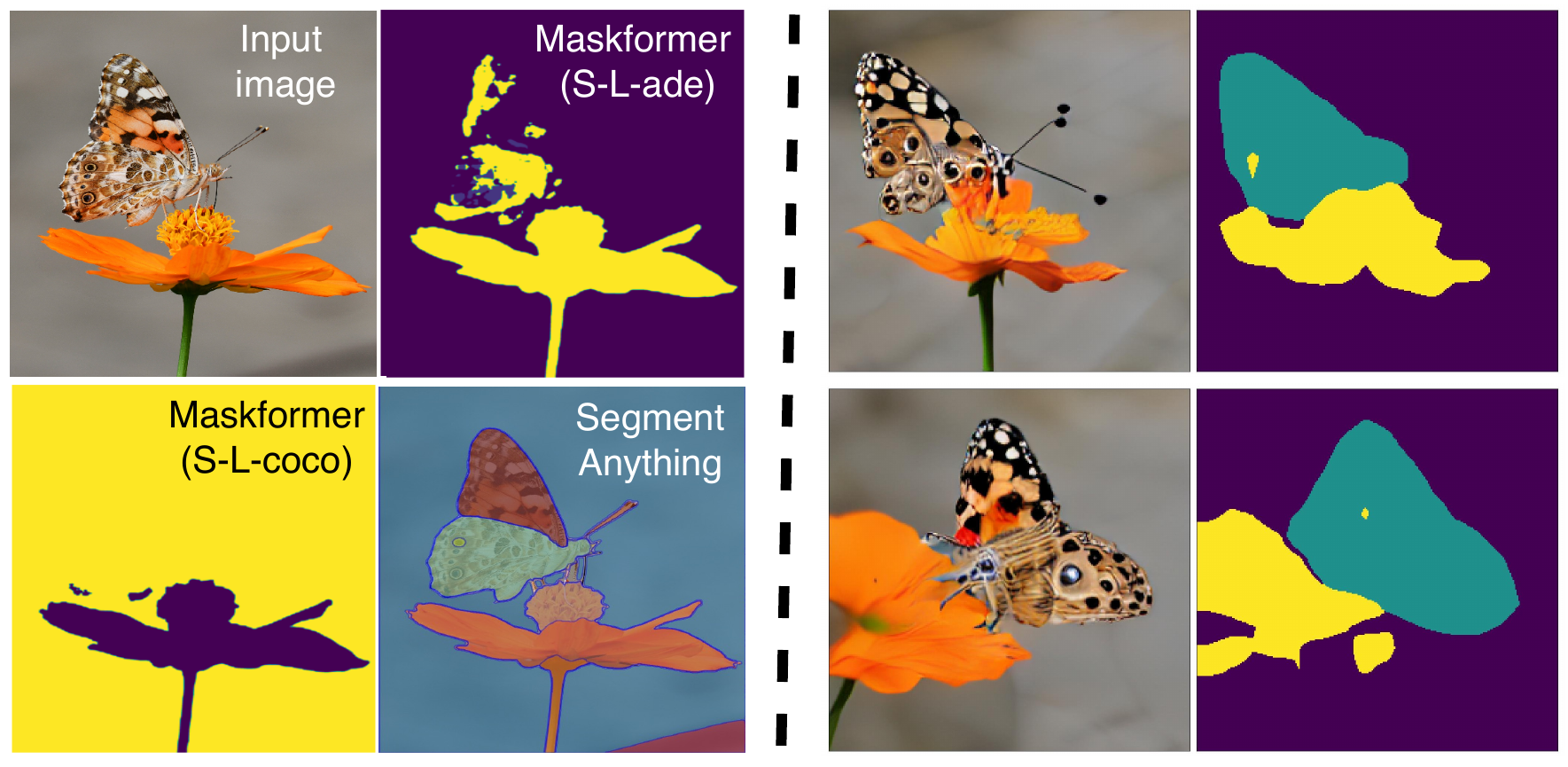}
    \textit{``a patterned \textcolor{teal}{\texttt{@} (butterfly)} resting on an orange \textcolor{olive}{\texttt{*} (flower)}''}
    \vspace{-2mm}
    \caption{
    \textbf{Learning concepts with a similar appearance.} MCPL's fully regularised version (right) demonstrates competitive cross-attention masks against segmentation models. Full stress tests are presented in the Appendix \ref{sec: challenge_cases}.
    }
    \label{fig:butterfly_highlight}
\end{figure}

\paragraph{Visualise concepts disentanglement and learning.}
To evaluate disentanglement and prompt-to-concept correlation, we visualise attention and attention masks for learnable prompts. Figures \ref{fig:visualise_gen_masks} and \ref{fig:comcepts_345_segment_highlight} display results for both natural and medical images. The visual outcomes align with earlier quantitative findings, affirming the effectiveness of our proposed MCPL method and regularisation terms.
In \Figref{fig:butterfly_highlight} our method demonstrates capability in learning concepts having similar colour, with cross-attention derived masks outperforming pre-trained segmentation models in terms of semantic accuracy.

\begin{figure*}[ht]
    \centering
    \includegraphics[width=\linewidth]{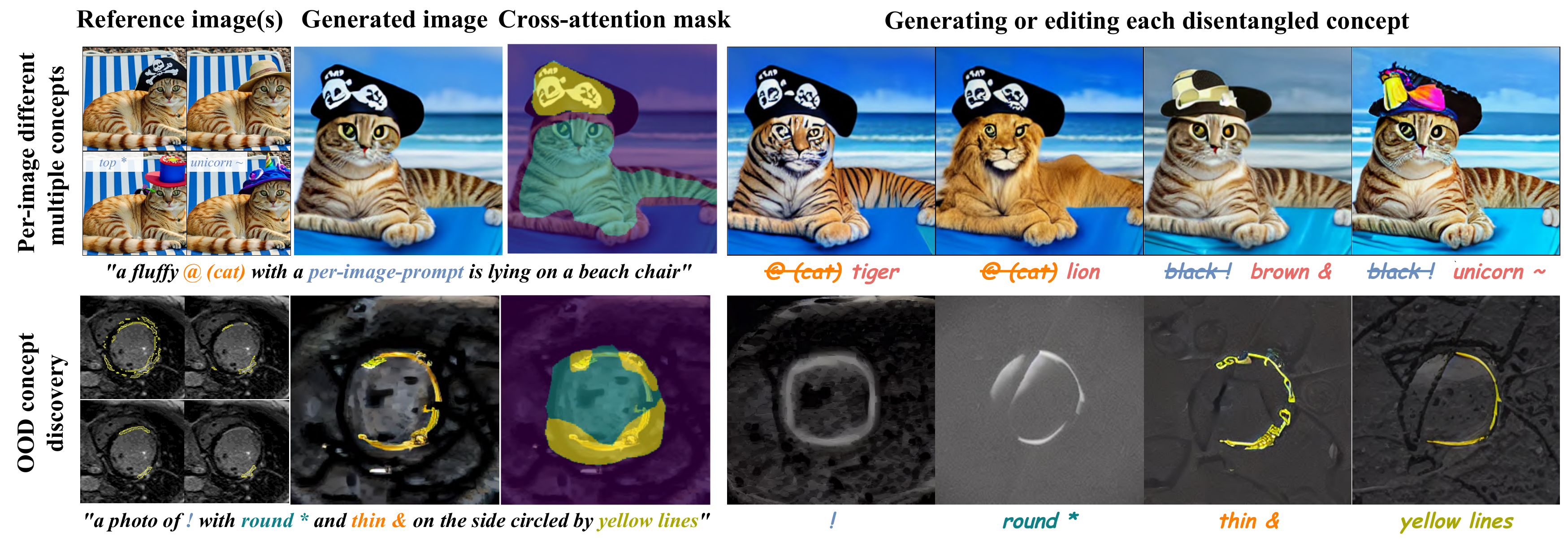}
    \vspace{-4mm}
    \caption{
    \textbf{MCPL learning and editing capabilities.}
    Top-row: discovering per-image different concepts with per-image specified string with input image from P2P \citep{hertz2022prompt}. Bottom-row: learning to disentangle multiple unseen concepts from cardiac MRI images with input images from LGE-CMR \citep{karim2016evaluation}. More examples in \Figref{fig:editing_capabilities}.
    \label{fig:editing_capabilities_sim}}
\end{figure*}

\paragraph{Image editing over disentangled concepts.}
Hypothesis generation in scientific fields can be accelerated by integrating new concepts into existing observations, a process that benefits from local image editing.
We demonstrate our method enables \textit{mask-free object-level learning, editing and quantification} (\Figref{fig:editing_capabilities_sim} top-row), with flexibility to \textit{handle per-image specified string to learn the different concepts within each image}.
Furthermore, our method can also learn unknown concepts from challenging out-of-distribution images (\Figref{fig:editing_capabilities_sim} bottom rows), \textit{opening an avenue of knowledge mining from pairs of textbook figures and captions}.
It is worth noting that, compared to BAS, our method does not rely on a separate segmentation model and mask to achieve local editing. Our method optimises the disentanglement of multiple concepts, leading to accurate word-concept correlation in the cross-attention hence supporting mask-free local editing method (e.g. P2P \cite{hertz2022prompt}) directly.

\paragraph{Learning abstract concepts.}
In this Appendix \ref{sec:compare_inspiration_tree}., we compare MCPL with Inspiration Tree (IT) \citep{vinker2023concept}, showcases MCPL's performance in both concepts exploration \Figref{fig:mcpl_inspiretree_explore} and combination \Figref{fig:mcpl_inspiretree_combine}.

\subsection{Ablation studies.}
\label{sec: ablation_intro}
We also conduct a set of ablation studies to assess various components and capabilities of our method, with details in the Appendix. They are: 
1) The \textit{MCPL-diverse} training strategy has demonstrated potential in learning tasks with varying concepts per image. Therefore, we performed further experiments to assess its effectiveness, with findings detailed in \Secref{sec: ablation_diverse_vs_one} confirming its efficacy.
2) Our language-driven approach benefits from the proposed adjective binding mechanism. To better understand its role, we conducted an ablation study detailed in \Secref{sec: ablation_adjective}, which confirmed its significance.
3) For a comprehensive evaluation, we visually compare our tuning-free method, which is not specifically designed for composing complex scenes when prompt interactions change, with SoTA composing-focused methods in complex scene tasks. This comparison is detailed in \Secref{sec: bas_compare} with promising results.

\section{Limitations and Conclusions}
MCPL enhances prompt-region semantic correlation through natural language instructions but may encounter difficulties in scenarios such as:
1) When concepts are linguistically indistinct, for example, multiple identical or highly similar instances that natural language struggles to differentiate.
2) In highly complex scenes containing many concepts with limited example images available, a challenge recognized by \cite{liu2023cones} and \cite{avrahami2023break}.

In conclusion, we introduced MCPL to tackle the novel challenge of mask-free learning of multiple concepts using images and natural language descriptions. This approach is expected to assist in the discovery of new concepts through natural language-driven human-machine interaction, potentially advancing task hypothesis generation and local image editing without requiring explicit knowledge of the new vision concept.

\newpage

\newpage

\section*{Impact Statement}
This paper presents work whose goal is to advance the field of Machine Learning. There are many potential societal consequences of our work, none which we feel must be specifically highlighted here.

\bibliography{icml2024_conference}

\begin{thebibliography}{33}
\providecommand{\natexlab}[1]{#1}
\providecommand{\url}[1]{\texttt{#1}}
\expandafter\ifx\csname urlstyle\endcsname\relax
  \providecommand{\doi}[1]{doi: #1}\else
  \providecommand{\doi}{doi: \begingroup \urlstyle{rm}\Url}\fi

\bibitem[Antonelli et~al.(2022)Antonelli, Reinke, Bakas, Farahani, Kopp-Schneider, Landman, Litjens, Menze, Ronneberger, Summers, et~al.]{antonelli2022medical}
Antonelli, M., Reinke, A., Bakas, S., Farahani, K., Kopp-Schneider, A., Landman, B.~A., Litjens, G., Menze, B., Ronneberger, O., Summers, R.~M., et~al.
\newblock The medical segmentation decathlon.
\newblock \emph{Nature communications}, 13\penalty0 (1):\penalty0 4128, 2022.

\bibitem[Avrahami et~al.(2023)Avrahami, Aberman, Fried, Cohen-Or, and Lischinski]{avrahami2023break}
Avrahami, O., Aberman, K., Fried, O., Cohen-Or, D., and Lischinski, D.
\newblock Break-a-scene: Extracting multiple concepts from a single image.
\newblock \emph{arXiv preprint arXiv:2305.16311}, 2023.

\bibitem[Caron et~al.(2021)Caron, Touvron, Misra, Jégou, Mairal, Bojanowski, and Joulin]{caron2021emerging}
Caron, M., Touvron, H., Misra, I., Jégou, H., Mairal, J., Bojanowski, P., and Joulin, A.
\newblock Emerging properties in self-supervised vision transformers, 2021.

\bibitem[Chen et~al.(2020)Chen, Kornblith, Norouzi, and Hinton]{chen2020simple}
Chen, T., Kornblith, S., Norouzi, M., and Hinton, G.
\newblock A simple framework for contrastive learning of visual representations.
\newblock In \emph{International conference on machine learning}, pp.\  1597--1607. PMLR, 2020.

\bibitem[Cheng et~al.(2021)Cheng, Schwing, and Kirillov]{cheng2021per}
Cheng, B., Schwing, A., and Kirillov, A.
\newblock Per-pixel classification is not all you need for semantic segmentation.
\newblock \emph{Advances in Neural Information Processing Systems}, 34:\penalty0 17864--17875, 2021.

\bibitem[Devlin et~al.(2018)Devlin, Chang, Lee, and Toutanova]{devlin2018bert}
Devlin, J., Chang, M.-W., Lee, K., and Toutanova, K.
\newblock Bert: Pre-training of deep bidirectional transformers for language understanding.
\newblock \emph{arXiv preprint arXiv:1810.04805}, 2018.

\bibitem[Gal et~al.(2022)Gal, Alaluf, Atzmon, Patashnik, Bermano, Chechik, and Cohen-Or]{gal2022textual}
Gal, R., Alaluf, Y., Atzmon, Y., Patashnik, O., Bermano, A.~H., Chechik, G., and Cohen-Or, D.
\newblock An image is worth one word: Personalizing text-to-image generation using textual inversion, 2022.
\newblock URL \url{https://arxiv.org/abs/2208.01618}.

\bibitem[Han et~al.(2023)Han, Li, Zhang, Milanfar, Metaxas, and Yang]{han2023svdiff}
Han, L., Li, Y., Zhang, H., Milanfar, P., Metaxas, D., and Yang, F.
\newblock Svdiff: Compact parameter space for diffusion fine-tuning.
\newblock \emph{arXiv preprint arXiv:2303.11305}, 2023.

\bibitem[Hertz et~al.(2022)Hertz, Mokady, Tenenbaum, Aberman, Pritch, and Cohen-Or]{hertz2022prompt}
Hertz, A., Mokady, R., Tenenbaum, J., Aberman, K., Pritch, Y., and Cohen-Or, D.
\newblock Prompt-to-prompt image editing with cross attention control.
\newblock 2022.

\bibitem[Johnson et~al.(2019)Johnson, Pollard, Berkowitz, Greenbaum, Lungren, Deng, Mark, and Horng]{johnson2019mimic}
Johnson, A.~E., Pollard, T.~J., Berkowitz, S.~J., Greenbaum, N.~R., Lungren, M.~P., Deng, C.-y., Mark, R.~G., and Horng, S.
\newblock Mimic-cxr, a de-identified publicly available database of chest radiographs with free-text reports.
\newblock \emph{Scientific data}, 6\penalty0 (1):\penalty0 317, 2019.

\bibitem[Karim et~al.(2016)Karim, Bhagirath, Claus, Housden, Chen, Karimaghaloo, Sohn, Rodr{\'\i}guez, Vera, Alb{\`a}, et~al.]{karim2016evaluation}
Karim, R., Bhagirath, P., Claus, P., Housden, R.~J., Chen, Z., Karimaghaloo, Z., Sohn, H.-M., Rodr{\'\i}guez, L.~L., Vera, S., Alb{\`a}, X., et~al.
\newblock Evaluation of state-of-the-art segmentation algorithms for left ventricle infarct from late gadolinium enhancement mr images.
\newblock \emph{Medical image analysis}, 30:\penalty0 95--107, 2016.

\bibitem[Kirillov et~al.(2023)Kirillov, Mintun, Ravi, Mao, Rolland, Gustafson, Xiao, Whitehead, Berg, Lo, et~al.]{kirillov2023segment}
Kirillov, A., Mintun, E., Ravi, N., Mao, H., Rolland, C., Gustafson, L., Xiao, T., Whitehead, S., Berg, A.~C., Lo, W.-Y., et~al.
\newblock Segment anything.
\newblock \emph{arXiv preprint arXiv:2304.02643}, 2023.

\bibitem[Kumari et~al.(2023)Kumari, Zhang, Zhang, Shechtman, and Zhu]{kumari2023multi}
Kumari, N., Zhang, B., Zhang, R., Shechtman, E., and Zhu, J.-Y.
\newblock Multi-concept customization of text-to-image diffusion.
\newblock In \emph{Proceedings of the IEEE/CVF Conference on Computer Vision and Pattern Recognition}, pp.\  1931--1941, 2023.

\bibitem[Lalande et~al.(2020)Lalande, Chen, Decourselle, Qayyum, Pommier, Lorgis, de~La~Rosa, Cochet, Cottin, Ginhac, et~al.]{lalande2020emidec}
Lalande, A., Chen, Z., Decourselle, T., Qayyum, A., Pommier, T., Lorgis, L., de~La~Rosa, E., Cochet, A., Cottin, Y., Ginhac, D., et~al.
\newblock Emidec: a database usable for the automatic evaluation of myocardial infarction from delayed-enhancement cardiac mri.
\newblock \emph{Data}, 5\penalty0 (4):\penalty0 89, 2020.

\bibitem[Lin et~al.(2014)Lin, Maire, Belongie, Hays, Perona, Ramanan, Doll{\'a}r, and Zitnick]{lin2014microsoft}
Lin, T.-Y., Maire, M., Belongie, S., Hays, J., Perona, P., Ramanan, D., Doll{\'a}r, P., and Zitnick, C.~L.
\newblock Microsoft coco: Common objects in context.
\newblock In \emph{Computer Vision--ECCV 2014: 13th European Conference, Zurich, Switzerland, September 6-12, 2014, Proceedings, Part V 13}, pp.\  740--755. Springer, 2014.

\bibitem[Liu et~al.(2023)Liu, Feng, Zhu, Zhang, Zheng, Liu, Zhao, Zhou, and Cao]{liu2023cones}
Liu, Z., Feng, R., Zhu, K., Zhang, Y., Zheng, K., Liu, Y., Zhao, D., Zhou, J., and Cao, Y.
\newblock Cones: Concept neurons in diffusion models for customized generation.
\newblock \emph{arXiv preprint arXiv:2303.05125}, 2023.

\bibitem[Ma \& Wang(2023)Ma and Wang]{ma2023segment}
Ma, J. and Wang, B.
\newblock Segment anything in medical images.
\newblock \emph{arXiv preprint arXiv:2304.12306}, 2023.

\bibitem[Menze et~al.(2014)Menze, Jakab, Bauer, Kalpathy-Cramer, Farahani, Kirby, Burren, Porz, Slotboom, Wiest, et~al.]{menze2014multimodal}
Menze, B.~H., Jakab, A., Bauer, S., Kalpathy-Cramer, J., Farahani, K., Kirby, J., Burren, Y., Porz, N., Slotboom, J., Wiest, R., et~al.
\newblock The multimodal brain tumor image segmentation benchmark (brats).
\newblock \emph{IEEE transactions on medical imaging}, 34\penalty0 (10):\penalty0 1993--2024, 2014.

\bibitem[Oord et~al.(2018)Oord, Li, and Vinyals]{oord2018representation}
Oord, A. v.~d., Li, Y., and Vinyals, O.
\newblock Representation learning with contrastive predictive coding.
\newblock \emph{arXiv preprint arXiv:1807.03748}, 2018.

\bibitem[Oquab et~al.(2023)Oquab, Darcet, Moutakanni, Vo, Szafraniec, Khalidov, Fernandez, Haziza, Massa, El-Nouby, et~al.]{oquab2023dinov2}
Oquab, M., Darcet, T., Moutakanni, T., Vo, H., Szafraniec, M., Khalidov, V., Fernandez, P., Haziza, D., Massa, F., El-Nouby, A., et~al.
\newblock Dinov2: Learning robust visual features without supervision.
\newblock \emph{arXiv preprint arXiv:2304.07193}, 2023.

\bibitem[Patashnik et~al.(2023)Patashnik, Garibi, Azuri, Averbuch-Elor, and Cohen-Or]{patashnik2023localizing}
Patashnik, O., Garibi, D., Azuri, I., Averbuch-Elor, H., and Cohen-Or, D.
\newblock Localizing object-level shape variations with text-to-image diffusion models, 2023.

\bibitem[Radford et~al.(2021)Radford, Kim, Hallacy, Ramesh, Goh, Agarwal, Sastry, Askell, Mishkin, Clark, Krueger, and Sutskever]{radford2021learning}
Radford, A., Kim, J.~W., Hallacy, C., Ramesh, A., Goh, G., Agarwal, S., Sastry, G., Askell, A., Mishkin, P., Clark, J., Krueger, G., and Sutskever, I.
\newblock Learning transferable visual models from natural language supervision, 2021.

\bibitem[Rombach et~al.(2022)Rombach, Blattmann, Lorenz, Esser, and Ommer]{rombach2022high}
Rombach, R., Blattmann, A., Lorenz, D., Esser, P., and Ommer, B.
\newblock High-resolution image synthesis with latent diffusion models.
\newblock In \emph{Proceedings of the IEEE/CVF conference on computer vision and pattern recognition}, pp.\  10684--10695, 2022.

\bibitem[Ruiz et~al.(2022)Ruiz, Li, Jampani, Pritch, Rubinstein, and Aberman]{ruiz2022dreambooth}
Ruiz, N., Li, Y., Jampani, V., Pritch, Y., Rubinstein, M., and Aberman, K.
\newblock Dreambooth: Fine tuning text-to-image diffusion models for subject-driven generation.
\newblock 2022.

\bibitem[Schickore(2022)]{sep-scientific-discovery}
Schickore, J.
\newblock {Scientific Discovery}.
\newblock In Zalta, E.~N. and Nodelman, U. (eds.), \emph{The {Stanford} Encyclopedia of Philosophy}. Metaphysics Research Lab, Stanford University, {W}inter 2022 edition, 2022.

\bibitem[Tewel et~al.(2023)Tewel, Gal, Chechik, and Atzmon]{tewel2023key}
Tewel, Y., Gal, R., Chechik, G., and Atzmon, Y.
\newblock Key-locked rank one editing for text-to-image personalization.
\newblock In \emph{ACM SIGGRAPH 2023 Conference Proceedings}, pp.\  1--11, 2023.

\bibitem[Tumanyan et~al.(2023)Tumanyan, Geyer, Bagon, and Dekel]{tumanyan2023plug}
Tumanyan, N., Geyer, M., Bagon, S., and Dekel, T.
\newblock Plug-and-play diffusion features for text-driven image-to-image translation.
\newblock In \emph{Proceedings of the IEEE/CVF Conference on Computer Vision and Pattern Recognition}, pp.\  1921--1930, 2023.

\bibitem[Van~der Maaten \& Hinton(2008)Van~der Maaten and Hinton]{van2008visualizing}
Van~der Maaten, L. and Hinton, G.
\newblock Visualizing data using t-sne.
\newblock \emph{Journal of machine learning research}, 9\penalty0 (11), 2008.

\bibitem[Vinker et~al.(2023)Vinker, Voynov, Cohen-Or, and Shamir]{vinker2023concept}
Vinker, Y., Voynov, A., Cohen-Or, D., and Shamir, A.
\newblock Concept decomposition for visual exploration and inspiration.
\newblock \emph{ACM Transactions on Graphics (TOG)}, 42\penalty0 (6):\penalty0 1--13, 2023.

\bibitem[Wang et~al.(2023)Wang, Fu, Du, Gao, Huang, Liu, Chandak, Liu, Van~Katwyk, Deac, et~al.]{wang2023scientific}
Wang, H., Fu, T., Du, Y., Gao, W., Huang, K., Liu, Z., Chandak, P., Liu, S., Van~Katwyk, P., Deac, A., et~al.
\newblock Scientific discovery in the age of artificial intelligence.
\newblock \emph{Nature}, 620\penalty0 (7972):\penalty0 47--60, 2023.

\bibitem[Wei et~al.(2023)Wei, Zhang, Ji, Bai, Zhang, and Zuo]{wei2023elite}
Wei, Y., Zhang, Y., Ji, Z., Bai, J., Zhang, L., and Zuo, W.
\newblock Elite: Encoding visual concepts into textual embeddings for customized text-to-image generation.
\newblock \emph{arXiv preprint arXiv:2302.13848}, 2023.

\bibitem[Wu et~al.(2020)Wu, Lischinski, and Shechtman]{wu2020stylespace}
Wu, Z., Lischinski, D., and Shechtman, E.
\newblock Stylespace analysis: Disentangled controls for stylegan image generation, 2020.

\bibitem[Zhou et~al.(2017)Zhou, Zhao, Puig, Fidler, Barriuso, and Torralba]{zhou2017scene}
Zhou, B., Zhao, H., Puig, X., Fidler, S., Barriuso, A., and Torralba, A.
\newblock Scene parsing through ade20k dataset.
\newblock In \emph{Proceedings of the IEEE conference on computer vision and pattern recognition}, pp.\  633--641, 2017.

\end{thebibliography}
\bibliographystyle{icml2024_conference}

\newpage
\onecolumn
\appendix
\section{Appendix}
\label{sec: appn}

\addtocontents{toc}{\protect\setcounter{tocdepth}{2}}

\subsection{User study for instruction following with prompt guided image editing.}
\label{sec: user}

\begin{figure}[H]
    \centering
    \begin{minipage}{0.334\linewidth}
        \includegraphics[width=\linewidth]{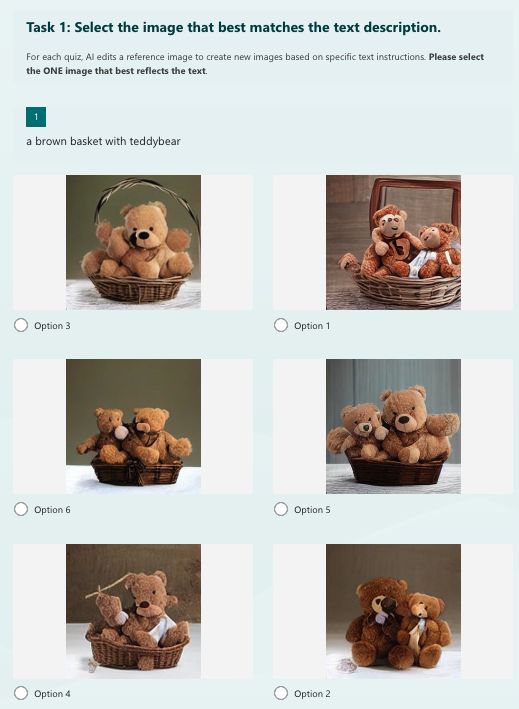}
    \end{minipage}
    \begin{minipage}{0.275\linewidth}
        \includegraphics[width=\linewidth]{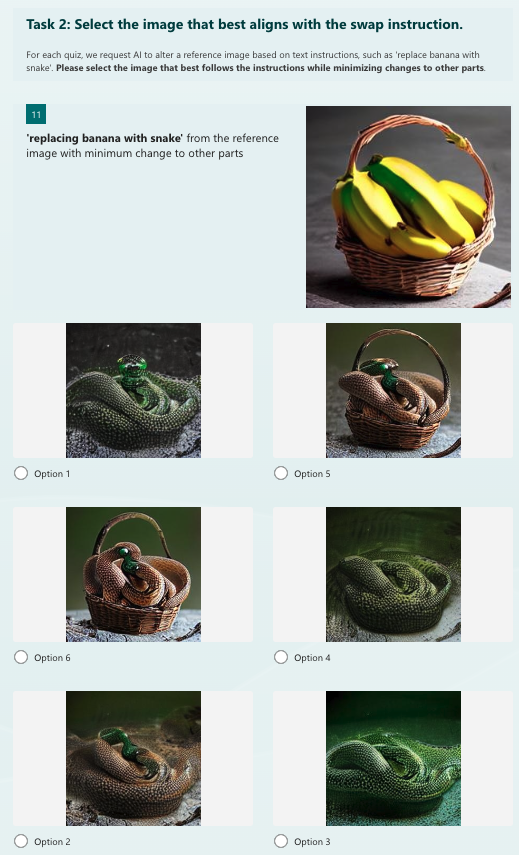}
    \end{minipage}
    \begin{minipage}{0.2675\linewidth}
        \includegraphics[width=\linewidth]{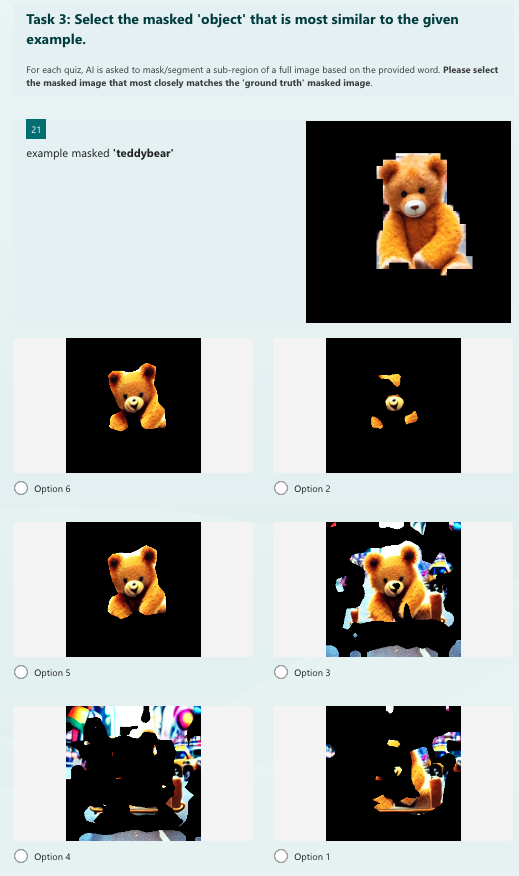}
    \end{minipage}
    \caption{
        Example of anonymized user studies of text alignment (left),  prompt following (middle) and semantic correspondence (right). 
        }
    \label{fig:user-study-example}
\end{figure}

\begin{itemize}
    \item We conduct anonymized user studies to evaluate \textbf{text alignment}, following Cones \citep{liu2023cones}, comparing edited images to baselines. Participants are instructed to select the image that most accurately represents the given text, with the sentence's learnable pseudo word replaced by specific editing terms, see the left example in \Figref{fig:user-study-example}.
    \item To assess the \textbf{prompt following}, we present the user the original image, followed by the edited versions with prompt instructions (e.g. replacing `cat’ with `panda’). We ask the users to select ``the image that best follows the instructions while minimizing changes to other parts”, see the middle example in \Figref{fig:user-study-example}.
    \item \textbf{Semantic correspondence} is evaluated by displaying the target prompt alongside both the masked ground truth and its edited counterpart, obscured by corresponding prompt cross-attention. Users are asked to identify the masked image that most closely resembles the 'ground truth'. This evaluation captures the accuracy of prompt-region correlation and the quality of generated objects, aligning with our research goals.
\end{itemize}

\begin{table}[ht]
\centering
\footnotesize
\begin{tabular}{@{}lcccccc@{}}
\toprule
Method & MCPL-all & MCPL-one & MCPL-all & MCPL-all & MCPL-one & MCPL-one \\
 &  &  & +PromptCL & +PromptCL & +PromptCL & +PromptCL \\
 &  &  &  & +AttnMask &  & +AttnMask \\
\midrule
Text Alignment (\%) & 7.18 & 3.33 & 7.95 & 25.64 & 15.38 & \textbf{40.51} \\
Prompt Following (\%) & 3.08 & 5.64 & 2.31 & 11.54 & 21.28 & \textbf{56.15} \\
Semantic Correspondence (natural images) (\%) & 3.59 & 7.18 & 3.08 & 5.13 & 26.67 & \textbf{54.36 }\\
Semantic Correspondence (medical images) (\%) & 6.15 & 5.13 & 6.67 & 3.08 & 13.33 & \textbf{65.64} \\
\bottomrule
\end{tabular}
\caption{
\textbf{User Study on Language-Driven Image Editing.} 
We collected 41 user studies, each study consists of three types of tasks (text alignment,  prompt following and semantic correspondence), each 10 and a total of 30 quizzes. For the semantic, 5 quizzes are from natural images and 5 from medical images. The value represents the percentage of users who think the image generated by the corresponding method is the best, following Cones \citep{liu2023cones}. All images are generated and edited with models learnt in our quantitative evaluation as presented in \Secref{sec: quantitative}. 
}
\label{tab:user_study}
\end{table}

\begin{figure}[H]
    \centering
    \begin{minipage}{0.495\linewidth}
        \includegraphics[width=\linewidth]{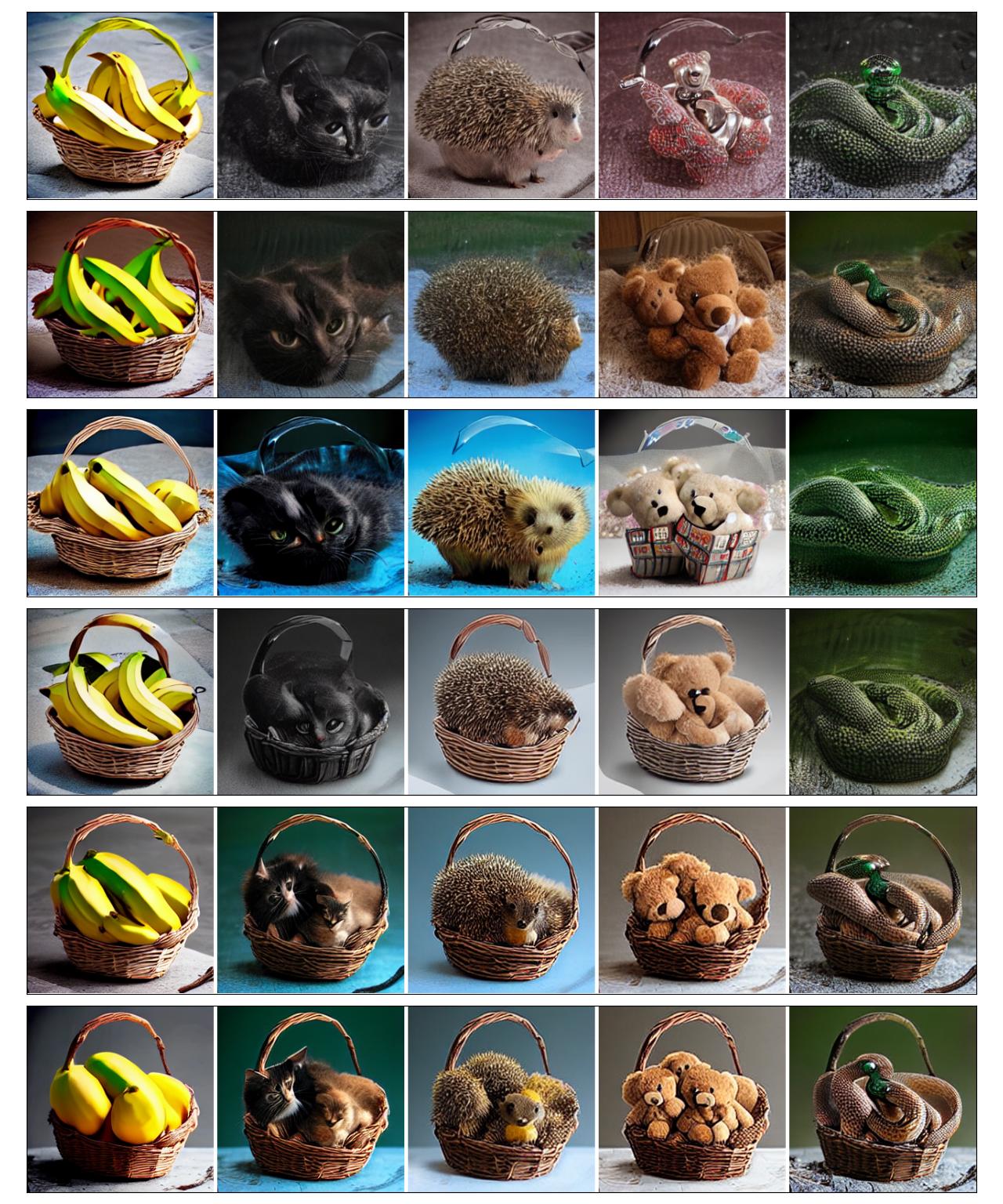}
        \footnotesize{Visual examples demonstrate the replacement of
        'bananas' with 'cat,' 'hedgehog,' 'teddy,' and 'snake,'
        sequentially from left to right.}
        \label{fig:first-editting}
    \end{minipage}
    \hfill 
    \begin{minipage}{0.495\linewidth}
        \includegraphics[width=\linewidth]{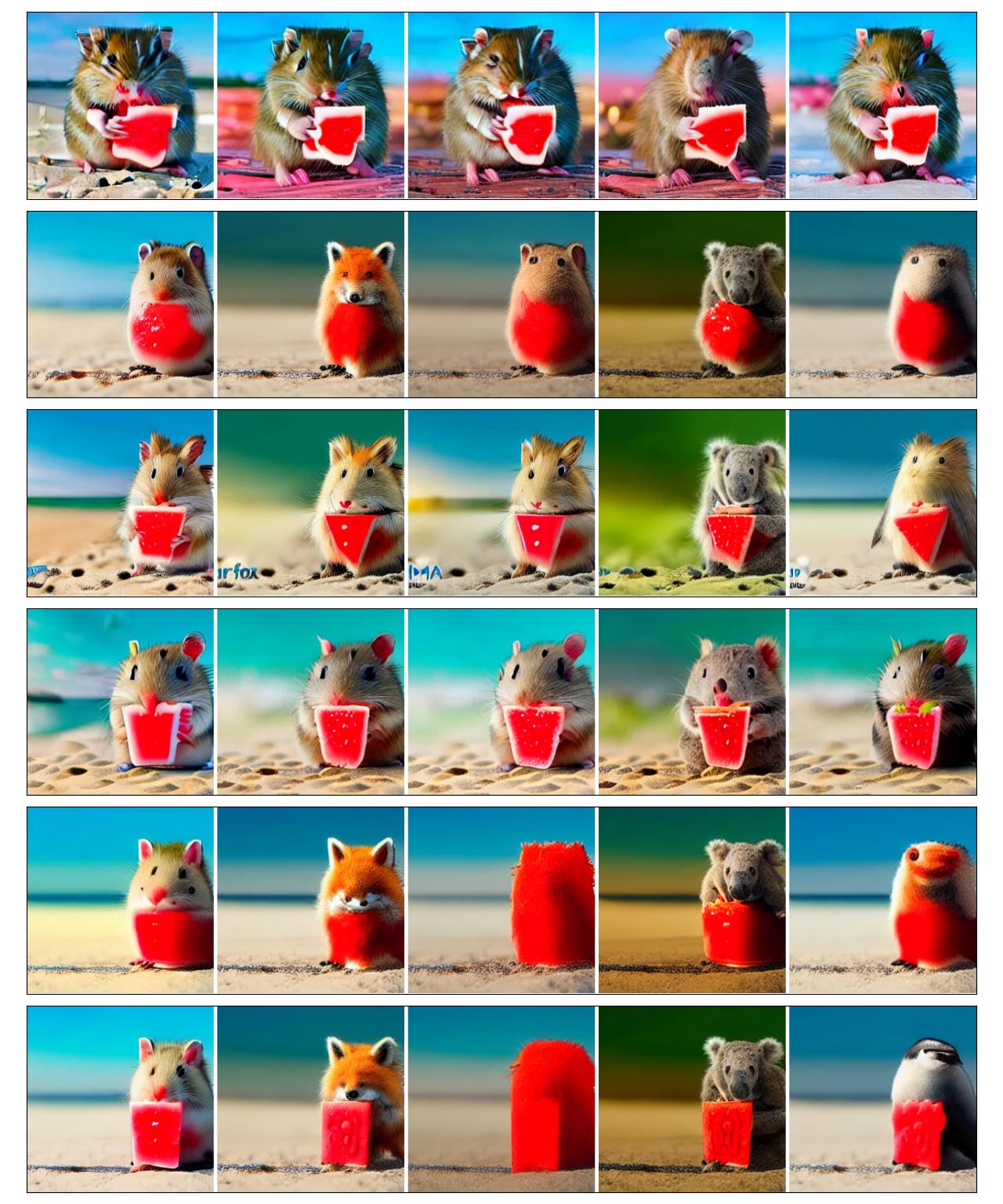}
        \footnotesize{Visual examples demonstrate the replacement of
        'hamster' with 'fox,' 'lamma,' 'koala,' and 'penguin,'
        sequentially from left to right.}
        \label{fig:second-editting}
    \end{minipage}
    \caption{
        Visualisation of text-guided image editing. 
        We compare all baseline methods across each row (from top to bottom): 
        1) MCPL-all, 
        2) MCPL-one, 
        3) MCPL-all+\textit{PromptCL}+\textit{Bind adj.}, 
        4) MCPL-all+\textit{AttnMask}+\textit{PromptCL}+\textit{Bind adj.}, 
        5) MCPL-one+\textit{PromptCL}+\textit{Bind adj.}, 
        6) MCPL-one+\textit{AttnMask}+\textit{PromptCL}+\textit{Bind adj.}.
        \textbf{The results corroborated the quantitative findings from the user study}: the fully regularized MCPL enhances the accuracy of prompt instruction editing by improving prompt-region correlation. Notably, there were instances of failed edits, for example, the model's inability to substitute `hamster' with `llama' in the third column of the right figure. This failure stems from the model's unfamiliarity with the word `llama', not from an incorrect correlation of regions.
        }
    \label{fig:overall-editting}
\end{figure}

\begin{itemize}
    \item Table~\ref{tab:user_study} demonstrates that our fully regularized version consistently outperforms all baseline methods across all metrics.
    \item The table further indicates that our method excels in tasks with tighter regional constraints, showing improved performance moving from broader image-level evaluations (text alignment) to more localized assessments (semantic correspondence). This underscores the effectiveness of our proposed method and regularization approach in enhancing text-semantic accuracy. 
    \item The visual comparisons in \Figref{fig:overall-editting} reinforce these quantitative findings.
\end{itemize}

\newpage
\subsection{Evaluating real image dataset}
\label{sec:real_images}

\begin{figure}[H]
    \centering
    \includegraphics[width=\linewidth]{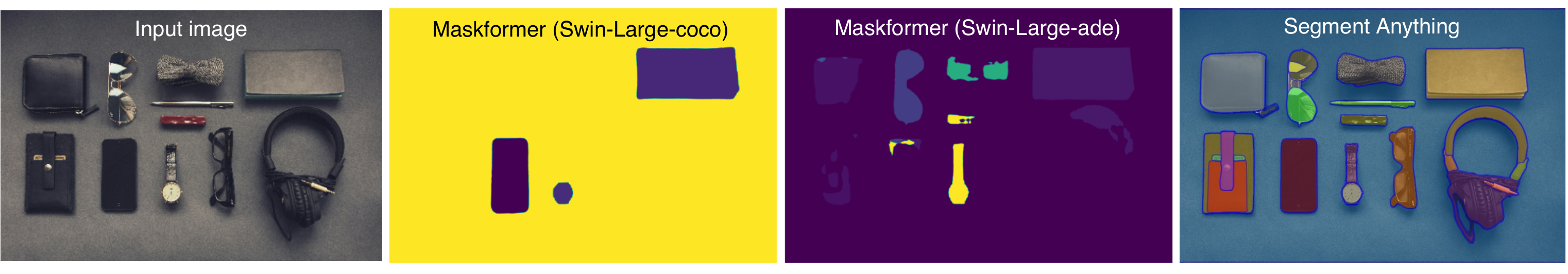}
    \caption{Challenge segmentation examples. 
    To streamline segmentation tasks, we utilize the MaskFormer model \citep{cheng2021per} trained on the COCO \citep{lin2014microsoft} and ADE20K \citep{zhou2017scene} datasets, along with the Segment Anything (SAM) model \citep{ma2023segment}. We observe that MaskFormer occasionally under-segments, whereas SAM is prone to over-segmentation. Consequently, manual adjustments are necessary to ensure the datasets are accurately prepared for evaluation.
    }
    \label{fig:difficult-segment}
\end{figure}

\paragraph{Real image collection and segmentation}
\begin{itemize}
    \item To evaluate our method with real images, we sourced a collection from Unsplash, following the precedent set by previous works like Custom Diffusion \citep{kumari2023multi} and Cones \citep{liu2023cones}. We utilized both automated tools (MaskFormer \citep{cheng2021per} and Segment Anything \citep{ma2023segment}) and manual segmentation for cases where automated methods were inadequate (refer to \Figref{fig:difficult-segment} for examples of challenging segmentation), resulting in a dataset of 137 unmasked and masked images across 50 classes.
    \item This dataset of real images, combined with our previously compiled natural and real medical multi-concept datasets, totals 1447 images, encompassing both masked objects and concepts. We plan to release this comprehensive dataset \textcolor{red}{\href{https://anonymous.4open.science/r/MCPL-rebuttal-630C}{here}}, addressing the need for more extensive evaluation datasets in multi-concept learning.
\end{itemize}

\begin{figure}[H]
    \centering
    \includegraphics[width=\linewidth]{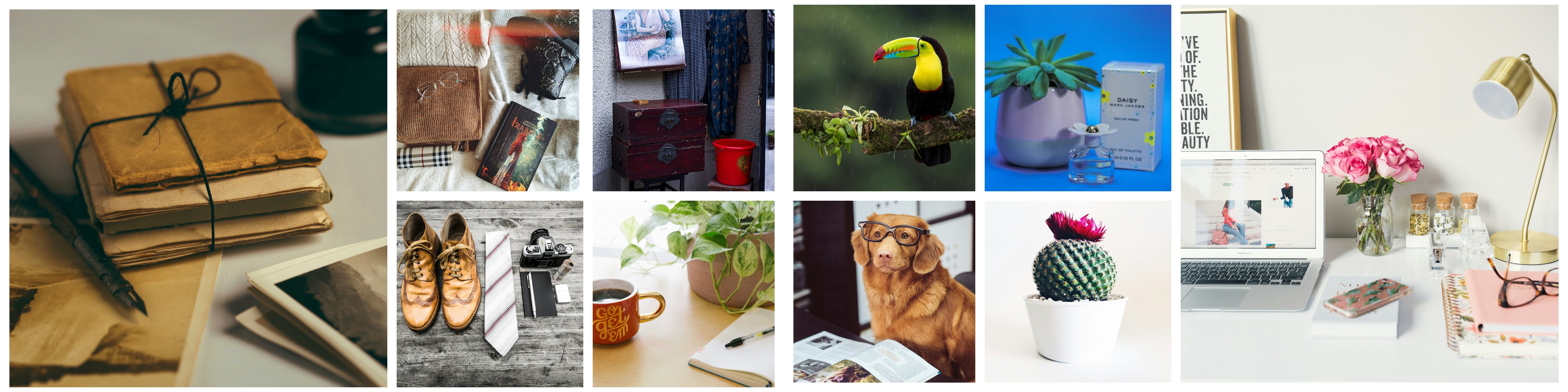}
    \caption{Visual examples of real images containing various number of objects.}
    \label{fig:real-highlight}
\end{figure}

\begin{figure}[H]
    \centering
    \begin{minipage}{0.495\linewidth}
        \includegraphics[width=\linewidth]{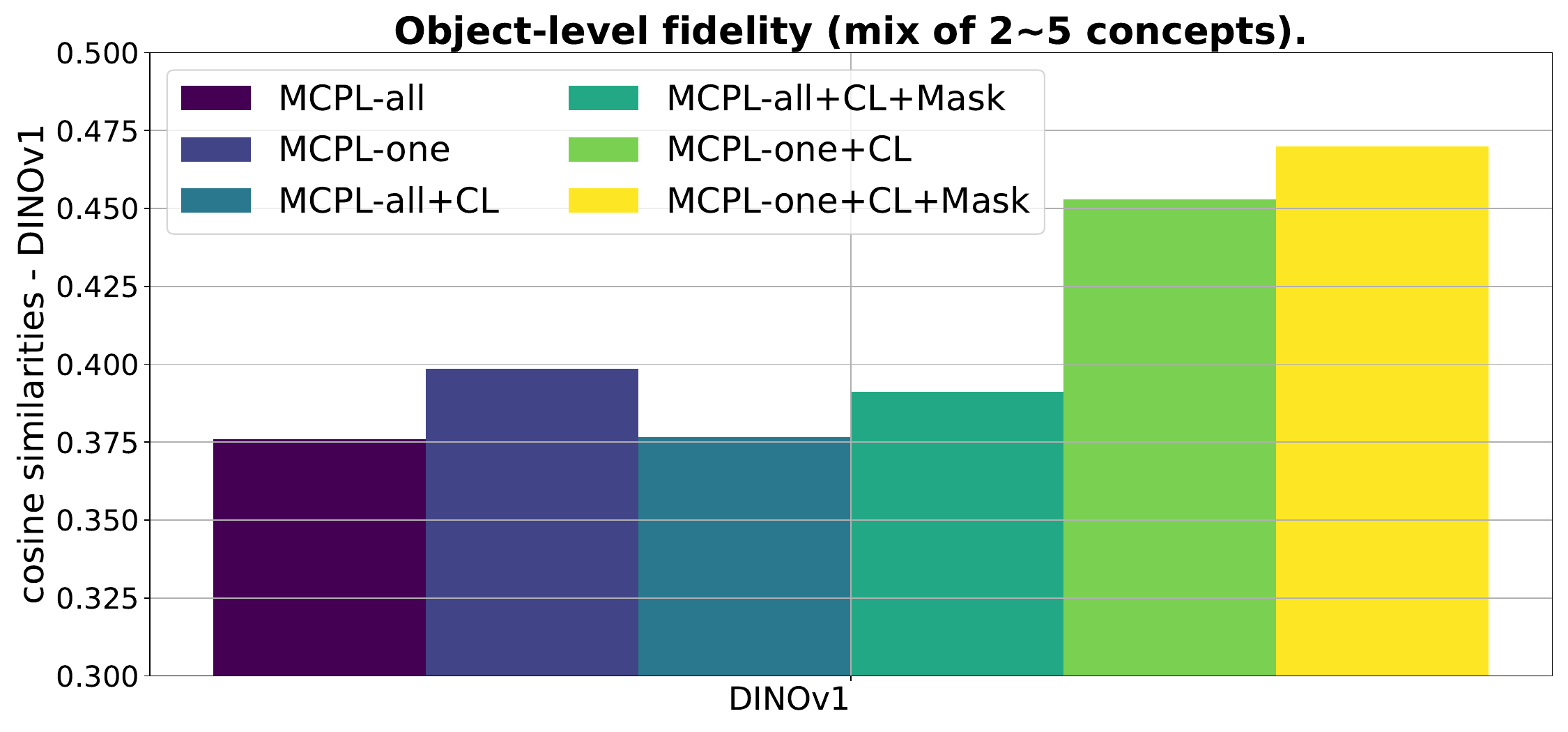}
        \footnotesize{Embedding similarity in learned object-level concepts compared to masked “ground truth” (real image)}
    \end{minipage}
    \hfill 
    \begin{minipage}{0.495\linewidth}
        \includegraphics[width=\linewidth]{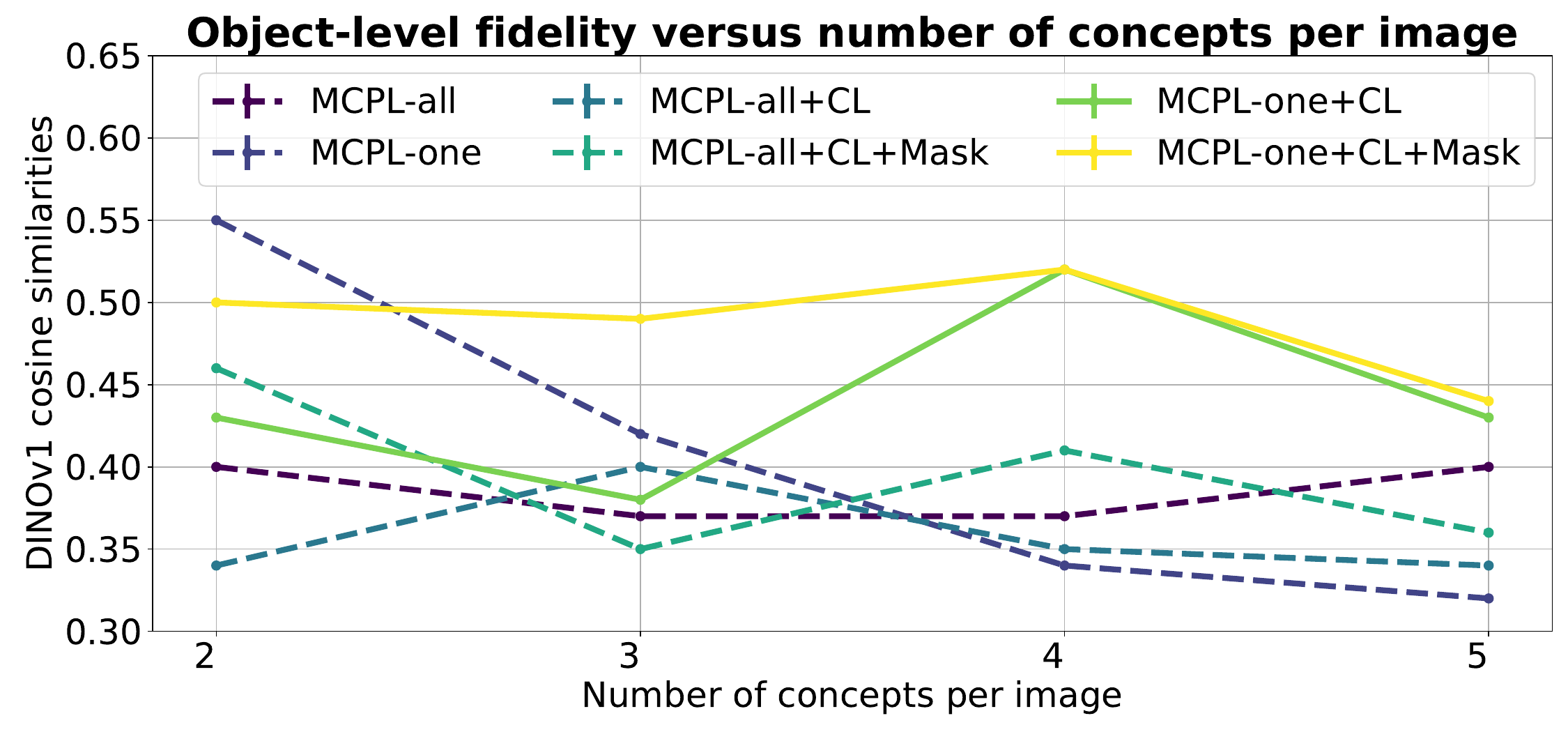}
        \footnotesize{Evaluate the learning as the number of concepts per image increases (real image).}
    \end{minipage}
    \caption{
        Quantitative results on real images 2-5 concepts per image and a total of 50 concepts, data point represents the pairwise cosine similarities measured by DINOv1. Also see the full object-level result in \Figref{fig:emb_similarity_real_345}}
    \label{fig:real-quantitative}
\end{figure}

\begin{figure*}
    \centering
    \includegraphics[width=.8\textwidth]{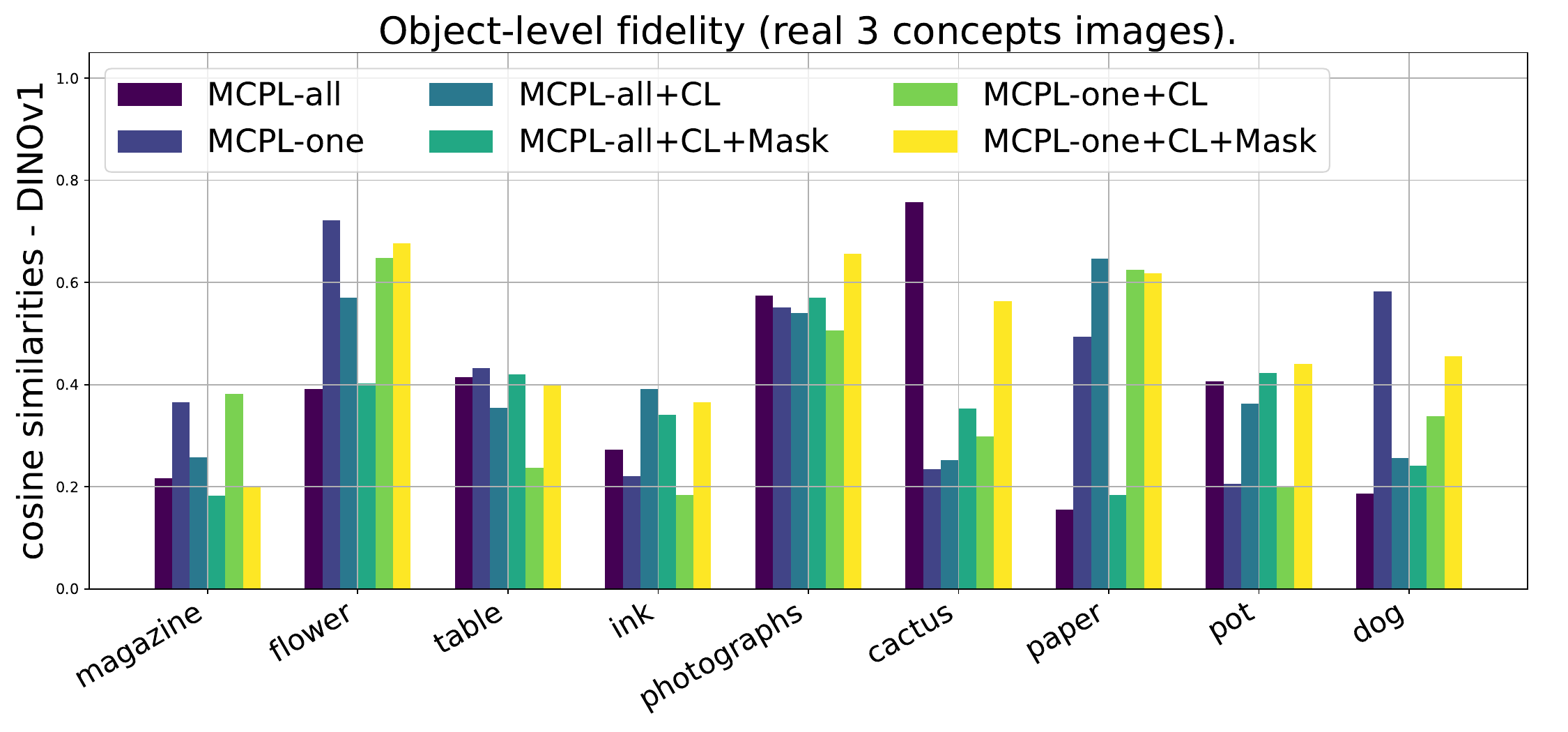}
    \includegraphics[width=.8\textwidth]{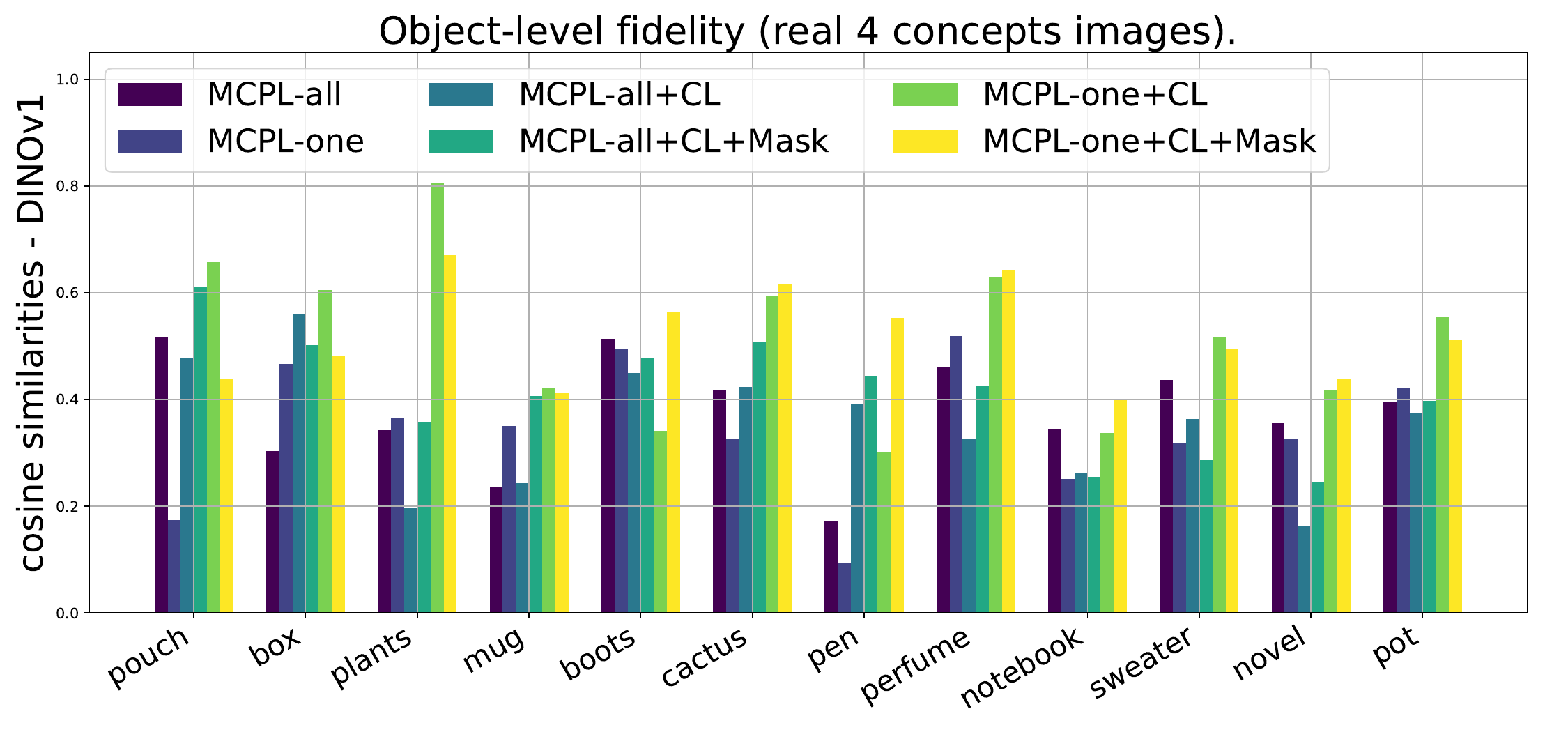}
    \includegraphics[width=.8\textwidth]{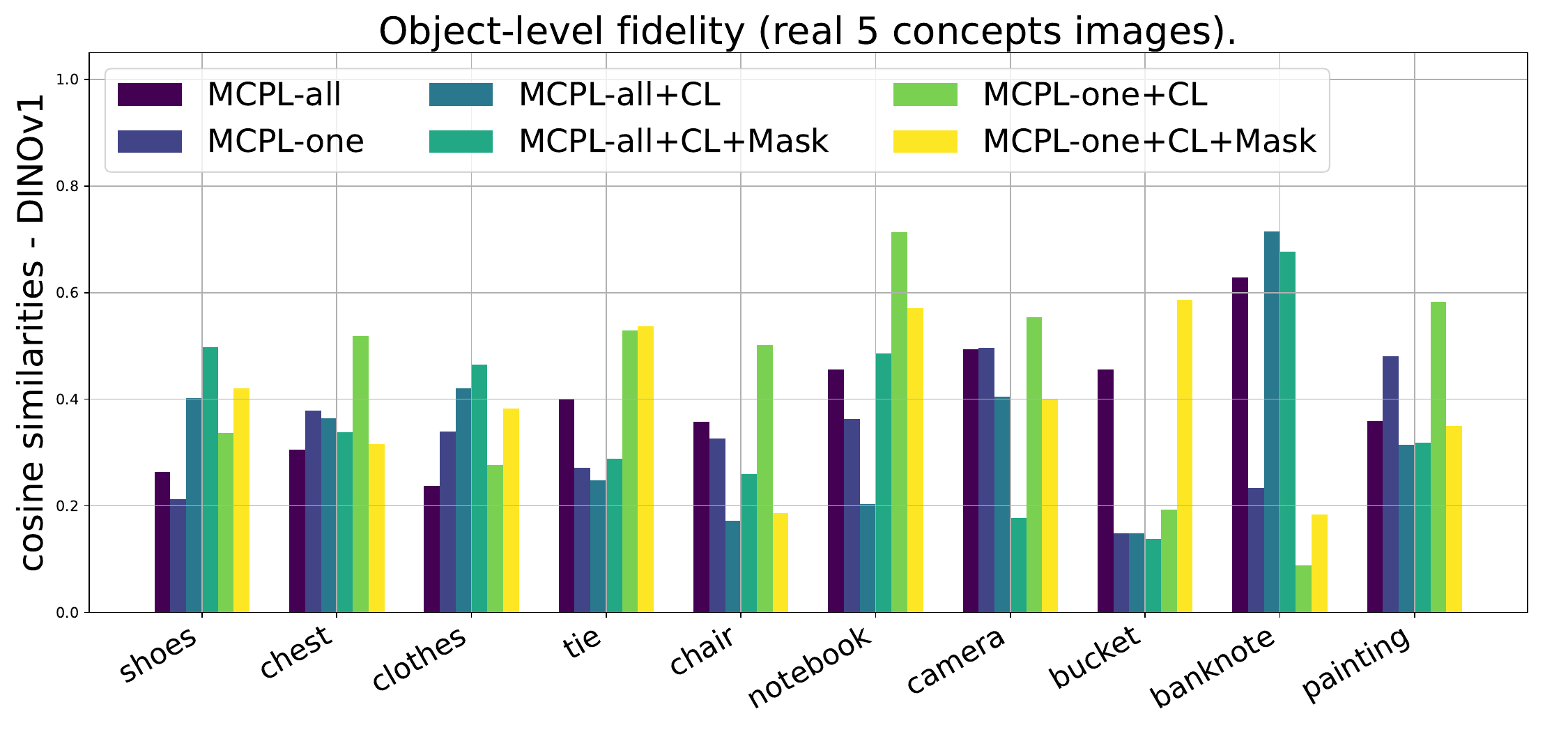}
    \caption{\footnotesize Real images (each containing 3,4 or 5 objects) per-object embedding similarity between the learned concept relative to the masked ``ground truth". Each plot computes image embeddings in the embedding spaces of DINOv1. We compare our base version adding variations of our proposed regularisation terms. \textbf{We observe our method consistently improving the accuracy of prompt-region correlation at the object level.}}
    \label{fig:emb_similarity_real_345}
\end{figure*}

\paragraph{Quantitative and qualitative evaluation.} We executed the complete evaluation process, as done with the generated natural image dataset, and observed consistent outcomes, where our method notably enhances text-semantic correlation. \textbf{Nevertheless, due to constrained time and resources, all learning was conducted using a single image. This limitation likely impacts learning efficacy, as previously identified in research} \citep{gal2022textual, vinker2023concept}.
\begin{itemize}
    \item We conducted learning exercises with MCPL and various baseline methods, calculating the embedding similarity of learned object-level concepts against the masked "ground truth" as indicated by DINOv1 (refer to the left figure in \Figref{fig:real-quantitative} for an overview and \Figref{fig:emb_similarity_real_345} for detailed object-level outcomes). 
    \item Additionally, we categorized the dataset based on the number of concepts per image to assess learning performance as the number of concepts per image grows. This analysis, depicted in the right figure in \Figref{fig:real-quantitative}, revealed a consistent challenge: as the number of concepts increases, the difficulty mirrors that observed in previous imagery annotation-based learning methods such as Cones \citep{liu2023cones} and Break-a-Scene \citep{avrahami2023break}.
\end{itemize}

\begin{figure}[ht]
    \centering
    \begin{minipage}[c]{0.12\linewidth}
        \centering
        \includegraphics[width=\linewidth]{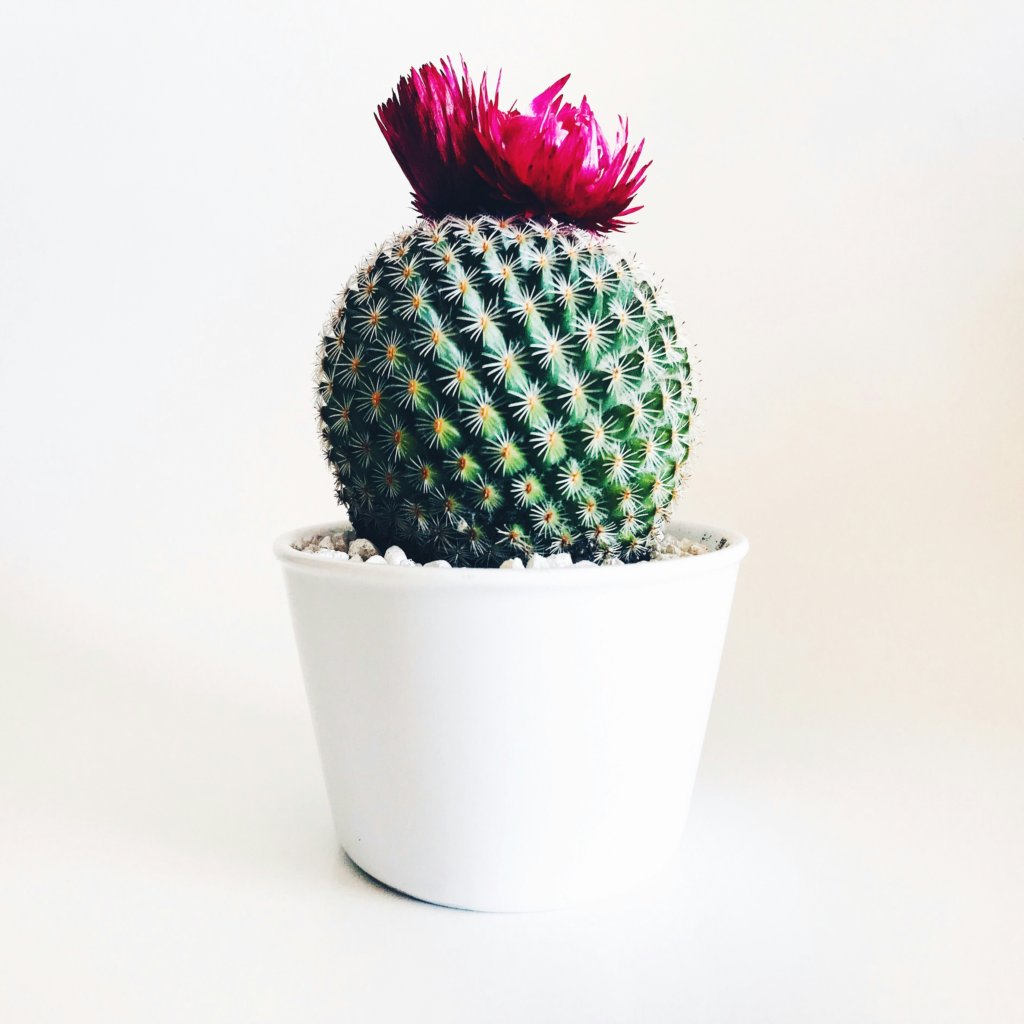}
    \end{minipage}
    \hfill
    \begin{minipage}[c]{0.87\linewidth}
        \centering
        \includegraphics[width=\linewidth]{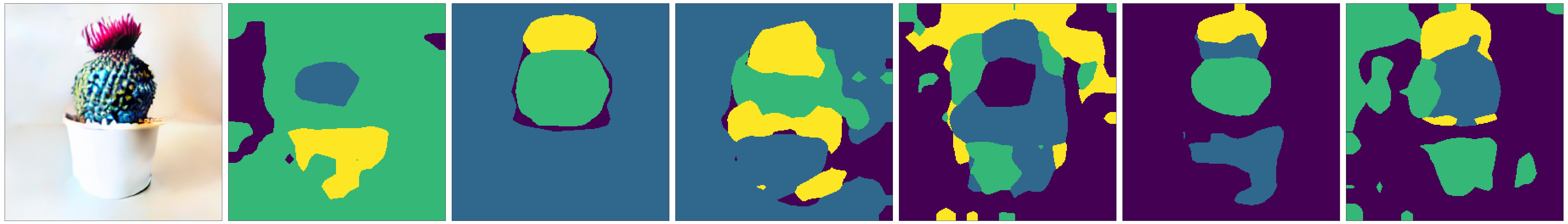}
    \end{minipage}
    \textit{``a green \textcolor{blue}{cactus/ token1} settled in a white \textcolor{teal}{pot/ token2} with red \textcolor{olive}{flower/ token3} on it''}
    
    \begin{minipage}[c]{0.12\linewidth}
        \centering
        \includegraphics[width=\linewidth]{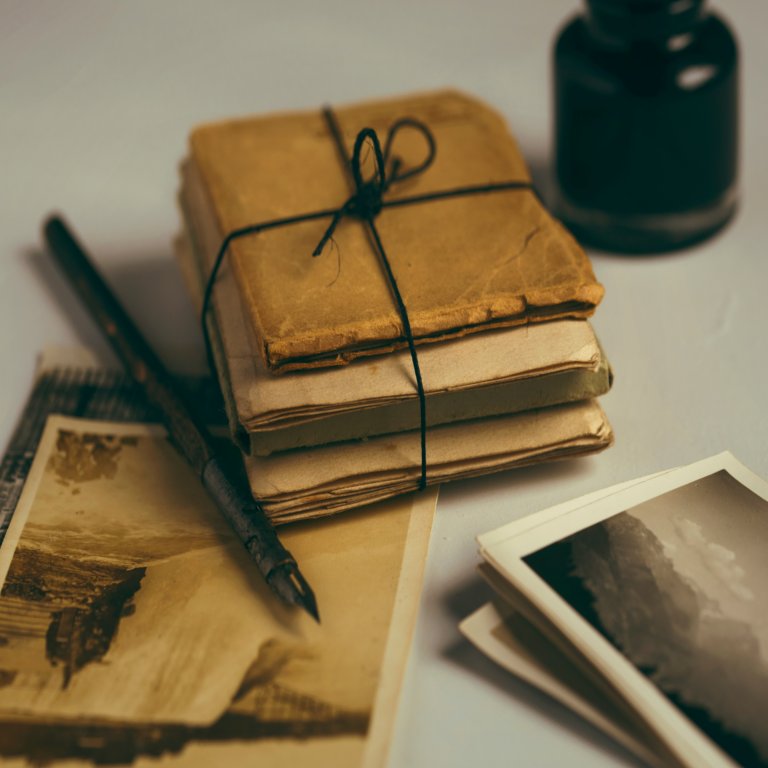}
    \end{minipage}
    \hfill
    \begin{minipage}[c]{0.87\linewidth}
        \centering
        \includegraphics[width=\linewidth]{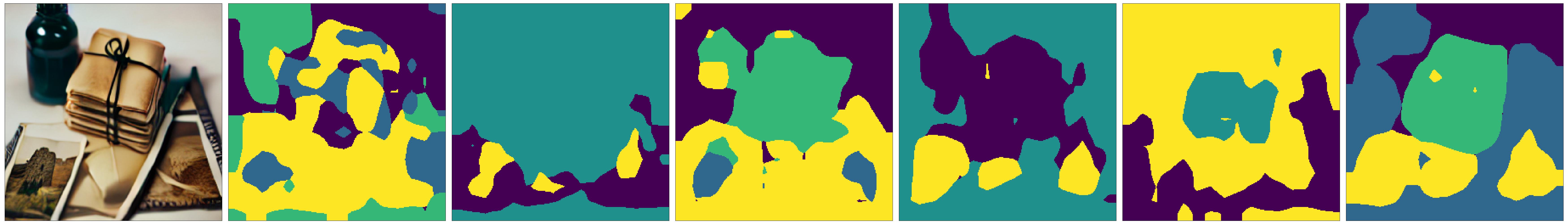}
    \end{minipage}
    \textit{``a black \textcolor{blue}{ink/ token1} a brown \textcolor{teal}{paper/ token2} and a stack \textcolor{olive}{photographs/ token3}''}
    
    \begin{minipage}[c]{0.12\linewidth}
        \centering
        \includegraphics[width=\linewidth]{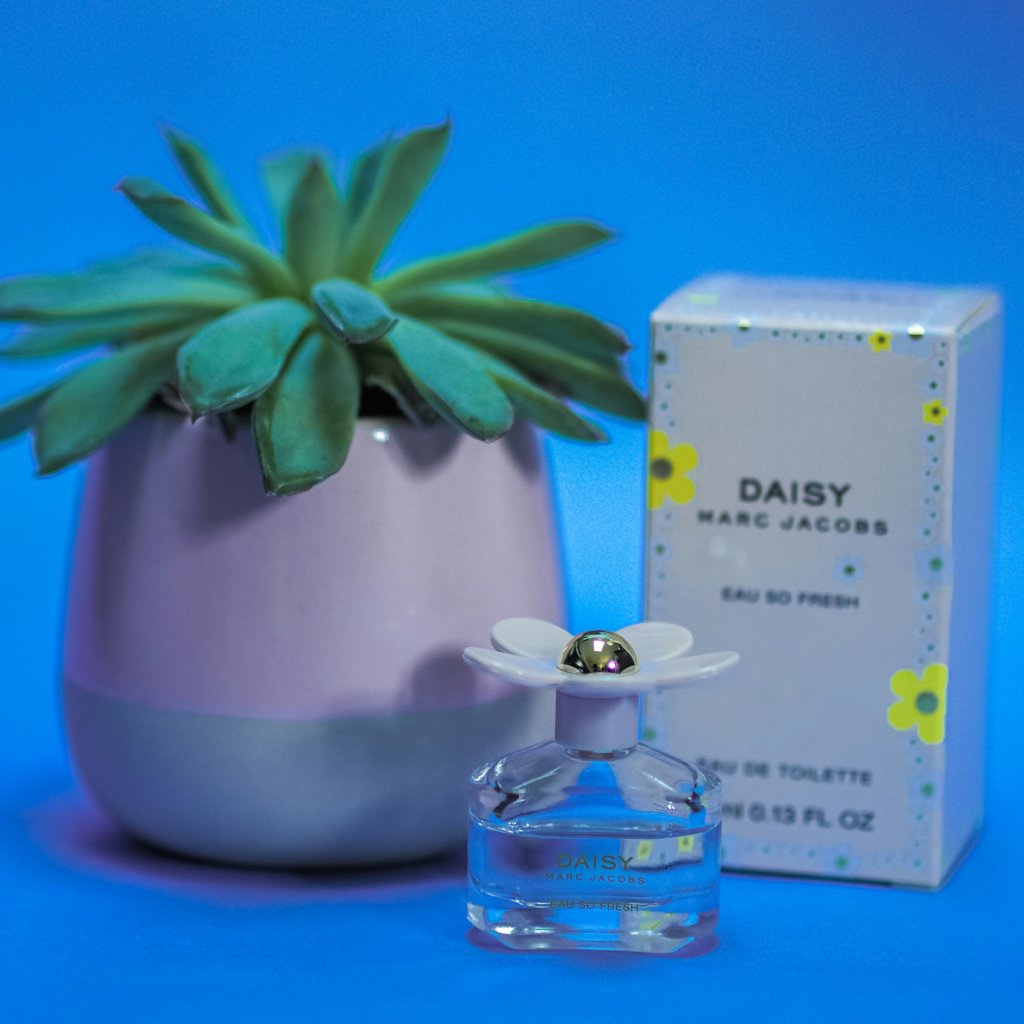}
    \end{minipage}
    \hfill
    \begin{minipage}[c]{0.87\linewidth}
        \centering 
        \includegraphics[width=\linewidth] {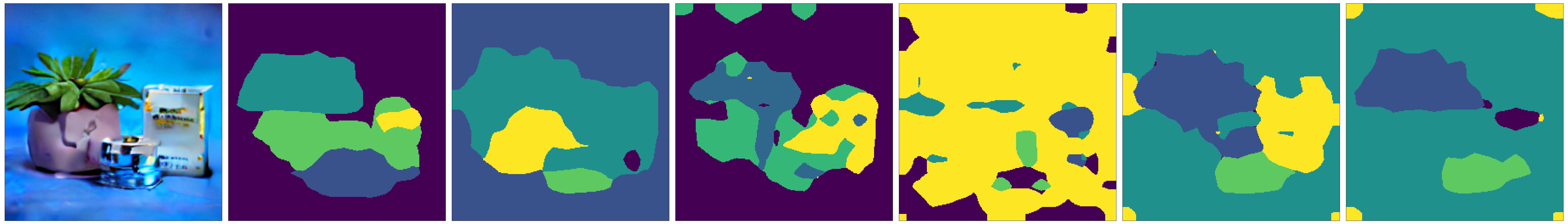}
    \end{minipage}
    \textit{``a green \textcolor{blue}{cactus/ token1} a pink \textcolor{teal}{pot/ token2} glass \textcolor{green}{perfume/ token3} and white \textcolor{olive}{box/ token4}''}
    
    \begin{minipage}[c]{0.12\linewidth}
        \centering
        \includegraphics[width=\linewidth]{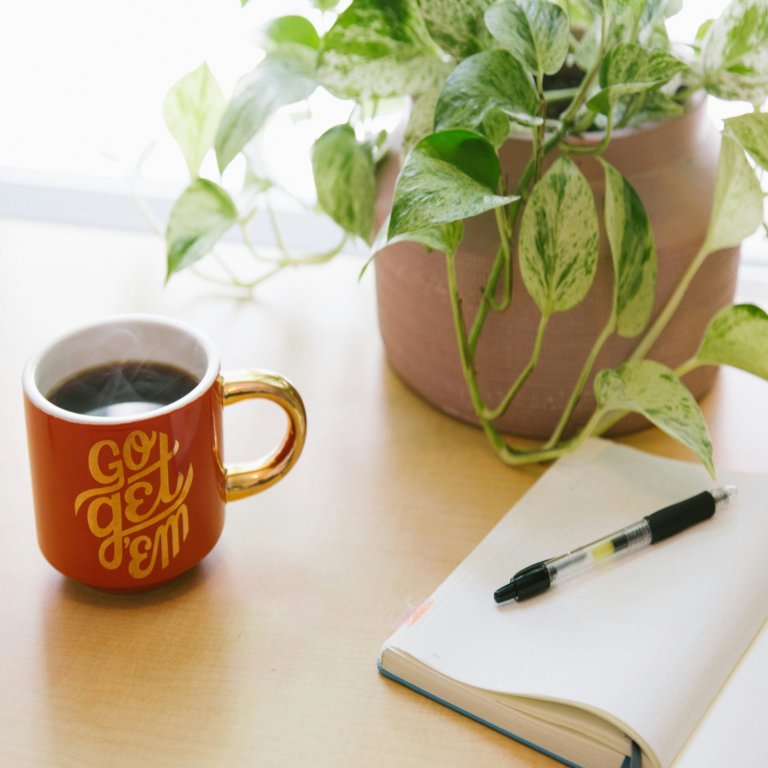}
    \end{minipage}
    \hfill
    \begin{minipage}[c]{0.87\linewidth}
        \centering
        \includegraphics[width=\linewidth]{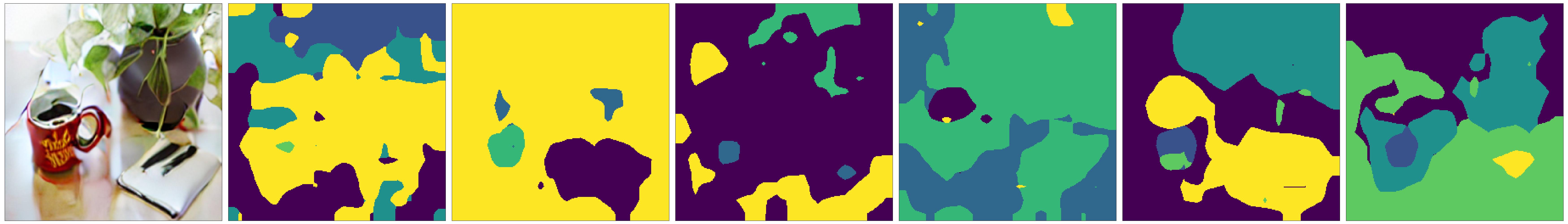}
    \end{minipage}
    \textit{``a red \textcolor{blue}{mug/ token1} green \textcolor{teal}{plants/ token2} blank \textcolor{green}{notebook/ token3} and a black \textcolor{olive}{pen/ token4}''}
    
    \caption{Visual results on real images. From left to right: 
        1) real input image, 
        2) generated image, 
        3) MCPL-all, 
        4) MCPL-one, 
        5) MCPL-all+\textit{PromptCL}+\textit{Bind adj.}, 
        6) MCPL-all+\textit{AttnMask}+\textit{PromptCL}+\textit{Bind adj.}, 
        7) MCPL-one+\textit{PromptCL}+\textit{Bind adj.}, 
        8) MCPL-one+\textit{AttnMask}+\textit{PromptCL}+\textit{Bind adj.}.
        \textbf{The final two columns showcase our optimised versions, consistently achieving higher accuracy in semantic correlation compared to the baseline methods.}
        }
    \label{fig:visual_real_simple}
\end{figure}

\begin{figure}[H]
    \centering
    \includegraphics[width=.9\linewidth]{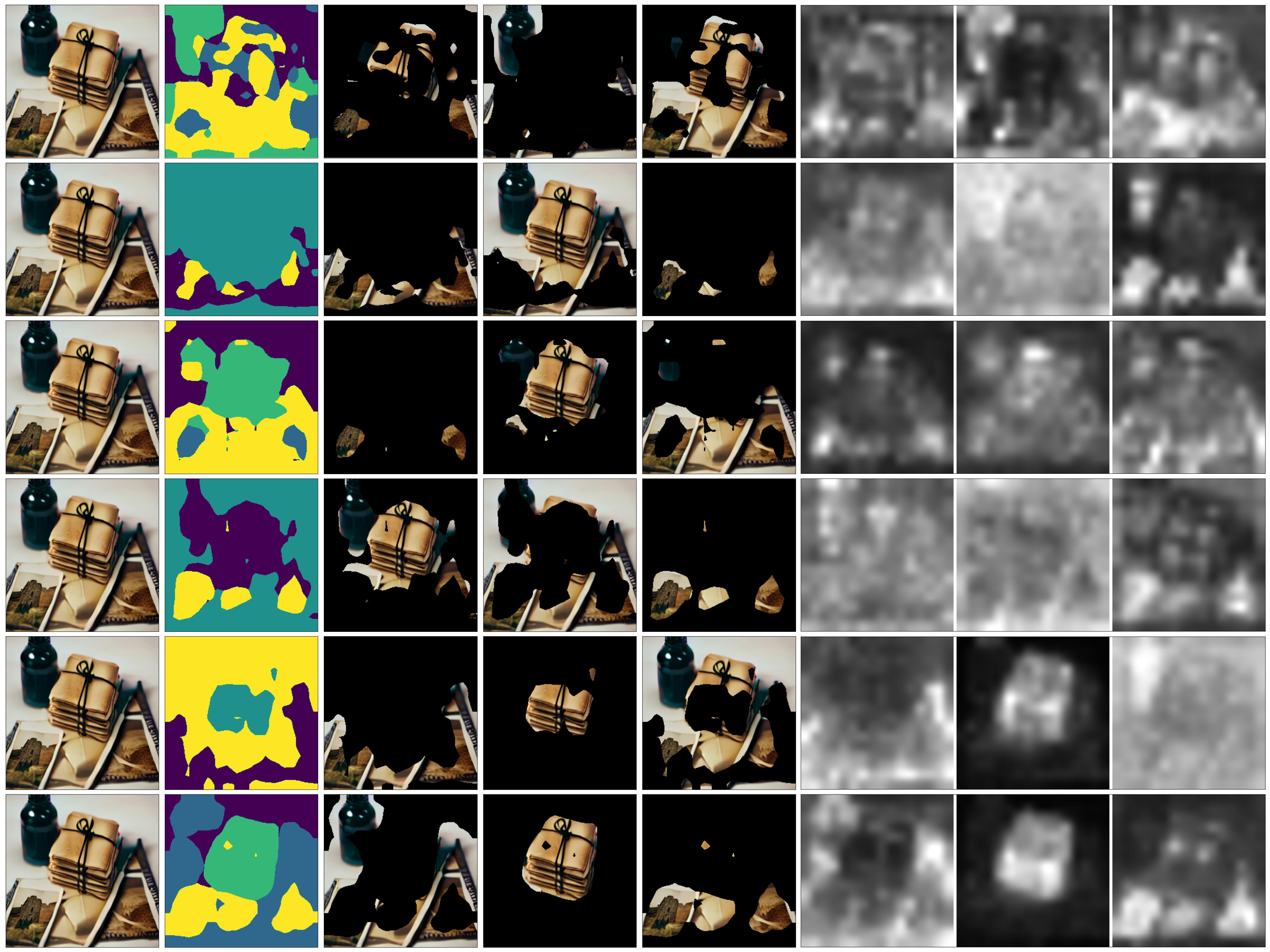}
    
    \textit{``a black \textcolor{blue}{ink/ token1} a brown \textcolor{teal}{paper/ token2} and a stack \textcolor{olive}{photographs/ token3}''}
    
    \includegraphics[width=.9\linewidth]{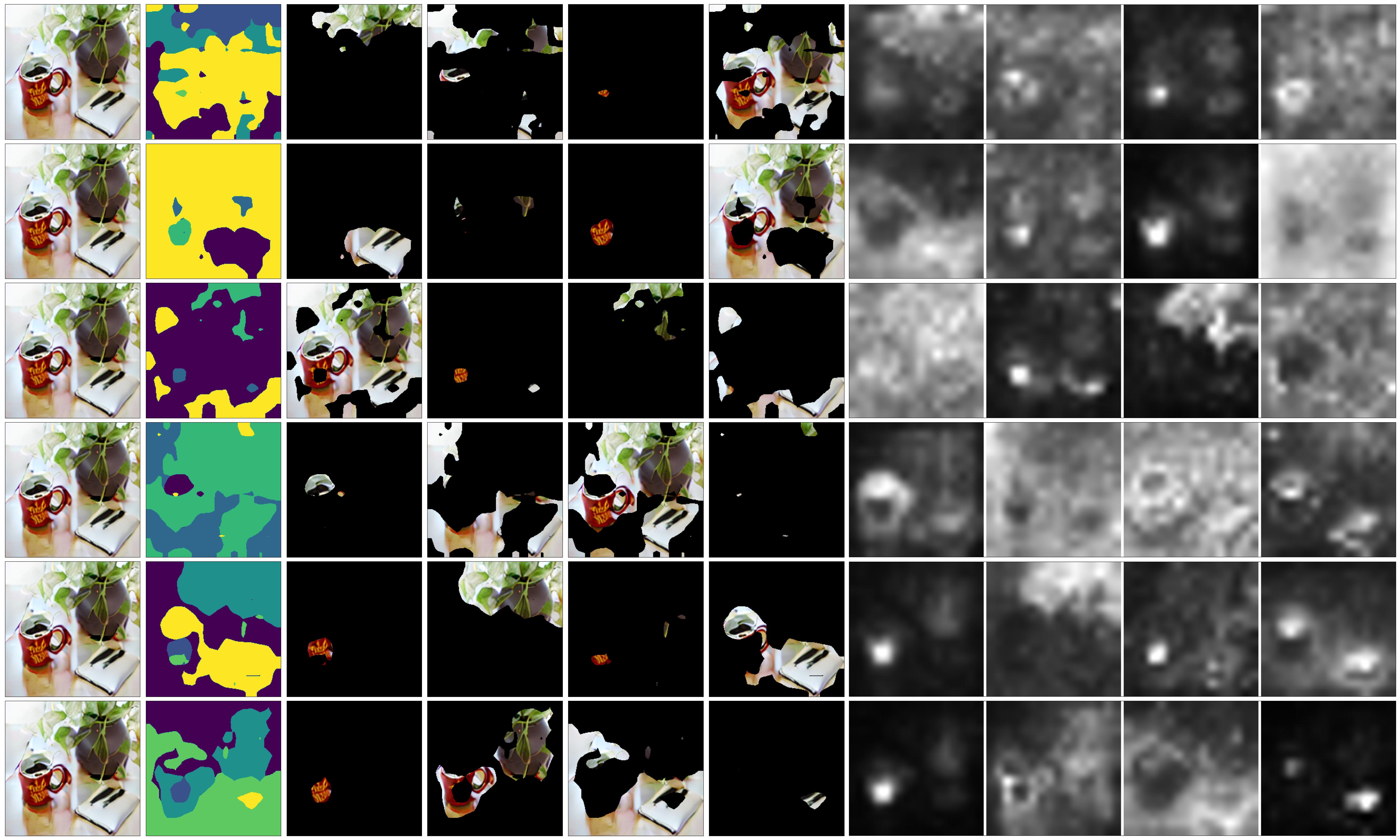}

    \textit{``a red \textcolor{blue}{mug/ token1} green \textcolor{teal}{plants/ token2} blank \textcolor{green}{notebook/ token3} and a black \textcolor{olive}{pen/ token4}''}
    \caption{
        Visualisation with mask and attention. 
        We compare all baseline methods across each row (from top to bottom): 
        1) MCPL-all, 
        2) MCPL-one, 
        3) MCPL-all+\textit{PromptCL}+\textit{Bind adj.}, 
        4) MCPL-all+\textit{AttnMask}+\textit{PromptCL}+\textit{Bind adj.}, 
        5) MCPL-one+\textit{PromptCL}+\textit{Bind adj.}, 
        6) MCPL-one+\textit{AttnMask}+\textit{PromptCL}+\textit{Bind adj.}.
        }
    \label{fig:visual_real_attn_ink}
\end{figure}

\begin{itemize}
    \item Visual outcomes presented in \Figref{fig:visual_real_simple} and \Figref{fig:visual_real_attn_ink} 1) validate our quantitative findings our fully regularized version consistently achieving higher accuracy in semantic correlation compared to the baseline methods and 2) highlight the increasing challenge as scenes become more complex.
    \item It's important to recognise that the observed performance decline is mainly attributed to 1) all learnings are performed with a single image due to constrained time and resources and 2) our method is implemented on the relatively less powerful LDM \citep{rombach2022high} model, operating at a resolution of $256 \times 256$ to enhance computational efficiency. 
\end{itemize}

\subsection{Comparing concepts discovery with Inspiration Tree}
\label{sec:compare_inspiration_tree}

\begin{figure}[H]
    \centering
    \includegraphics[width=.8\linewidth]{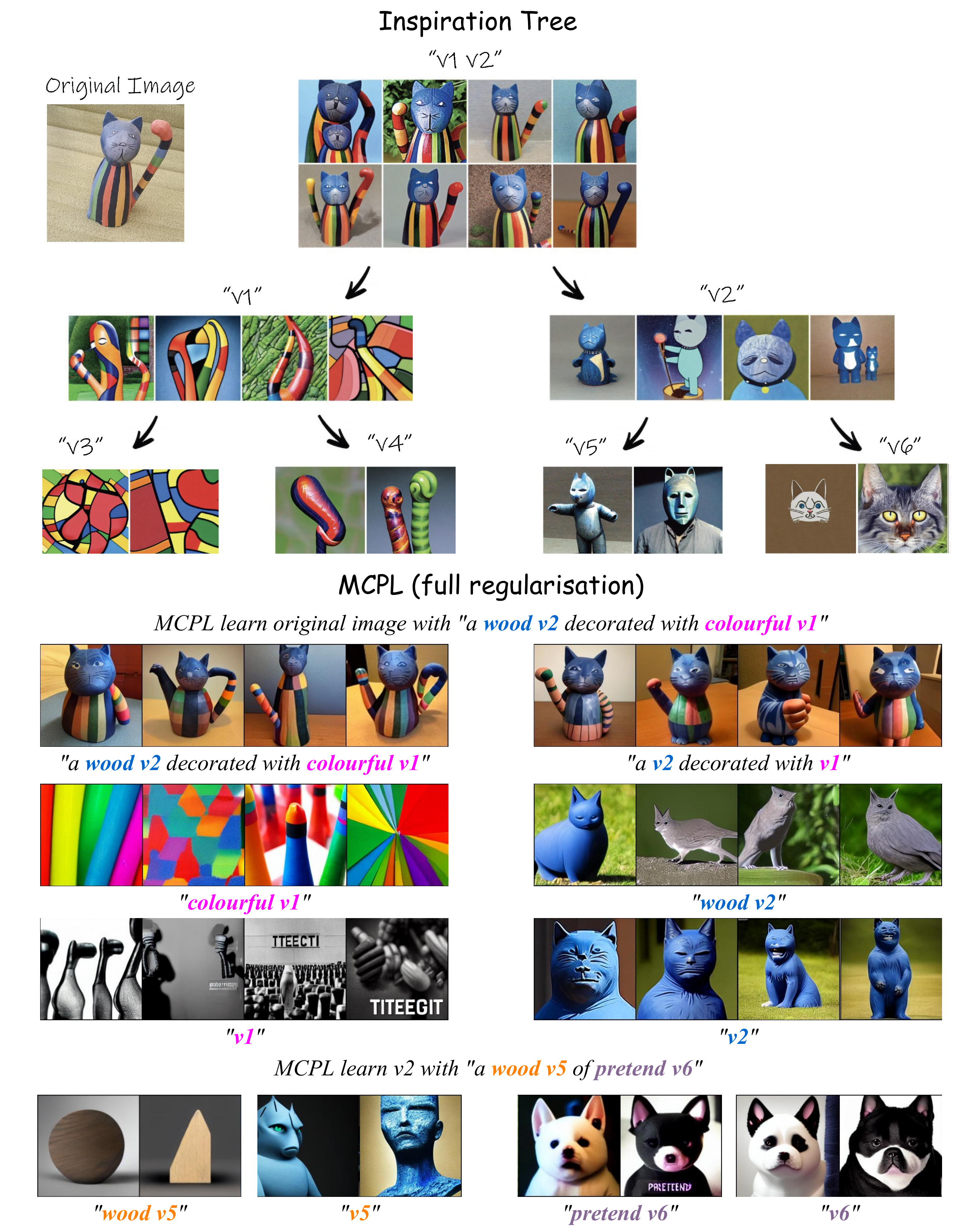}
    \vspace{-4mm}
    \caption{Comparing MCPL with Inspiration Tree \citep{vinker2023concept} in \textbf{exploring concepts from image}. We observe: 
    1) \textbf{Although MCPL was not specifically designed for abstract concept learning, it demonstrates decent performance in comparison to IT, even in its deeper layers.}
    2) Adjectives assume a different role from their function in previous tasks focused on semantic differentiation. Typically, an adjective defines an abstract 'style,' like 'colourful'. Yet, in certain instances, the adjective can overpower the underlying concept of interest, as seen in comparisons like `wood v5' versus `wood v2'. 
    }
    \label{fig:mcpl_inspiretree_explore}
\end{figure}

Inspiration Tree (IT) \citep{vinker2023concept} shares a similar spirit as our work in discovering multiple concepts from images, diverging from the language-guided discovery characteristic of MCPL to allow for unstructured concept exploration. 
MCPL excels in instruction following discovery, adept at identifying concepts distinct in both visual and linguistic aspects, whereas IT specializes in unveiling abstract and subtle concepts independently of direct human guidance. 
\textbf{We view MCPL and IT as complementary, enhancing different aspects of concept identification across diverse fields like arts, science, and healthcare.} 
In this section, we compare MCPL with IT in both concepts exploration \Figref{fig:mcpl_inspiretree_explore} and combination \Figref{fig:mcpl_inspiretree_combine}.

\vspace{-2mm}

\begin{figure}[H]
    \centering
    \includegraphics[width=.8\linewidth]{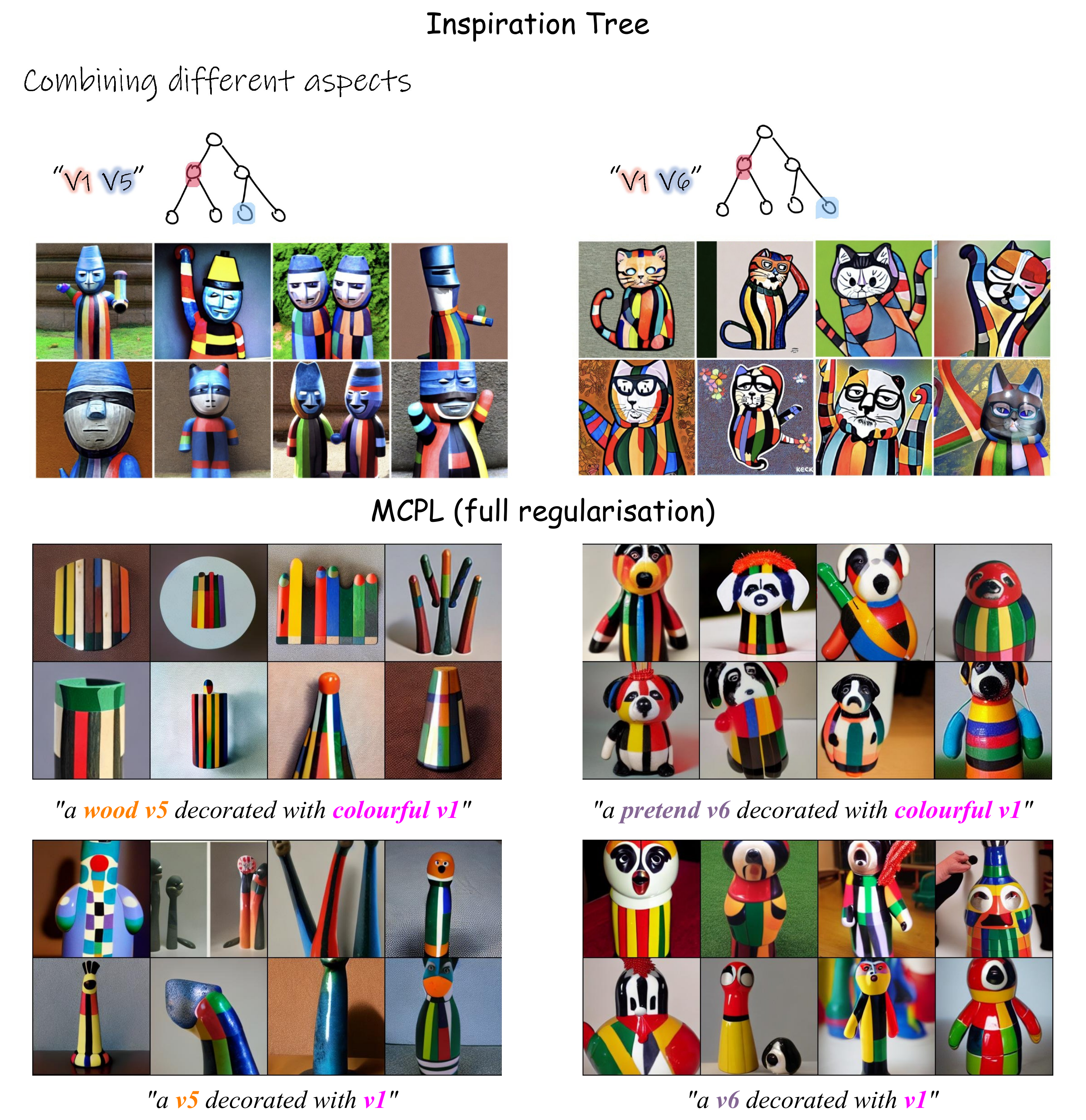}
    \caption{Comparing MCPL with Inspiration Tree \citep{vinker2023concept} in \textbf{combining separately learnt concepts into one image}. We observe: 
    1) MCPL showcases notable effectiveness in comparison to IT, despite its initial focus on semantic regions. It adeptly discerns underlying abstract styles and integrates them into new scenes.
    2) Adjectives exhibit varied influences: terms like `colourful' and `pretend' effectively direct `v1' and `v6' towards accurately learning and reproducing the intended concept, even after removing `colourful' in the combination process. Conversely, `wood' dominates the intended concept, misguiding `v5'.
    }
    \label{fig:mcpl_inspiretree_combine}
\end{figure}

\begin{figure}[H]
    \centering
    \includegraphics[width=.8\linewidth]{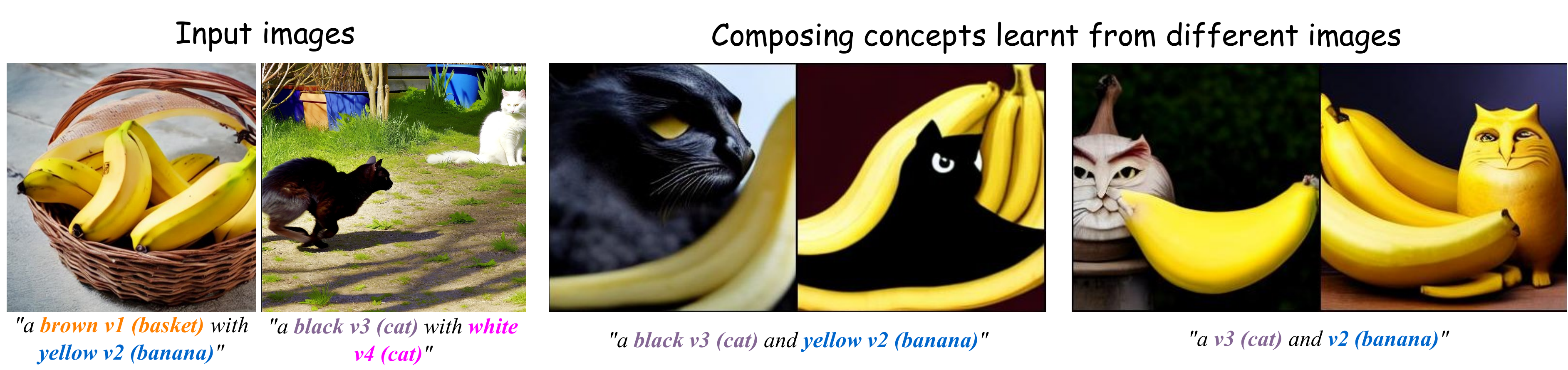}
    \caption{Combination of objects from different images. 
    We observe: 
    1) MCPL demonstrates a good ability to combine concepts without updating the diffusion model parameters \textbf{but only update text embedding}.
    2) Yet, the performance of this combination approach has limitations. As \textbf{MCPL is aimed at identifying semantic regions within image-text pairs}, for more sophisticated image editing, it is advisable to use \textbf{MCPL as a language-driven mask proposer}. The identified masked objects can then serve as inputs for mask-based techniques like Break-a-Scene \citep{avrahami2023break}, which updates the diffusion model parameters for enhanced recomposition.
    }
    \label{fig:mcpl_inspiretree_combine}
\end{figure}

\newpage
\subsection{Stress test with challenge images}
\label{sec: challenge_cases}

\begin{figure}[H]
    \centering
    \includegraphics[width=\linewidth]{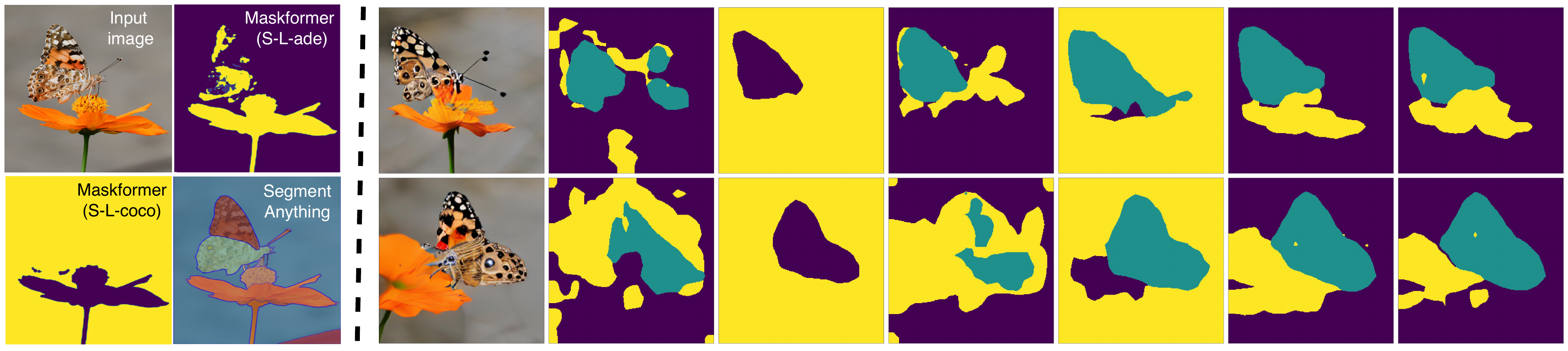}
    \textit{``a patterned \textcolor{teal}{butterfly/ token1} resting on an orange \textcolor{olive}{flower/ token2}''}
    \caption{
    \textcolor{teal}{\textbf{Successful stress test}}
    in learning concepts with \textbf{a similar colour}.
    On the left, the input image and masks predicted by pre-trained segmentation models are displayed, \textbf{highlighting the difficulty in distinguishing textures between two concepts}. On the right, our results are presented, with the last two columns demonstrating our optimized approach using various learning strategies, which consistently outperforms baseline methods including segmentation models in terms of semantic accuracy.
    From left to right we have:
    1) MCPL generated image, 
    2) MCPL-all, 
    3) MCPL-one, 
    4) MCPL-all+\textit{PromptCL}+\textit{Bind adj.}, 
    5) MCPL-all+\textit{AttnMask}+\textit{PromptCL}+\textit{Bind adj.}, 
    6) MCPL-one+\textit{PromptCL}+\textit{Bind adj.}, 
    7) MCPL-one+\textit{AttnMask}+\textit{PromptCL}+\textit{Bind adj.}.
    }
    \label{fig:challenge_success_butterfly}
\end{figure}

\begin{figure}[H]
    \centering
    \includegraphics[width=\linewidth]{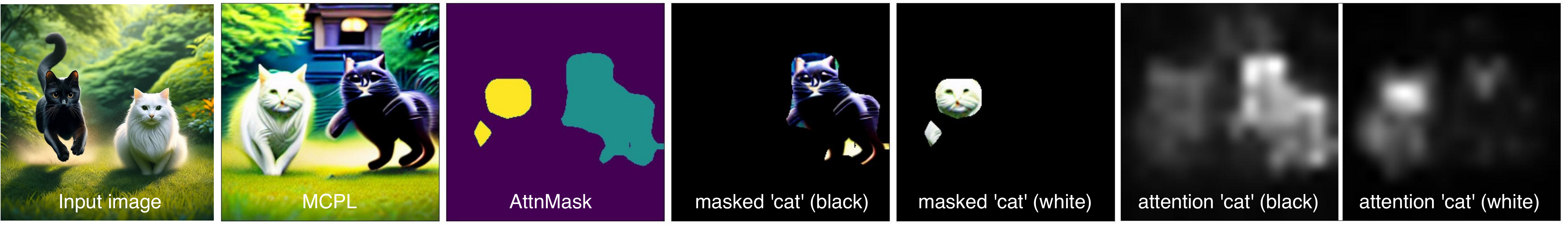}
    \textit{``a running black \textcolor{teal}{cat/ token1} and a sitting white \textcolor{olive}{cat/ token2}''}
    \caption{
    \textcolor{teal}{\textbf{Successful stress test}}
    in learning concepts with \textbf{the same category} --- two cats.
    From left to right, we present the input image, the MCPL-generated image with full regularisation, the semantic masks, masked concepts, and per-prompt cross-attention. \textbf{We demonstrate MCPL's capability to identify the same category when the semantic appearances are distinct in both visual and linguistic aspects}. 
    }
    \label{fig:challenge_success_two_cats}
\end{figure}

\begin{figure}[H]
    \centering
    \includegraphics[width=\linewidth]{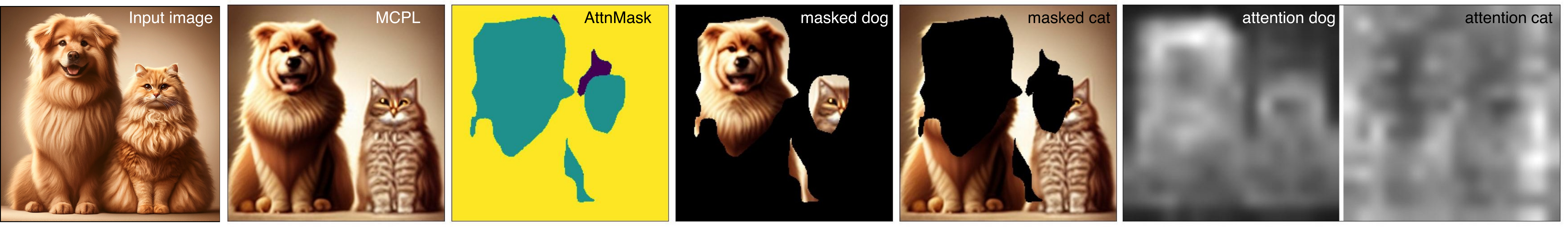}
    \textit{``a smiling \textcolor{teal}{dog/ token1} and an elegant \textcolor{olive}{cat/ token2}''}
    \caption{
    \textcolor{orange}{\textbf{Mixed stress test}}
    in learning concepts with \textbf{a similar appearance (i.e., shared adjectives)} --- a dog and a cat with a similar colour.
    From left to right, we present the input image and our fully regularised result. 
    In this scenario, MCPL demonstrates varied performance, distinguishing the animal's face from its body instead of by category. This issue stems from the indistinct appearance, particularly linguistically. \textbf{Since adjectives critically guide the model's focus, learning falters when the adjective does not effectively differentiate regions}.
    }
    \label{fig:challenge_mix_flur_dog_cat}
\end{figure}

\begin{figure}[H]
    \centering
    \includegraphics[width=\linewidth]{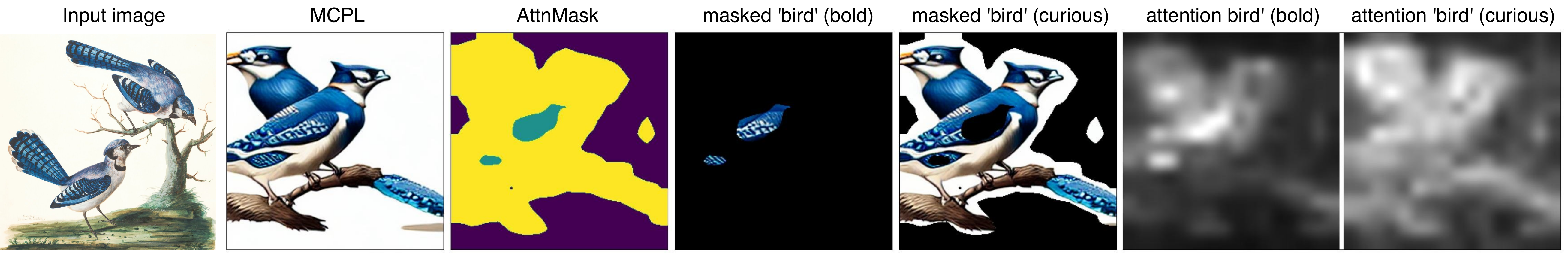}
    \textit{``a bold \textcolor{teal}{bird/ token1} on top of a curious \textcolor{olive}{bird/ token2}''}
    \caption{
    \textcolor{purple}{\textbf{Failed stress test}}
    in learning concepts with \textbf{visually similar concepts} --- a photo of two identical birds.
    From left to right, we present the input image and our fully regularised result. 
    In an extreme case where two birds are identical, MCPL fails to differentiate between them. \textbf{MCPL aims to accurately learn multiple concepts following language instructions. However, when concepts are linguistically indistinguishable, learning falters.} In these instances, we recommend adopting learning methods with imagery annotation such as Cones \citep{liu2023cones}, Breas-a-Scene \citep{avrahami2023break} or less reliant on human input, like the Inspiration Tree \citep{vinker2023concept}, as discussed in Section \ref{sec:compare_inspiration_tree}.
    }
    \label{fig:challenge_failed_two_birds}
\end{figure}

\begin{figure}[H]
    \centering
    \includegraphics[width=\linewidth]{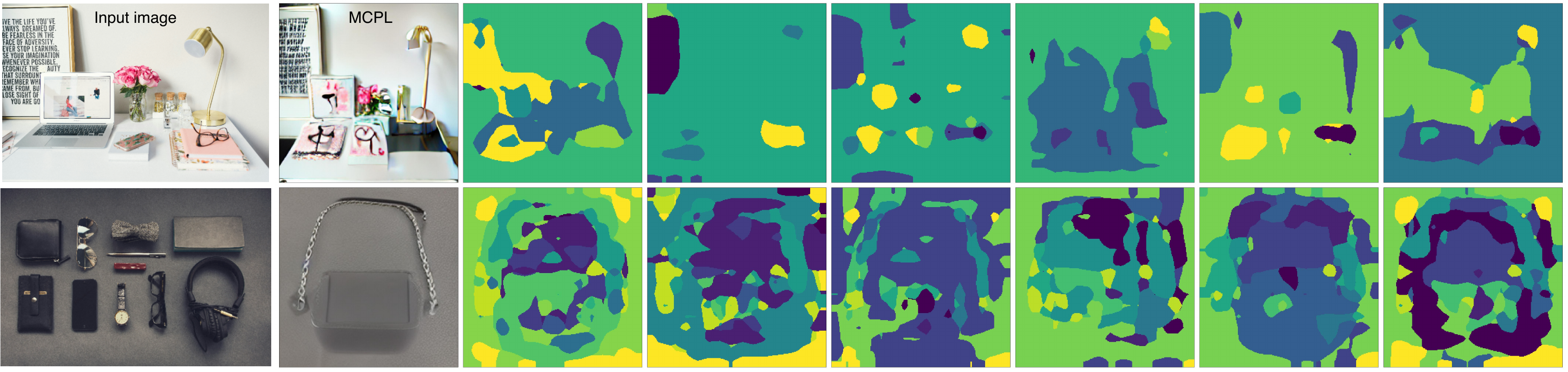}
    \caption{
    \textcolor{purple}{\textbf{Failed stress test}}
    in learning concepts with \textbf{a large number of concepts from a single image}.
    MCPL struggles to learn a large number of concepts from a single image, a limitation also noted in previous image annotation-based multi-concept learning methods like Cones \citep{liu2023cones} and Break-a-Scene \citep{avrahami2023break}. This challenge is exacerbated by a few key factors:
    1) \textbf{Limited Example Images}: As highlighted by \cite{gal2022textual} and \cite{vinker2023concept}, learning is more efficient with 4 to 6 example images showcasing the same concepts. Yet, sourcing multiple images containing a broad array of concepts for learning is difficult.
    2) \textbf{Weak Backbone Model}: Our methodology employs the LDM \citep{rombach2022high} model, chosen for its computational efficiency despite being less powerful and operating at a resolution of $256 \times 256$. This limitation hampers the model's ability to reconstruct complex scenes (see the second example) leading to inaccurately associated concepts.
    \textbf{Alternative Solution}: Despite these challenges, we propose combining the MCPL-diverse strategy with the random crop technique introduced by Break-a-Scene \citep{avrahami2023break}. MCPL-diverse is a strategy to learn per-image specified concepts, while random cropping helps in subsampling concepts from a larger image leading to increased examples. This combination enhances the ability to learn numerous concepts from a single image, as shown in our six objects example in Figure \ref{fig:bas_compare}, where MCPL achieves results comparable to the mask-based BAS method \citep{avrahami2023break}.
    }
    \label{fig:challenge_failed_large_num_complex}
\end{figure}

\begin{itemize}
    \item To fully explore the boundary of our method, we conducted a set of challenging experiments including learning concepts with similar colour (\Figref{fig:challenge_success_butterfly}), same category (\Figref{fig:challenge_success_two_cats}), similar appearance (\Figref{fig:challenge_mix_flur_dog_cat}, \Figref{fig:challenge_failed_two_birds}) and learning a large number of concepts from a single image (\Figref{fig:challenge_failed_large_num_complex}).
\end{itemize}

\subsection{Limitations}
\label{sec: limitations}

MCPL enhances prompt-region correlation through natural language instructions but may encounter difficulties in scenarios such as:
\begin{itemize}
    \item When concepts are linguistically indistinct, for example, multiple identical or highly similar concepts that natural language struggles to differentiate.
    \item In highly complex scenes containing many concepts with limited example images available, a challenge recognized by \cite{liu2023cones} and \cite{avrahami2023break}.
\end{itemize}

\newpage

\newpage
\subsection{More editing examples}

\begin{figure}[ht]
    \centering
    \includegraphics[width=1\linewidth]{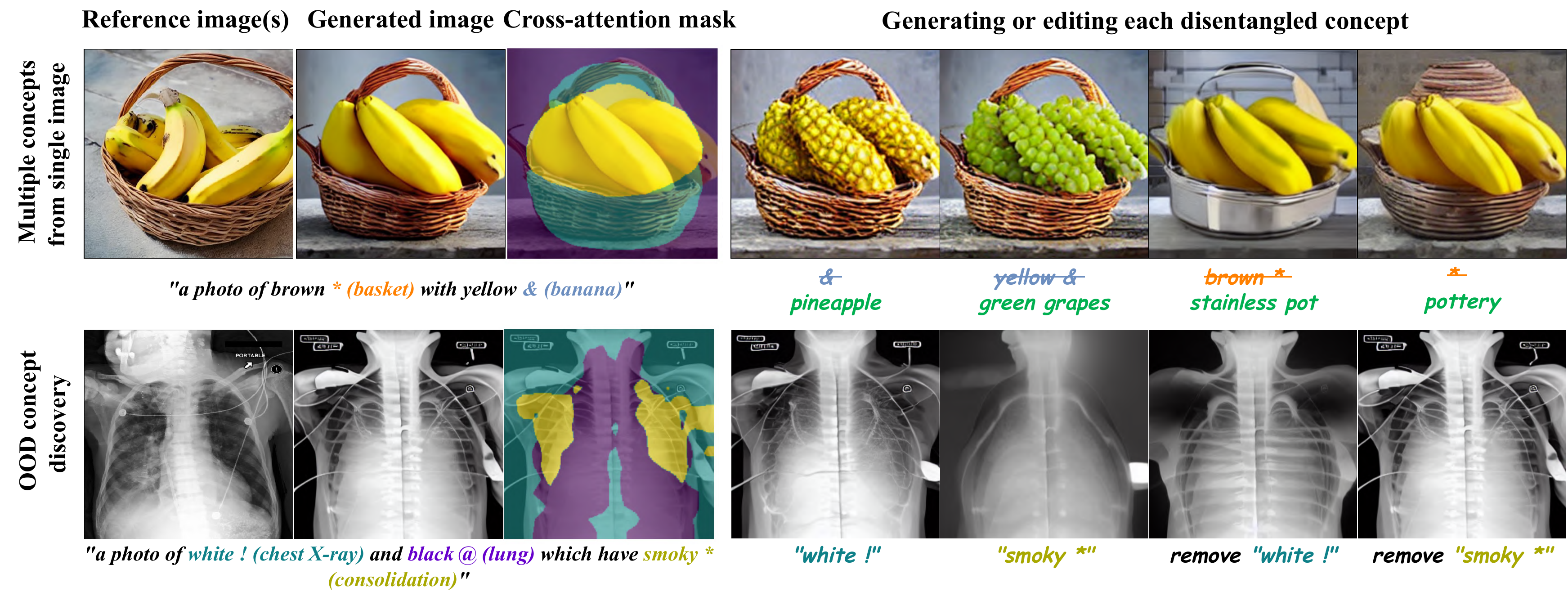}
    \vspace{-8mm}
    \caption{
    \textbf{MCPL learning and editing capabilities.}
    Top-row: learning and editing multiple concepts with a single string with input image from our generated 2-concept dataset (\ref{sec:dataset}). Bottom-row: learning to disentangle multiple unseen concepts from chest X-ray images with input image from MIMIC-CXR dataset \citep{johnson2019mimic}.}
    \label{fig:editing_capabilities}
\end{figure}

\subsection{Full t-SNE results}
\label{sec: full_tsne}

To assess disentanglement, we calculate and visualise the t-SNE projections. We approximate the ``ground truth" using features learned through \textit{Textual Inversion} on per-concept masked images. For benchmarking, we compare our method with the SoTA mask-based multi-concept learning method, \textit{Break-A-Scene (BAS)}, which acts as a performance benchmark but isn't directly comparable. 
Additionally, we assess variants integrating our proposed regularization techniques to align with the learning objectives. 

\begin{figure}[H]
    \centering
    \includegraphics[width=1\linewidth]{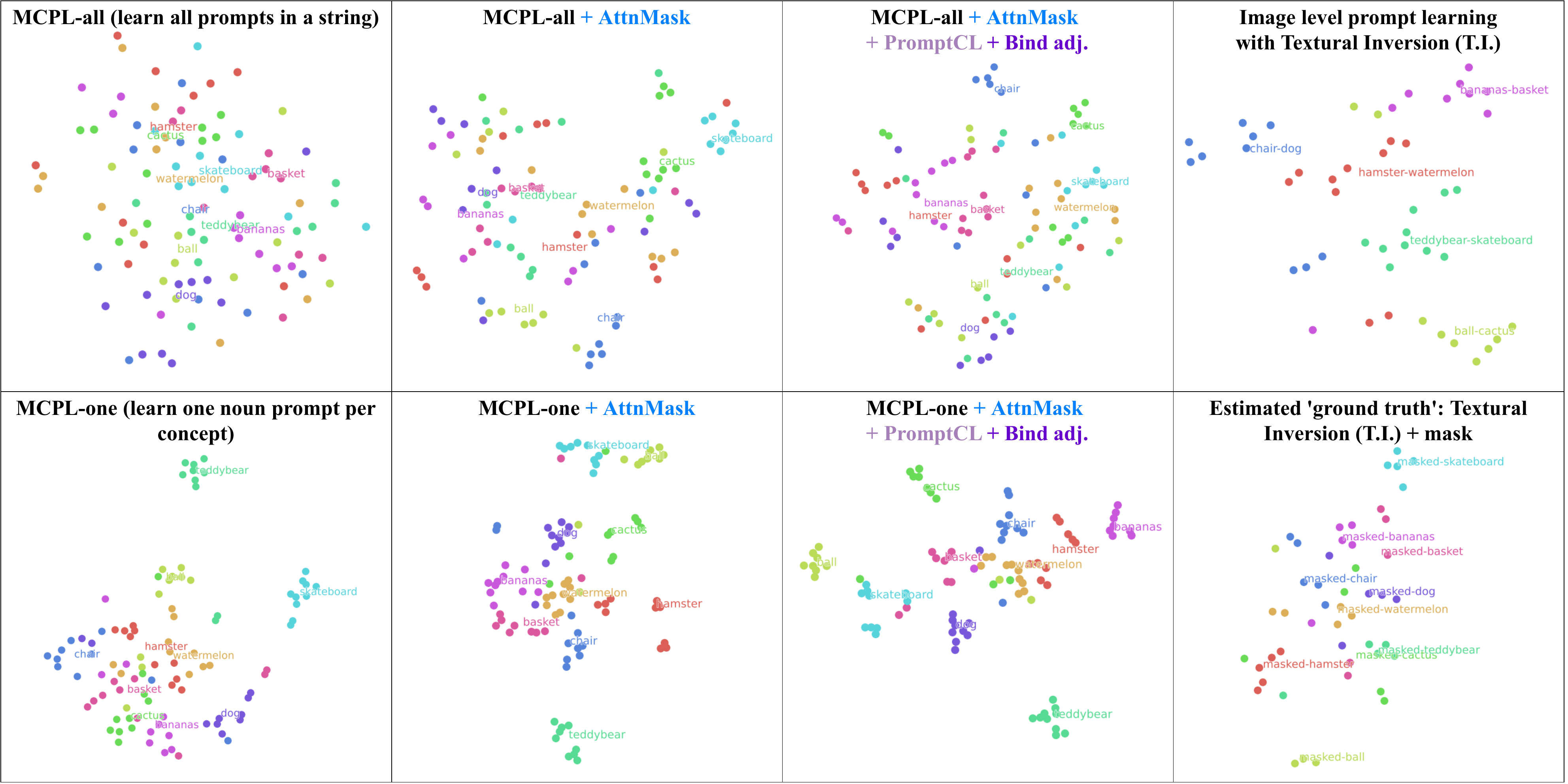}
    \caption{The t-SNE visualisations of learned prompt-concept features (comparing all variants) on the in-distribution natural (top) dataset. The results confirmed \textbf{our \textit{MCPL-one} combined with all regularisation terms can effectively distinguish all learned concepts} compared to all baselines. }
    \label{fig:t_sne_full}
\end{figure}

\begin{figure}[H]
    \centering
    \includegraphics[width=1\linewidth]{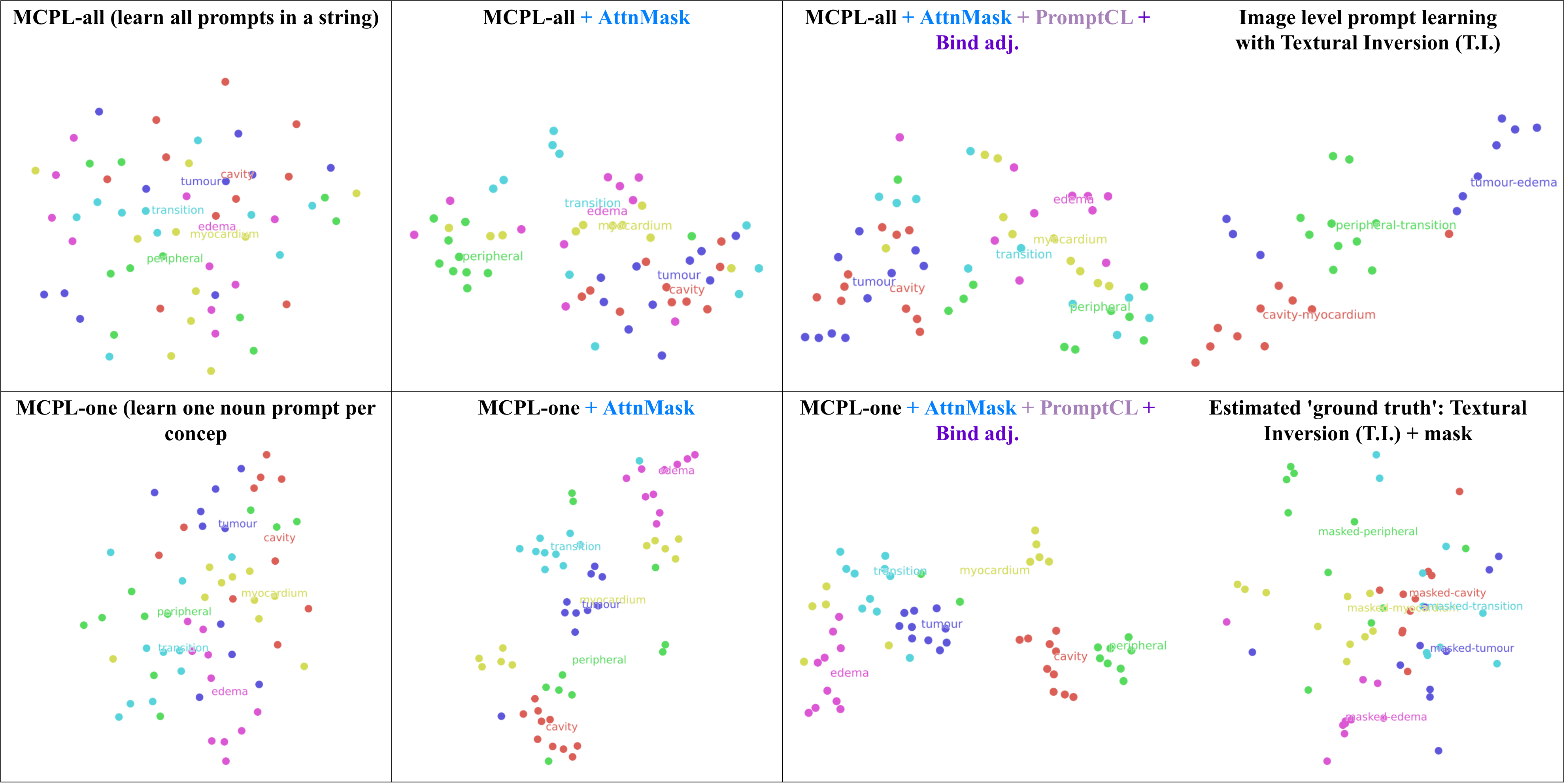}
    \caption{The t-SNE visualisations of learned prompt-concept features (comparing all variants) on the out-distribution medical (bottom) dataset. The results confirmed \textbf{our \textit{MCPL-one} combined with all regularisation terms can effectively distinguish all learned concepts} compared to all baselines. }
    \label{fig:t_sne_full_med}
\end{figure}

\subsection{All object-level embedding similarity of the learned concept relative to the estimated
“ground truth”.}
\label{sec: emb_similarity_all_objects}
\begin{figure}[H]
    \centering

    \begin{minipage}{0.49\textwidth}
        \centering
        \includegraphics[width=\textwidth]{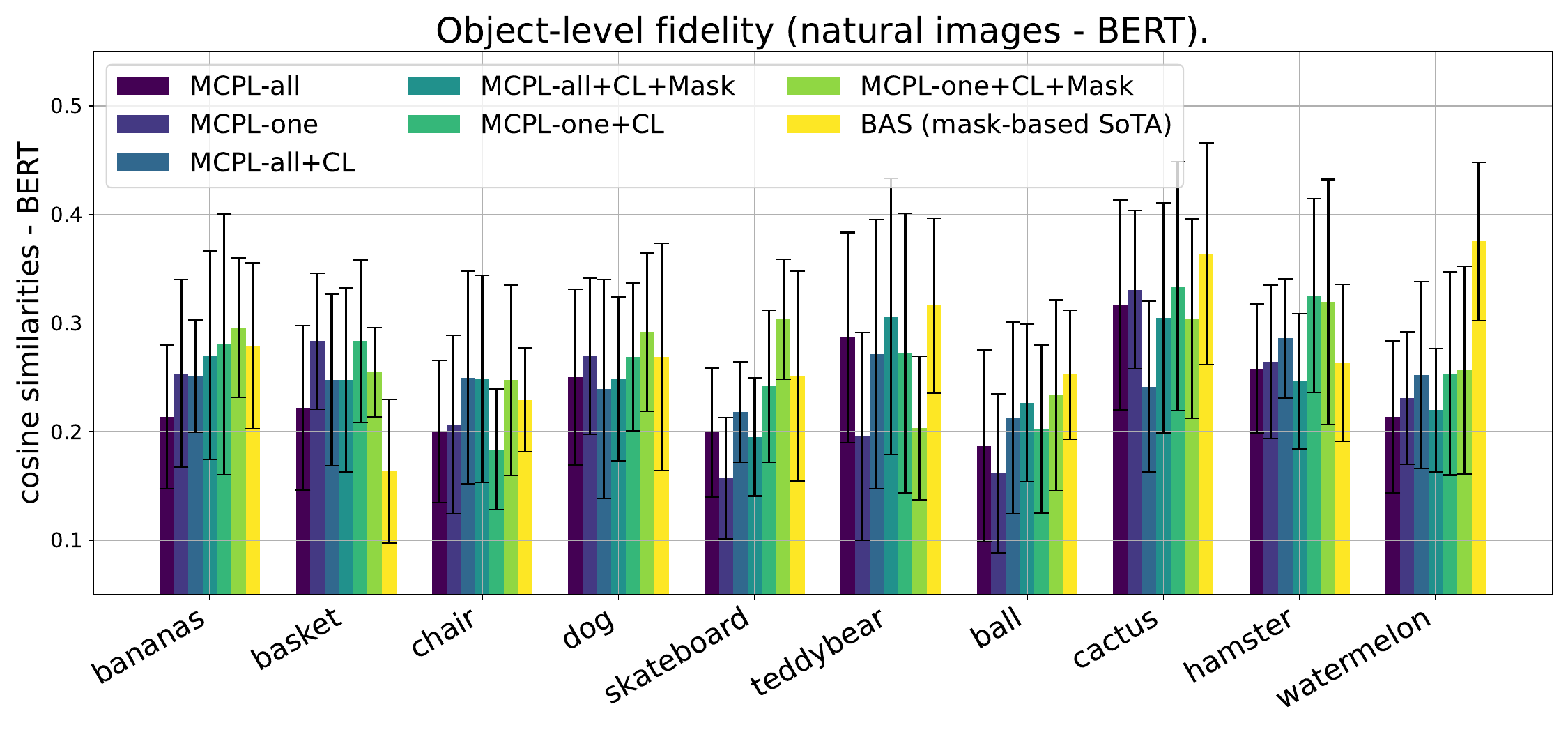} 
    \end{minipage}
    \hfill
    \begin{minipage}{0.49\textwidth}
        \centering
        \includegraphics[width=\textwidth]{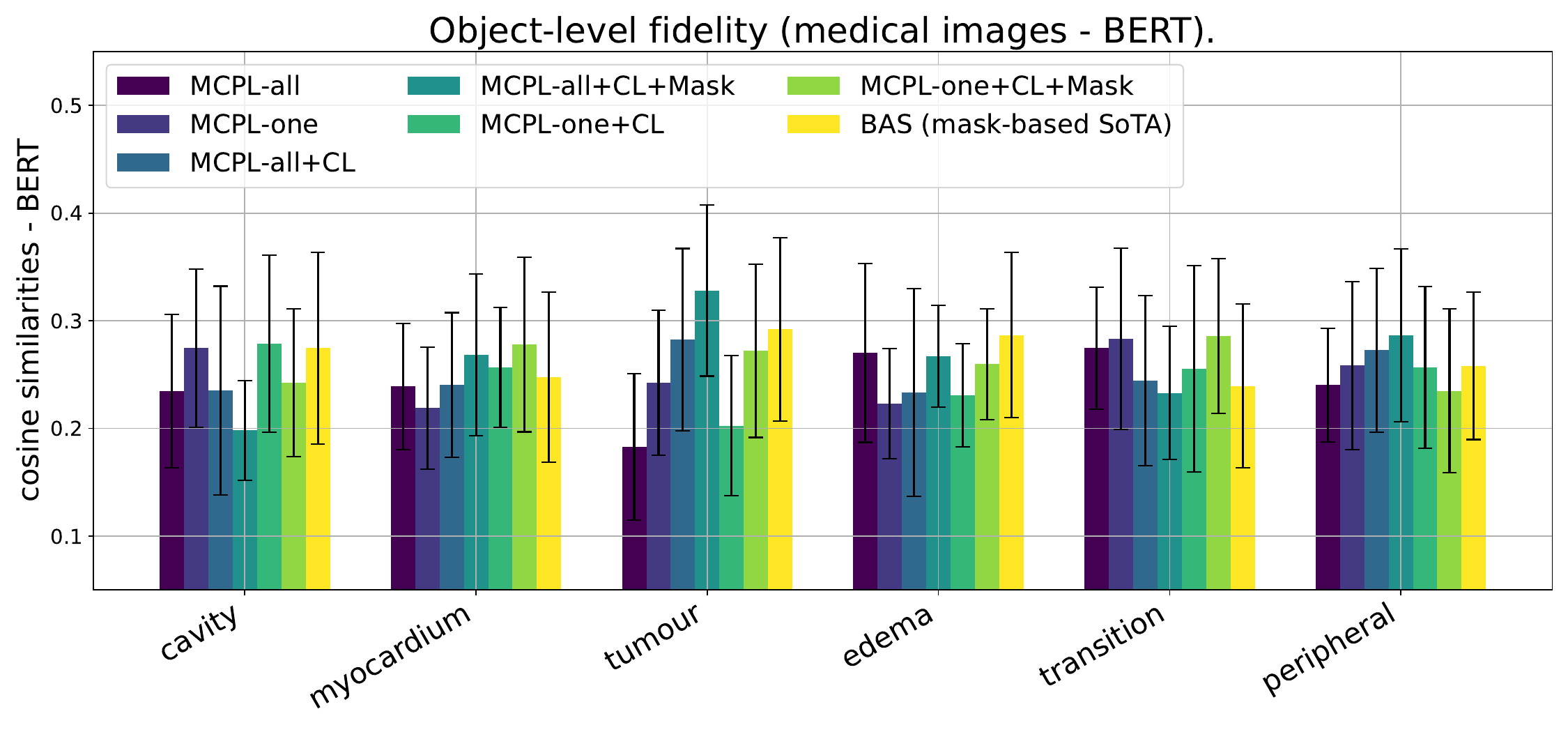}
    \end{minipage}
    
    \begin{minipage}{0.49\textwidth}
        \centering
        \includegraphics[width=\textwidth]{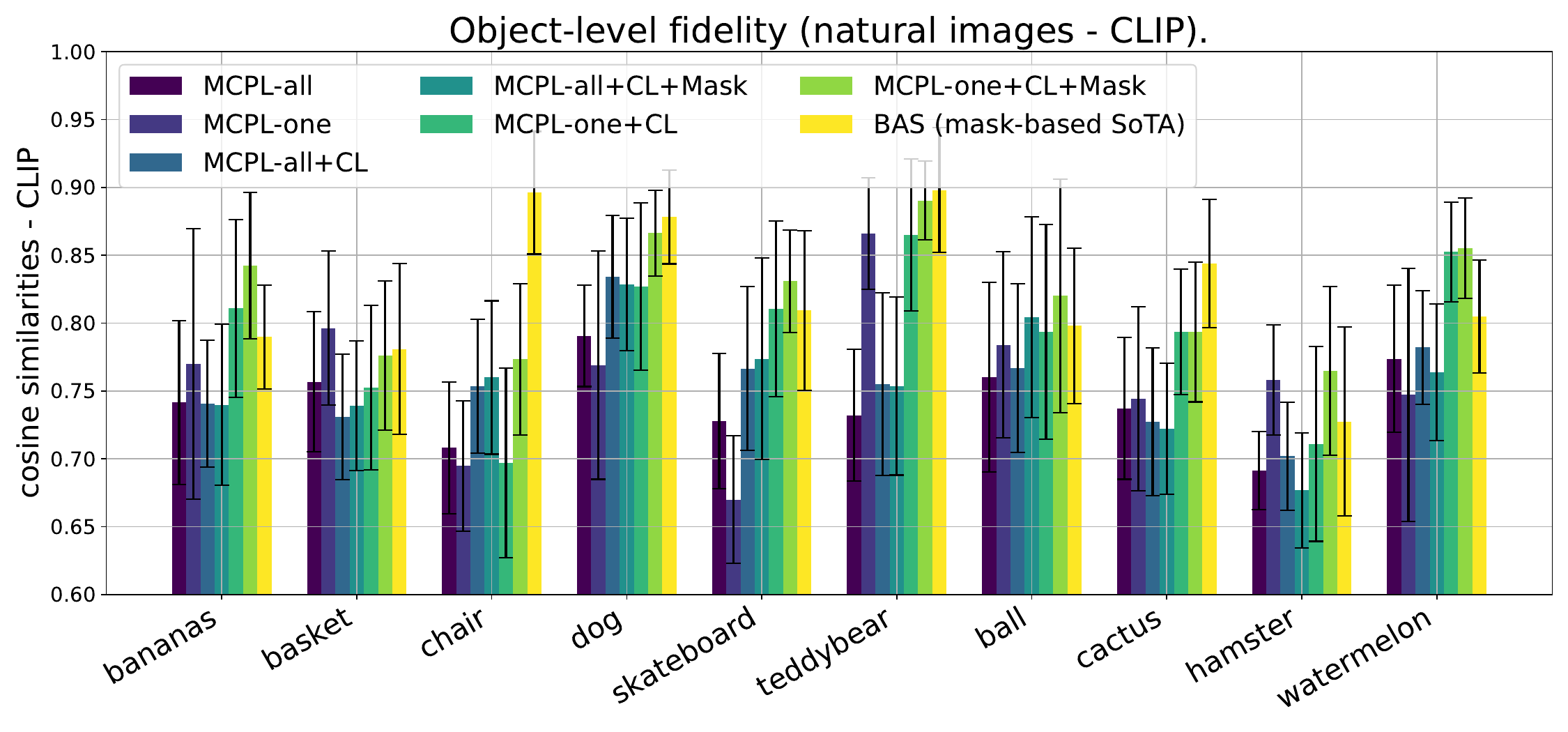} 
    \end{minipage}
    \hfill
    \begin{minipage}{0.49\textwidth}
        \centering
        \includegraphics[width=\textwidth]{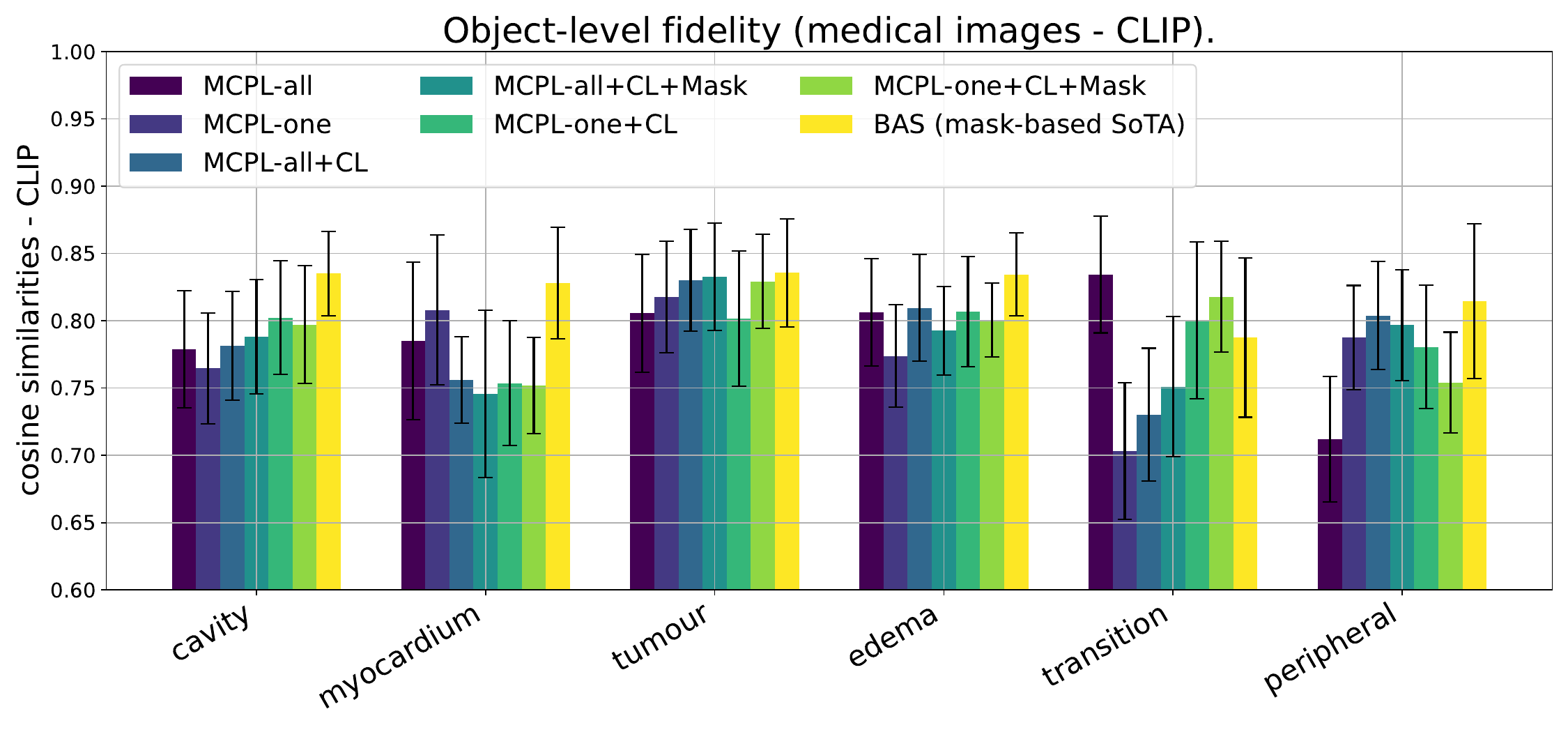}
    \end{minipage}
    
    \begin{minipage}{0.49\textwidth}
        \centering
        \includegraphics[width=\textwidth]{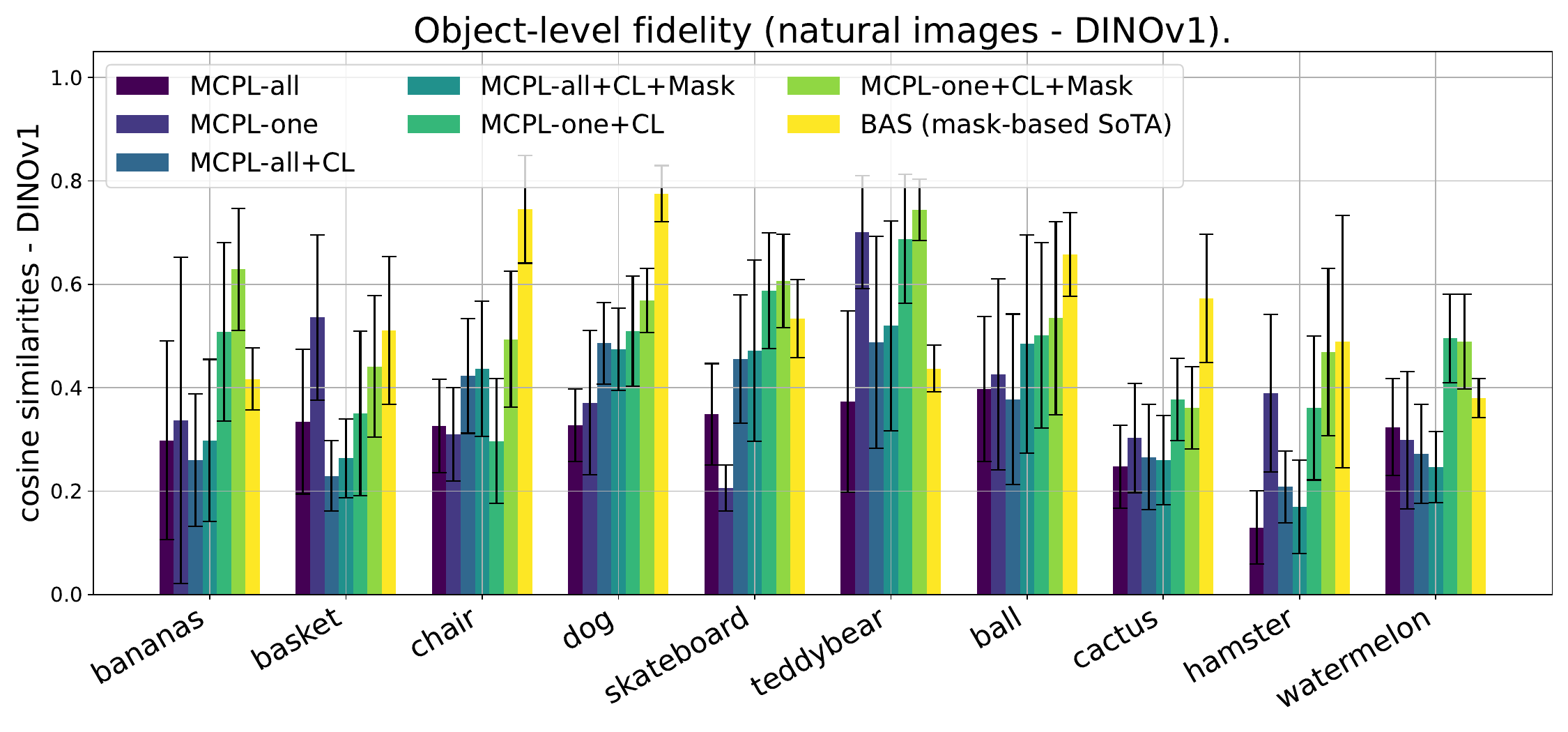} 
    \end{minipage}
    \hfill
    \begin{minipage}{0.49\textwidth}
        \centering
        \includegraphics[width=\textwidth]{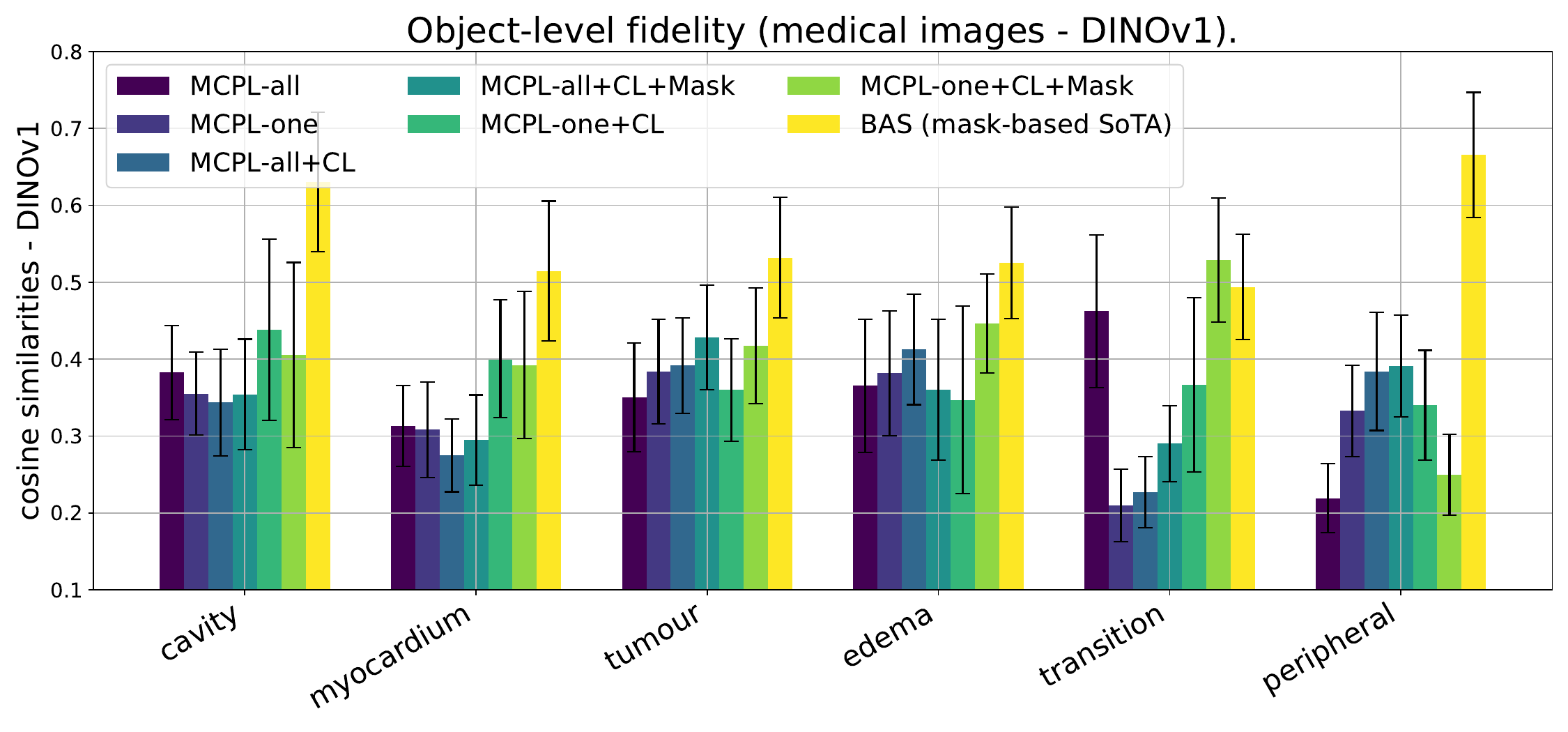}
    \end{minipage}
    

    \begin{minipage}{0.49\textwidth}
        \centering
        \includegraphics[width=\textwidth]{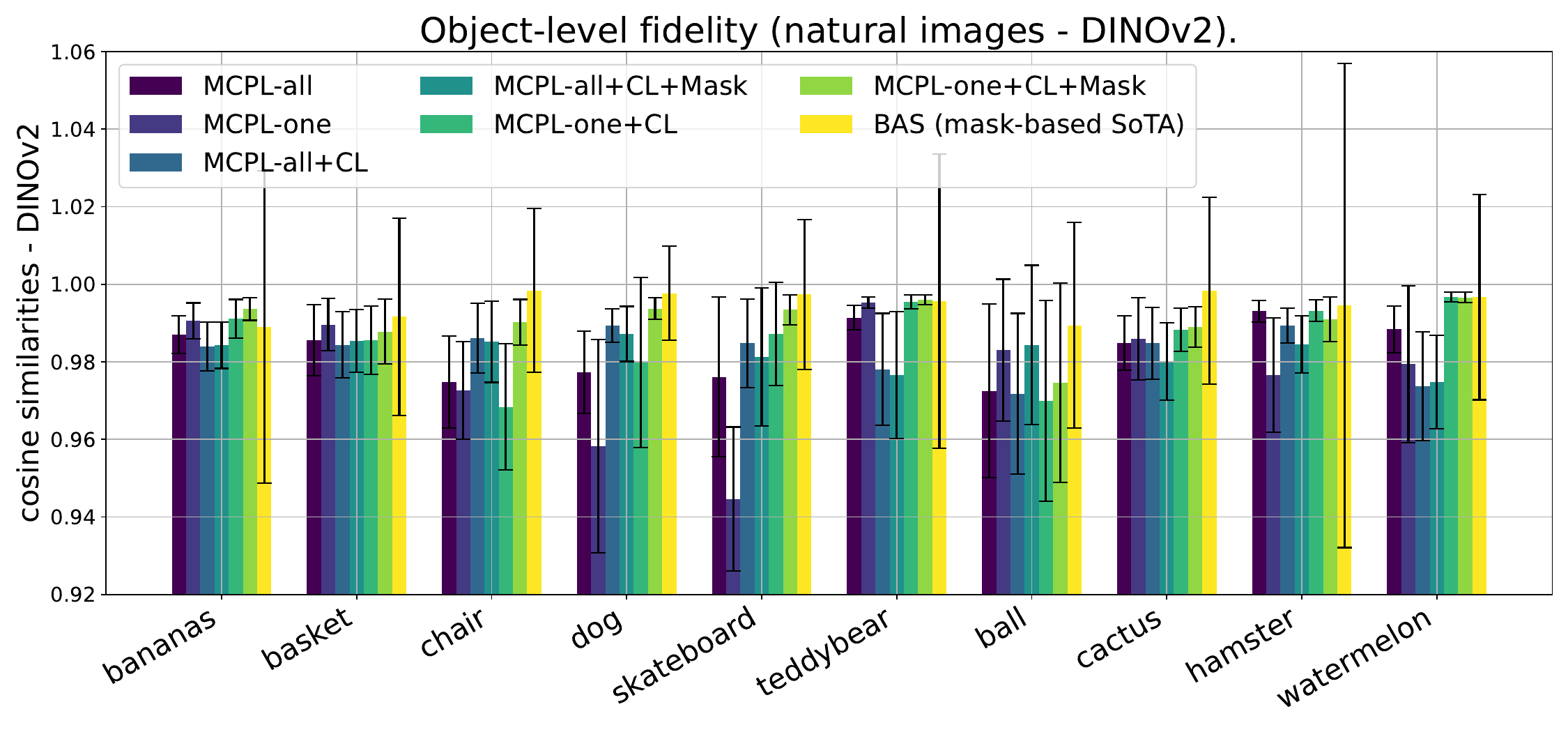} 
    \end{minipage}
    \hfill
    \begin{minipage}{0.49\textwidth}
        \centering
        \includegraphics[width=\textwidth]{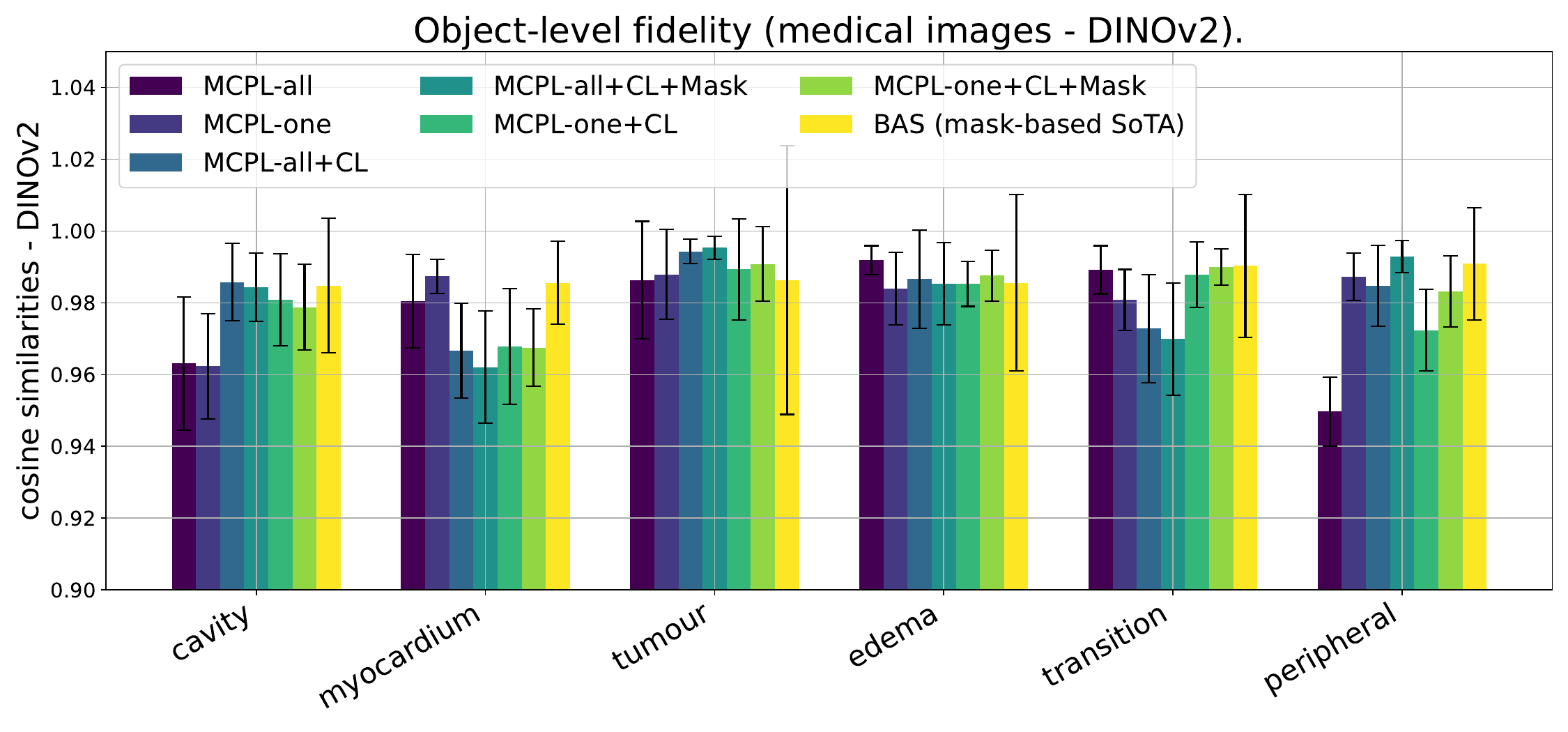}
    \end{minipage}

    \caption{\footnotesize Two-concepts natural (left column) and medical (right column) per-object embedding similarity between the learned concept relative to the masked ``ground truth". Each plot computes either the textural or image embeddings in one of the four embedding spaces (BERT, CLIP, DINOv1 and DINOv2). Each bar represents 4000 (natural images) or 6000 (medical images) pairwise cosine similarities. We compare our base version adding variations of our proposed regularisation terms. We also compare against the state-of-the-art (SoTA) mask-based learning method, Break-A-Scene (BAS) \cite{avrahami2023break}.}
    \label{fig: all_emb_two_concepts}
\end{figure}

\begin{figure*}
    \centering
    \includegraphics[width=.8\textwidth]{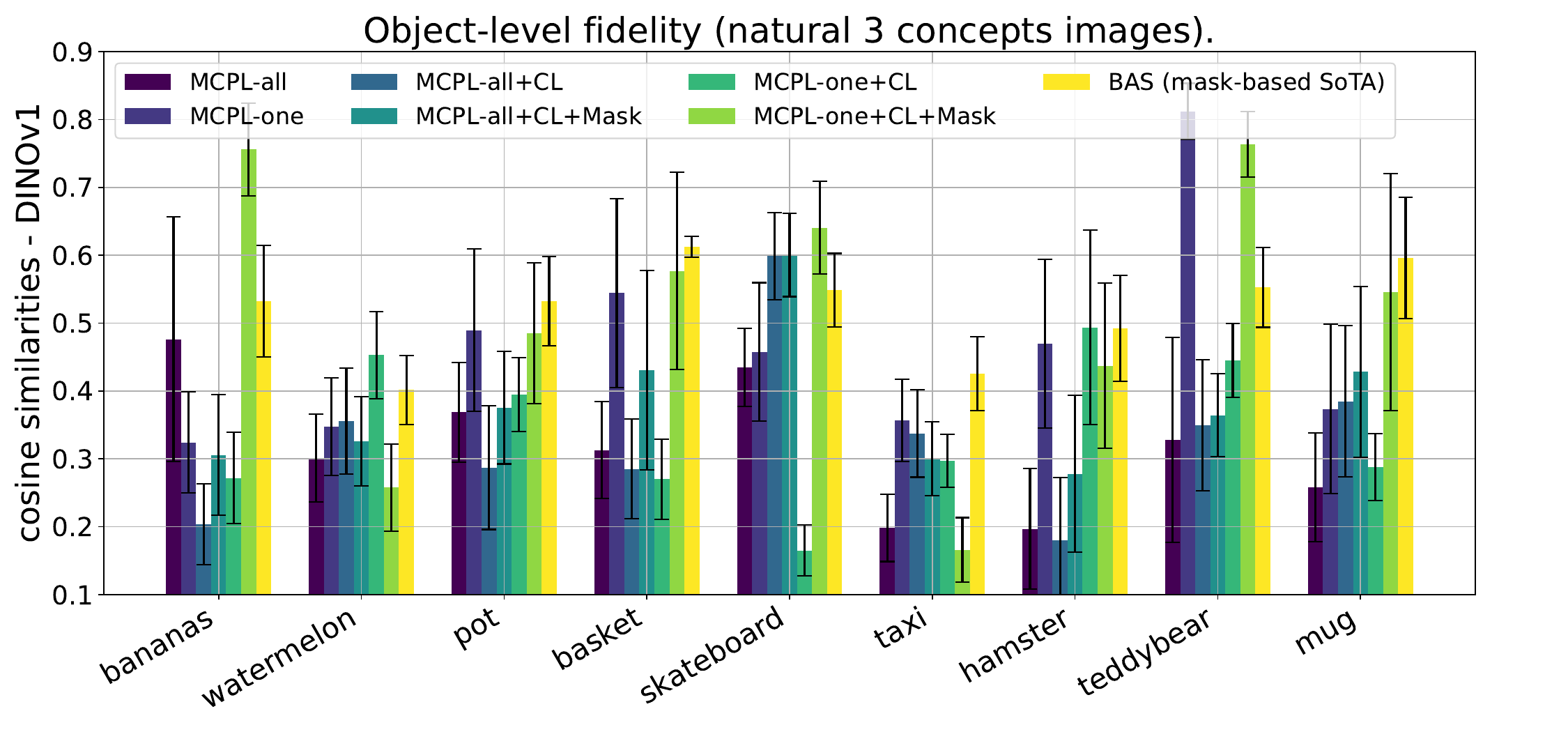}
    \includegraphics[width=.8\textwidth]{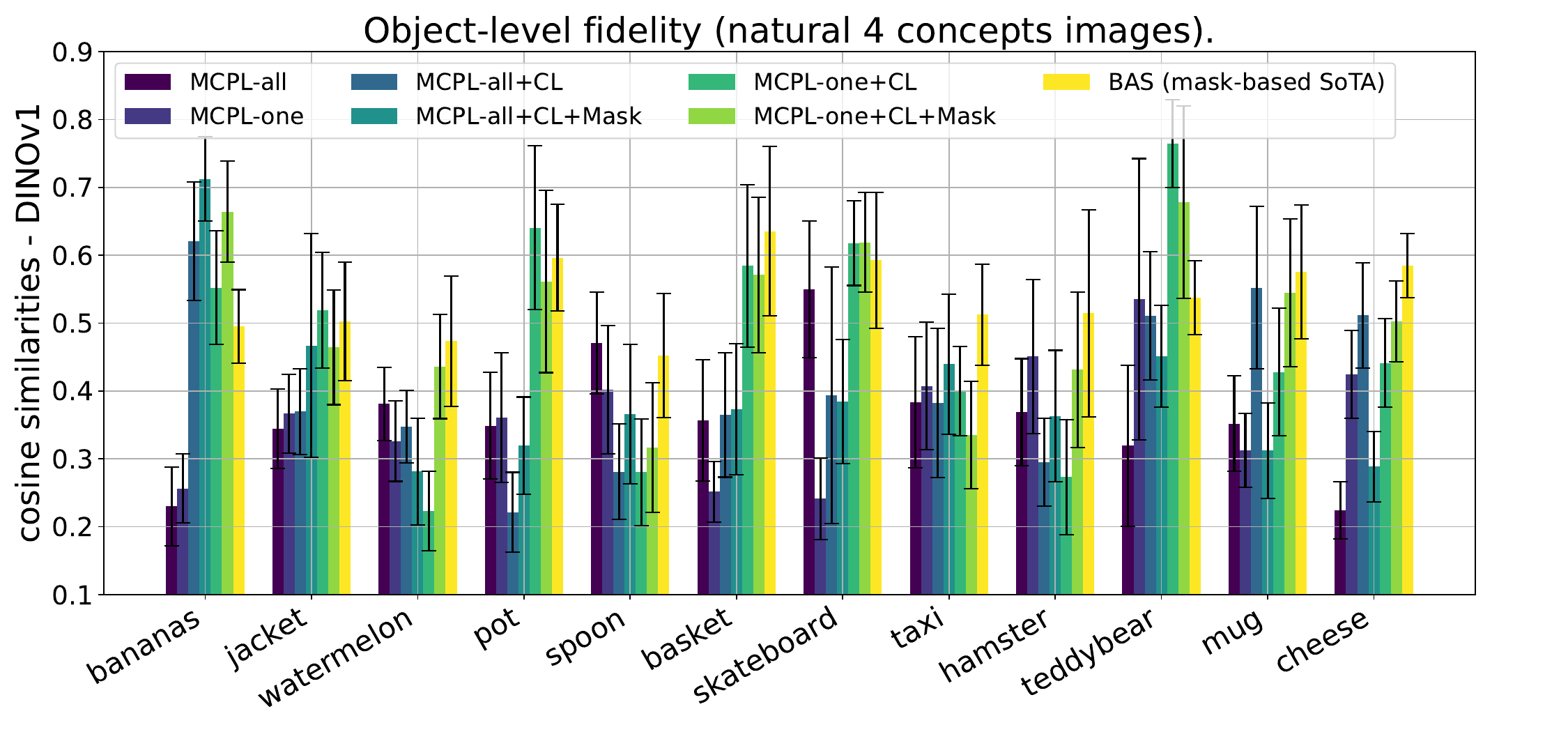}
    \includegraphics[width=.8\textwidth]{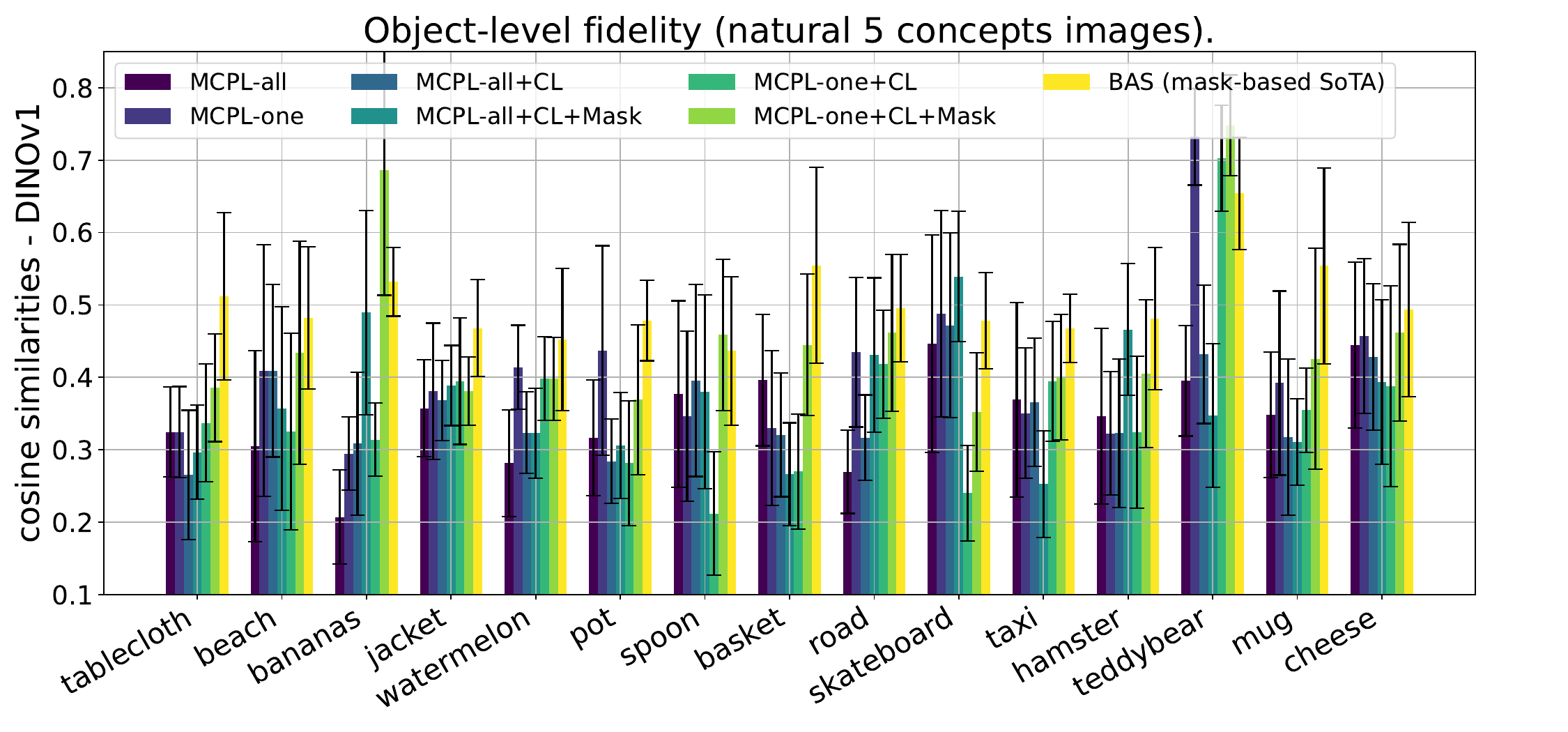}
    \caption{\footnotesize Natural images (each containing 3,4 or 5 objects) per-object embedding similarity between the learned concept relative to the masked ``ground truth". Each plot computes image embeddings in the embedding spaces of DINOv1. We compare our base version adding variations of our proposed regularisation terms. \textbf{Notably, our method sometimes surpasses the SoTA mask-based technique, Break-A-Scene \cite{avrahami2023break}, at the object level such as bananas and teddybear.}}
    \label{fig: emb_similarity_345}
\end{figure*}

To assess how well our method preserves object-level semantic and textual details, we evaluate both prompt and image fidelity. Our experiments, depicted in Figures \ref{fig: all_emb_two_concepts} and \ref{fig: emb_similarity_345}, involve learning from dual-concept natural and medical images, and from 3 to 5 natural images. Results \textbf{consistently demonstrate that our method adding all proposed regularisation terms accurately learns object-level embeddings}, in comparison with masked ground truth in pre-trained embedding spaces such as BERT, CLIP, DINOv1, and DINOv2. Notably, our method sometimes surpasses the state-of-the-art mask-based technique, Break-A-Scene, at the object level. This underscores the effectiveness and potential of our mask-free, language-driven approach to learning multiple concepts from a single image.

\newpage

\subsection{Full motivational experiment results}
\label{sec:full_motivation}
\begin{figure}[H]
    \centering
    \includegraphics[width=1\linewidth]{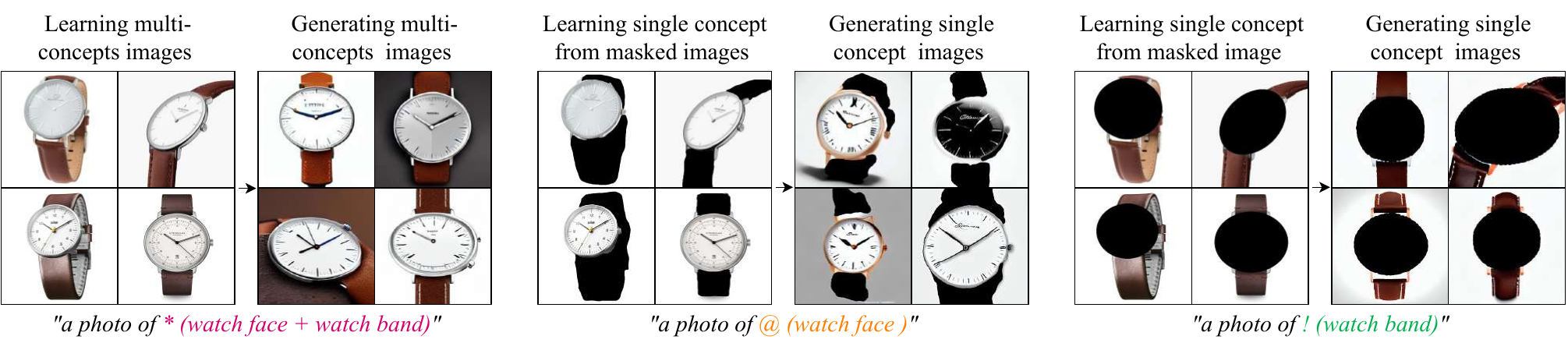}
    \vspace*{-6mm}
    \caption{\textbf{Motivational study with watch images.} We learn embeddings using Textual Inversion on both unmasked multi-concept images (``watch face" and ``watch band") and masked single-concept images (``watch face" or ``watch band").}
    \label{fig: motivation}
\end{figure}

\paragraph{Do multiple distinct embeddings arise from the same image?} 
To understand the possibility of learning multiple concepts within a frozen textual embedding space, we explored whether \textit{Textual Inversion} can discern semantically distinct concepts from processed images, each highlighting a single concept. Following \cite{wu2020stylespace}, we used images with manual masks to isolate concepts, as seen in \Figref{fig: motivation}. We applied \textit{Textual Inversion} to these images to learn embeddings for the unmasked or masked images. \textit{Our findings indicate that when focusing on isolated concepts, \textit{Textual Inversion} can successfully learn distinct embeddings, as validated by the generated representations of each concept.}

\begin{figure}[H]
    \centering
    \includegraphics[width=.8\linewidth]{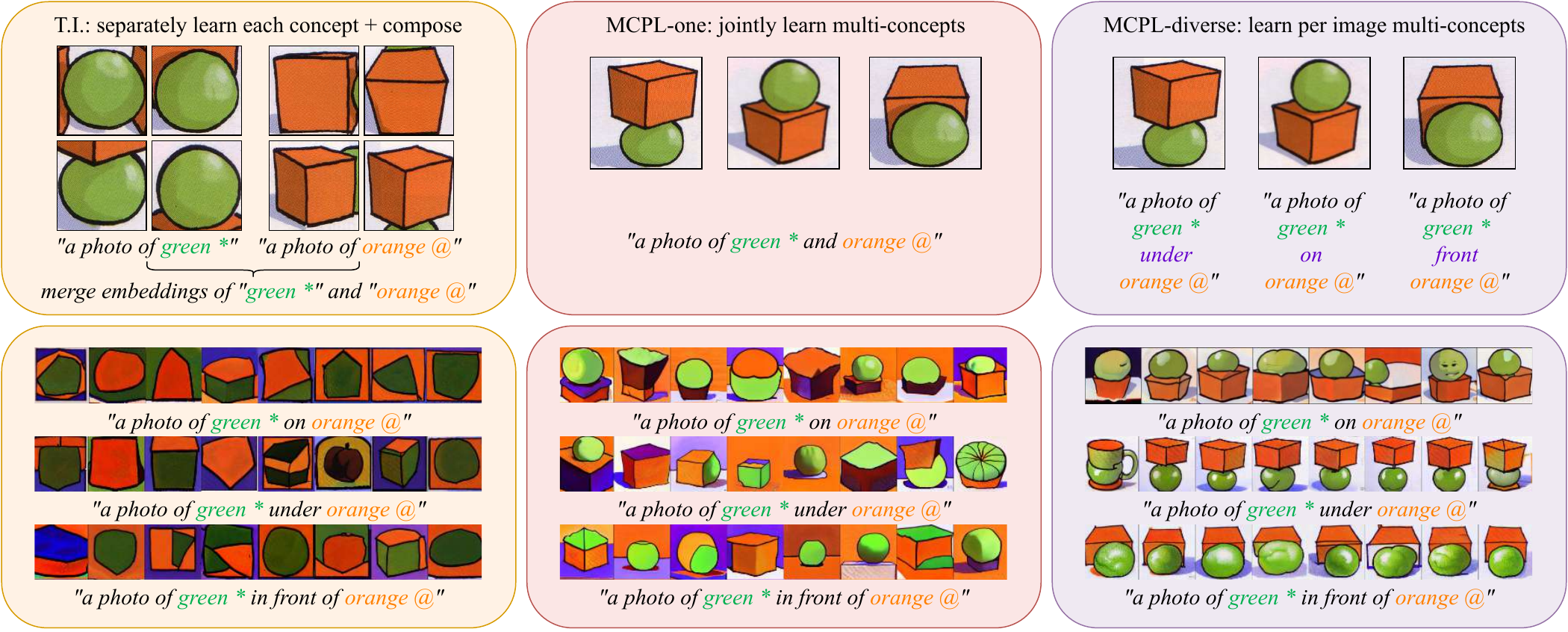}
    \caption{\textbf{Learning and Composing ``ball" and ``box"}. We learned the concepts of ``ball" and ``box" using different methods (top row) and composed them into unified scenes (bottom row). We compare three learning methods: \textit{Textual Inversion} \citep{gal2022textual}, which learns each concept separately from isolated images (left); \textit{MCPL-one}, which jointly learns both concepts from uncropped examples using a single prompt string (middle); and \textit{MCPL-diverse}, which advances this by learning both concepts with per-image specific relationships (right).}
    \label{fig: MCPL_vs_TI}
\end{figure}

\vspace{-4mm}
\paragraph{Is separate learning of concepts sufficient for multi-object image generation?}
While separate learning with carefully sampled or masked images in a multi-object scene deviates from our objective, it is valuable to evaluate its effectiveness. Specifically, we use Textual Inversion to separately learn concepts like ``ball" and ``box" from carefully cropped images, as shown in \Figref{fig: MCPL_vs_TI}. We then attempt to compose images using strings that combine these concepts, such as ``a photo of a green ball on orange box." \textit{Our results indicate that the accurate composition of multi-object images remains challenging, even when individual concepts are well-learned.}

\vspace{-4mm}
\subsection{Full ablation results of assessing regularisation terms with cross-attention}
\label{sec: full_ablation}
\begin{figure}[h]
  \centering
  \begin{minipage}{.515\linewidth}
    \centering
    \includegraphics[width=1\linewidth]{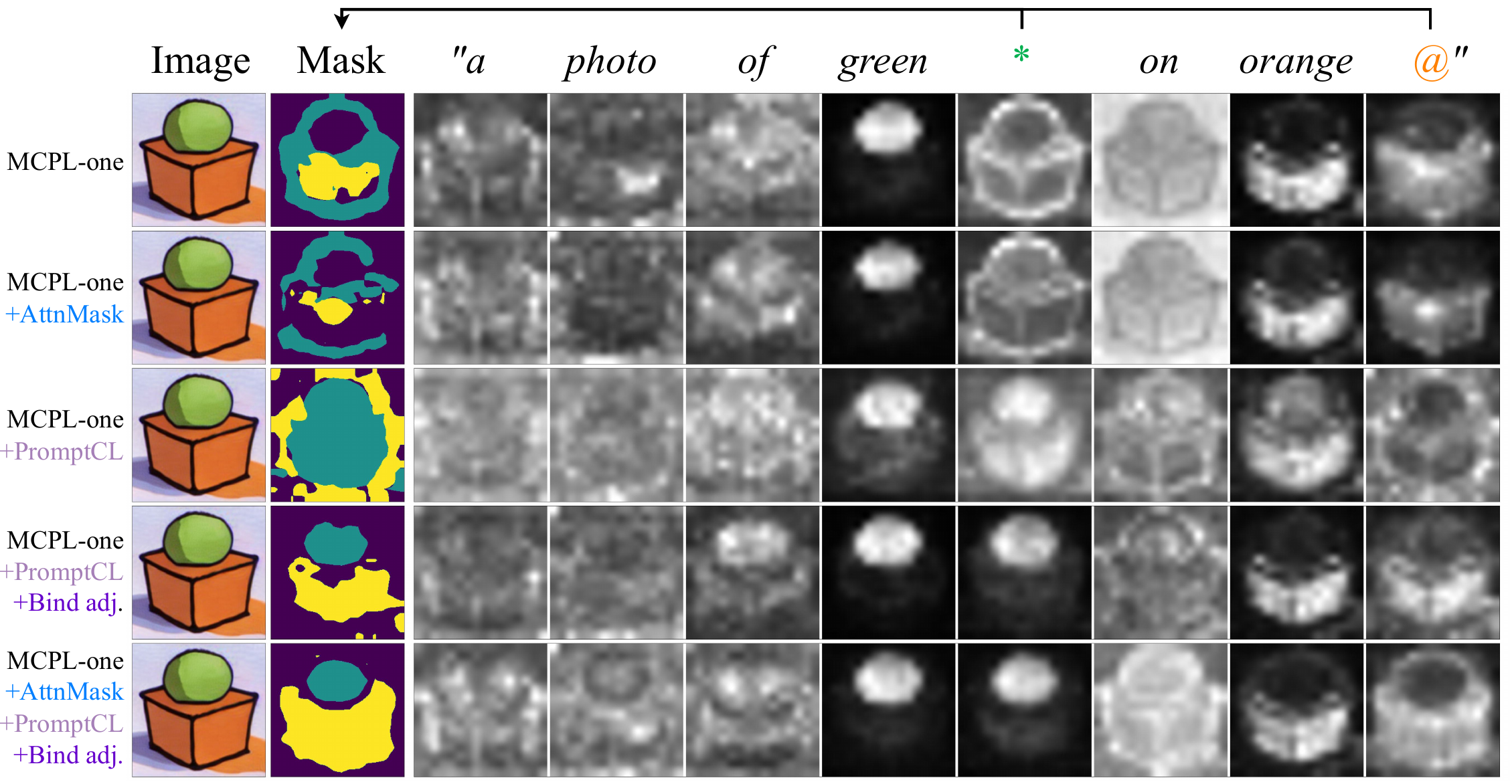}
  \end{minipage}%
  \begin{minipage}{.485\linewidth}
    \centering
    \includegraphics[width=1\linewidth]{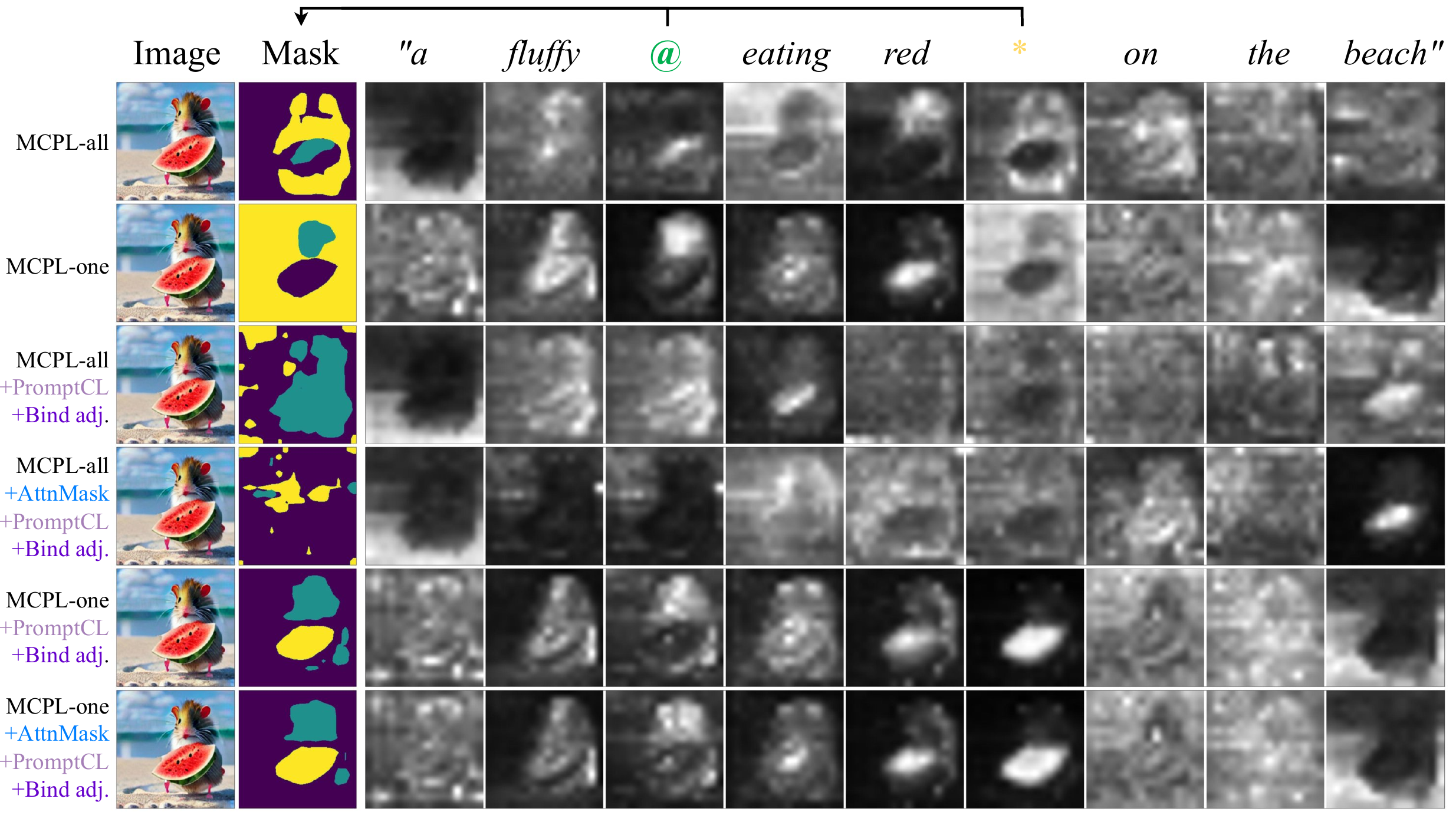}
  \end{minipage}
  \caption{Enhancing object-level prompt-concept correlation in MCPL using proposed \textit{AttnMask}, \textit{PromptCL} and \textit{Bind adj.} regularisation techniques. 
  We use average cross-attention maps to quantify the correlation of each prompt with its corresponding object-level concept. Additionally, we construct attention-based masks from multiple selected prompts for the concepts of interest. The visual results confirm that \textbf{incorporating all of the proposed regularisation terms enhances concept disentanglement, whereas applying them in isolation yields suboptimal outcomes.}}
  
    \label{fig:improve_MCPL}
\end{figure}

\begin{figure}[H]
\centering
\renewcommand{\arraystretch}{1.5} 

\begin{tabular}{c c c}
\hspace*{-.5cm}
\includegraphics[trim={0 0 0 0},clip,width=0.315\textwidth]{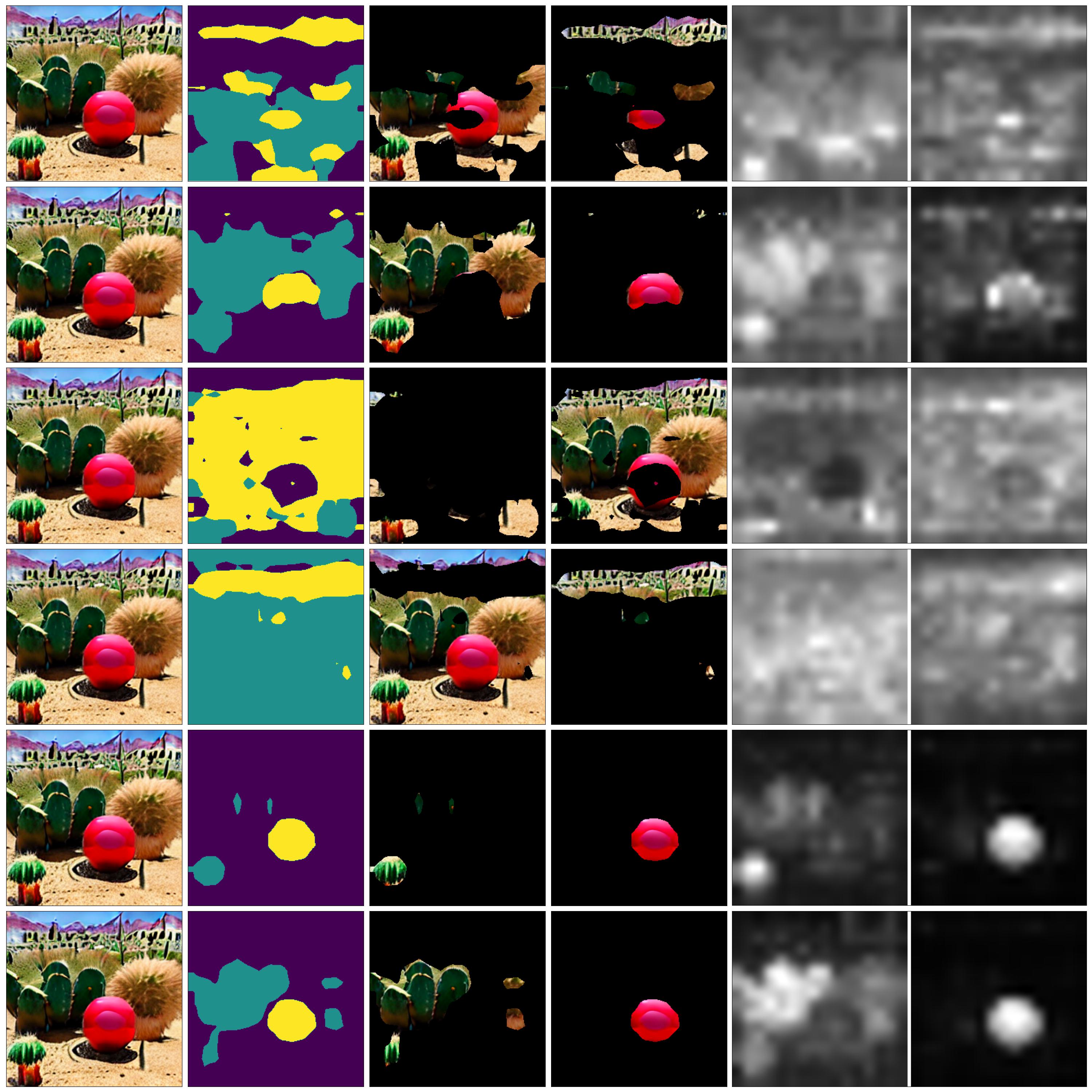} &
\includegraphics[trim={0 0 0 0},clip,width=0.315\textwidth]{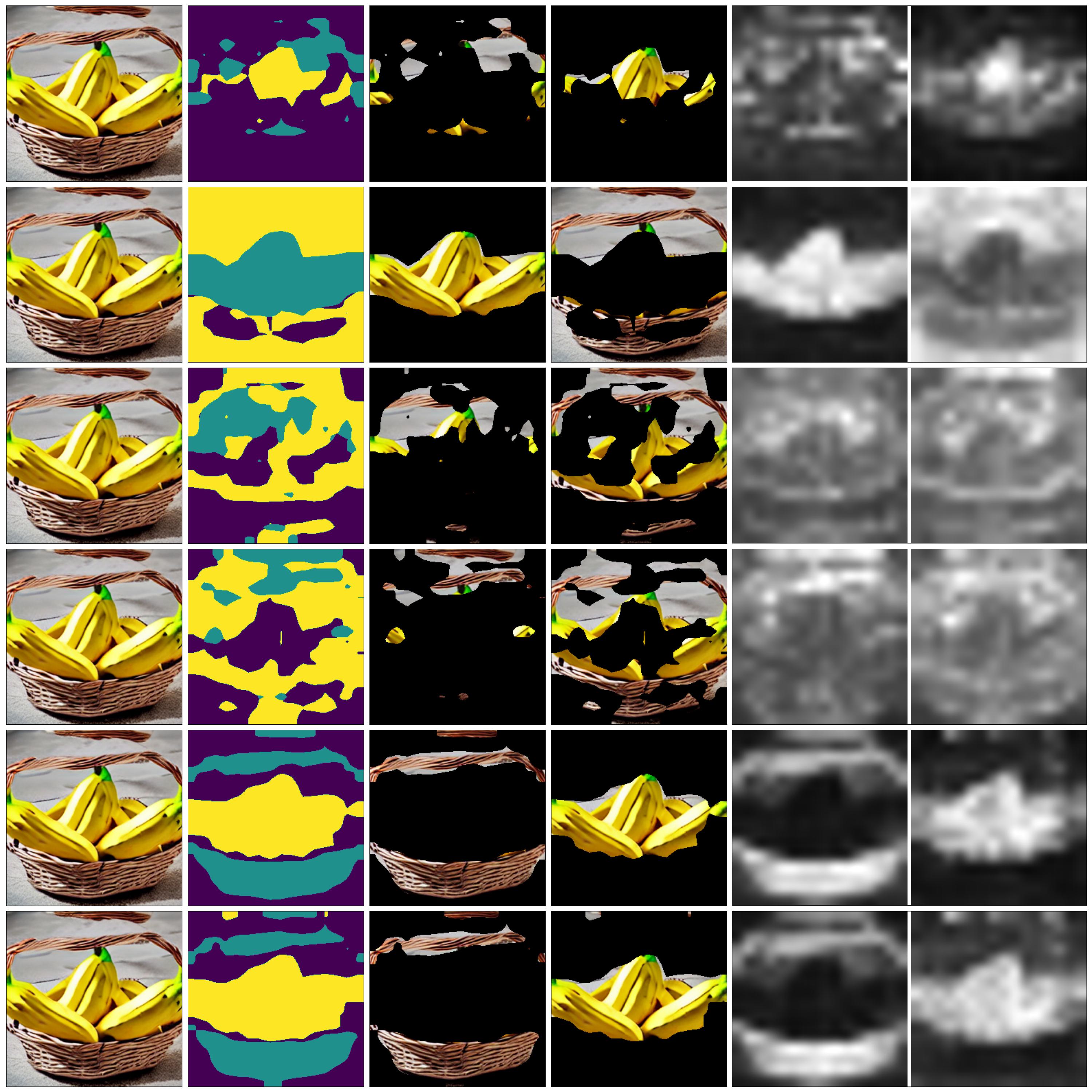} &
\includegraphics[trim={0 0 0 0},clip,width=0.315\textwidth]{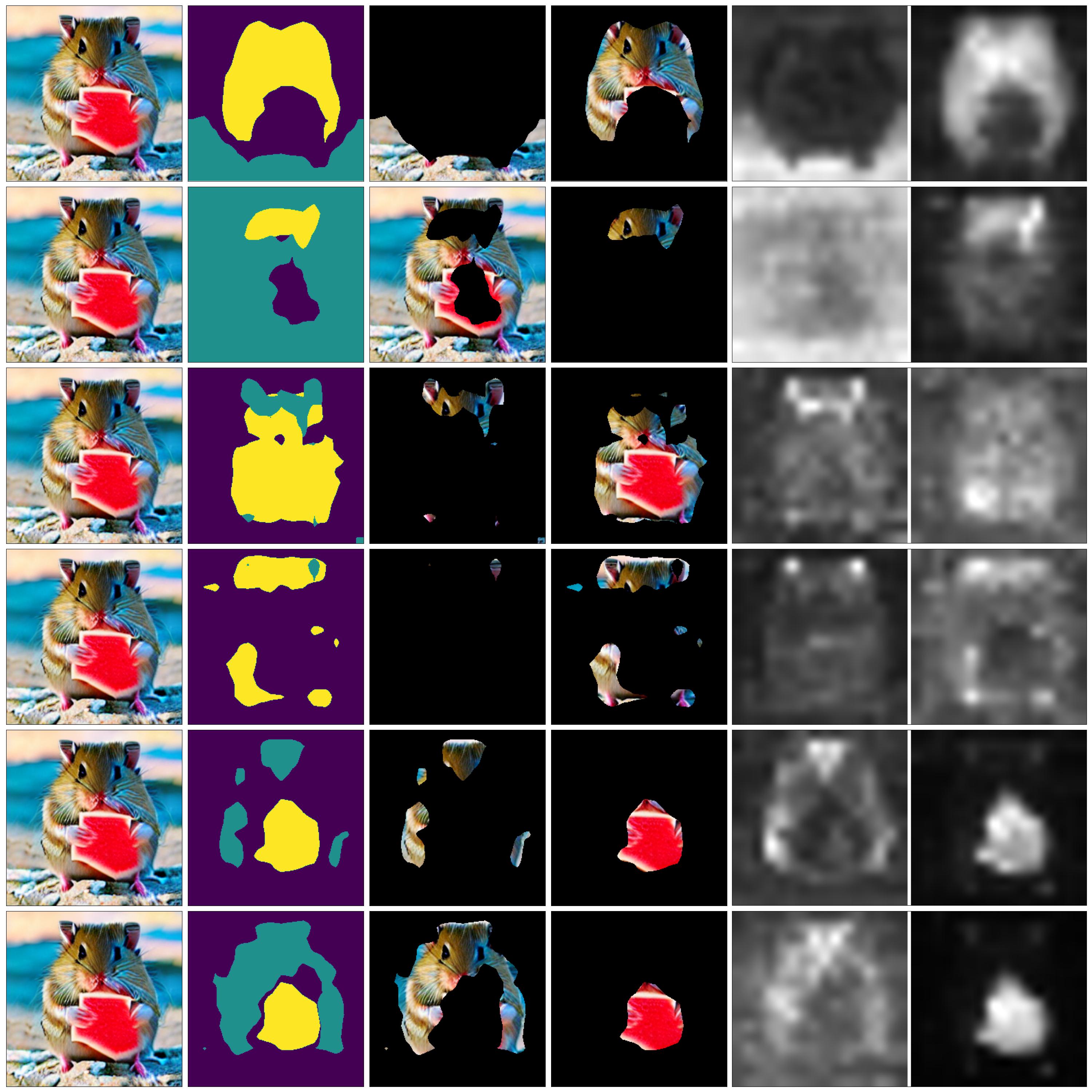} \\
Ball Cactus & Bananas Basket & Hamster Watermelon \\
\end{tabular}

\begin{tabular}{c c c}
\hspace*{-.5cm}
\includegraphics[trim={0 0 0 0},clip,width=0.315\textwidth]{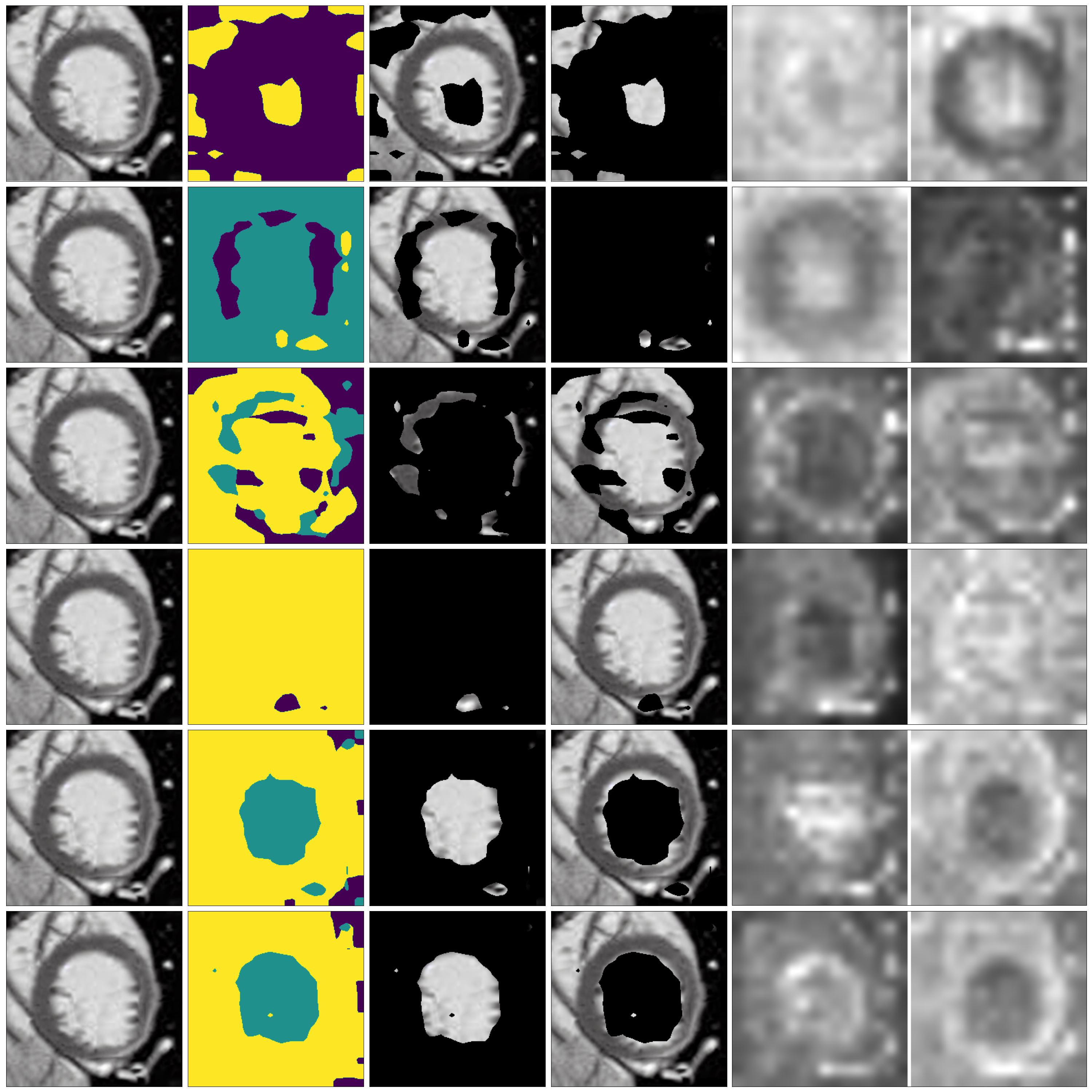} &
\includegraphics[trim={0 0 0 0},clip,width=0.315\textwidth]{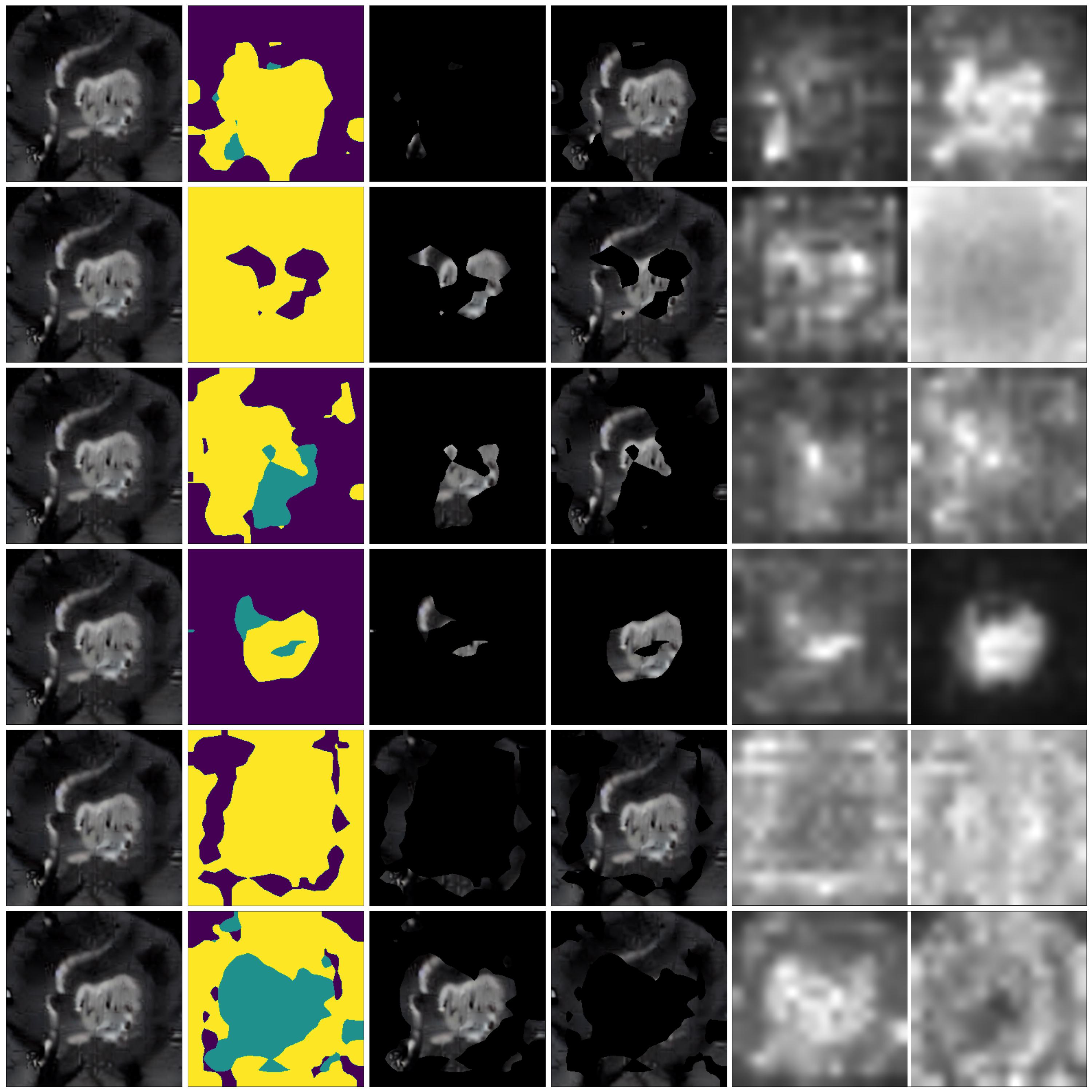} &
\includegraphics[trim={0 0 0 0},clip,width=0.315\textwidth]{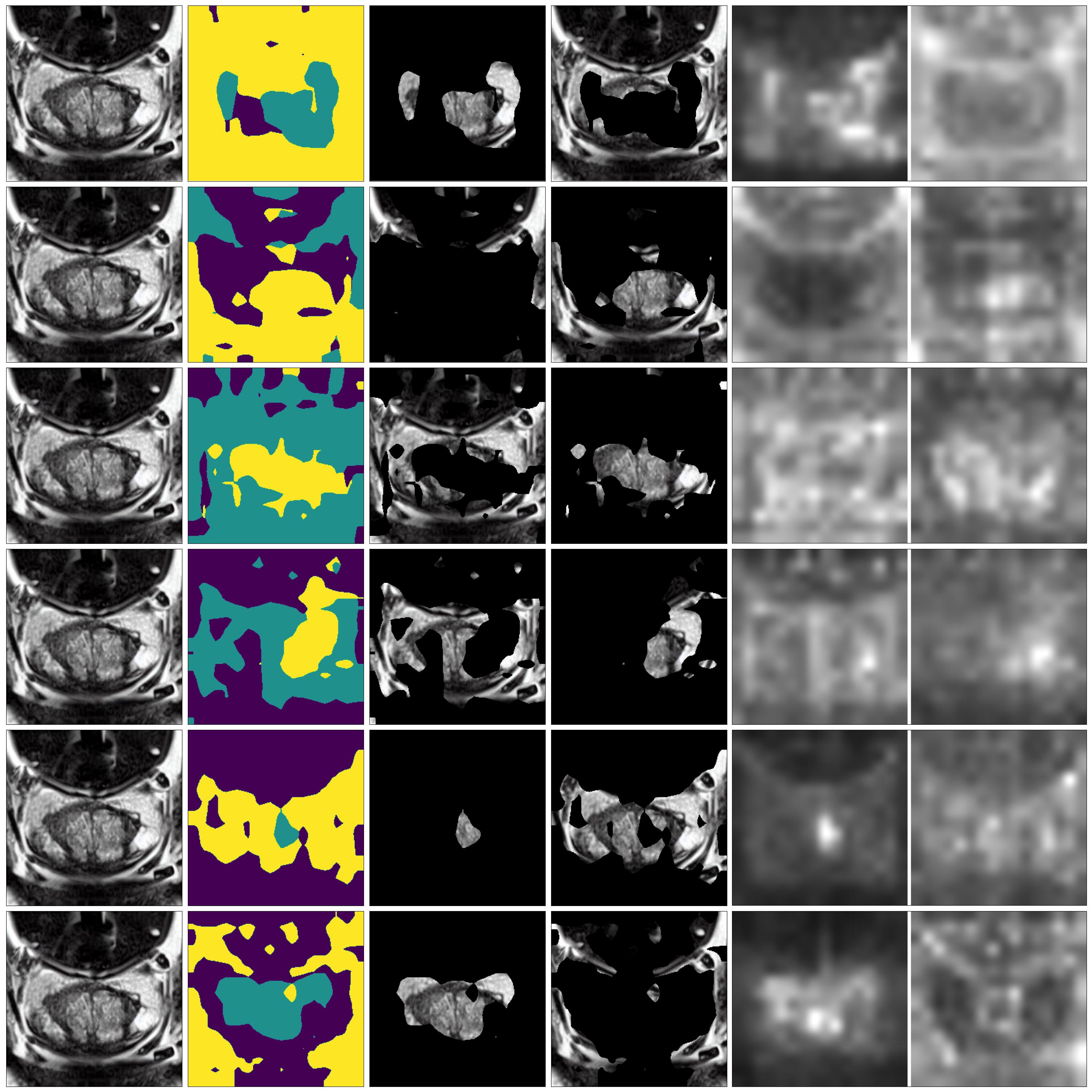} \\
Cavity Myocardium & Tumour Edema & Peripheral Transition \\
\end{tabular}

\caption{Visualisation of concepts (two concepts). 
We compare all baseline methods across each row (from top to bottom): 
1) MCPL-all, 
2) MCPL-one, 
3) MCPL-all+\textit{PromptCL}+\textit{Bind adj.}, 
4) MCPL-all+\textit{AttnMask}+\textit{PromptCL}+\textit{Bind adj.}, 
5) MCPL-one+\textit{PromptCL}+\textit{Bind adj.}, 
6) MCPL-one+\textit{AttnMask}+\textit{PromptCL}+\textit{Bind adj.}.
\textbf{The results on both the natural and medical images confirmed our conclusion} --- the inclusion of all proposed regularisation terms (toward the bottom row) consistently demonstrated their effectiveness in enhancing the accuracy of prompt-concept correlation.
}
\label{fig:visualise_learned_masked}
\end{figure}

\begin{figure}[H]
\centering
\renewcommand{\arraystretch}{1.5} 

\begin{tabular}{c c}
\hspace*{-.3cm}
\includegraphics[trim={0 0 0 0},clip,width=0.445\textwidth]{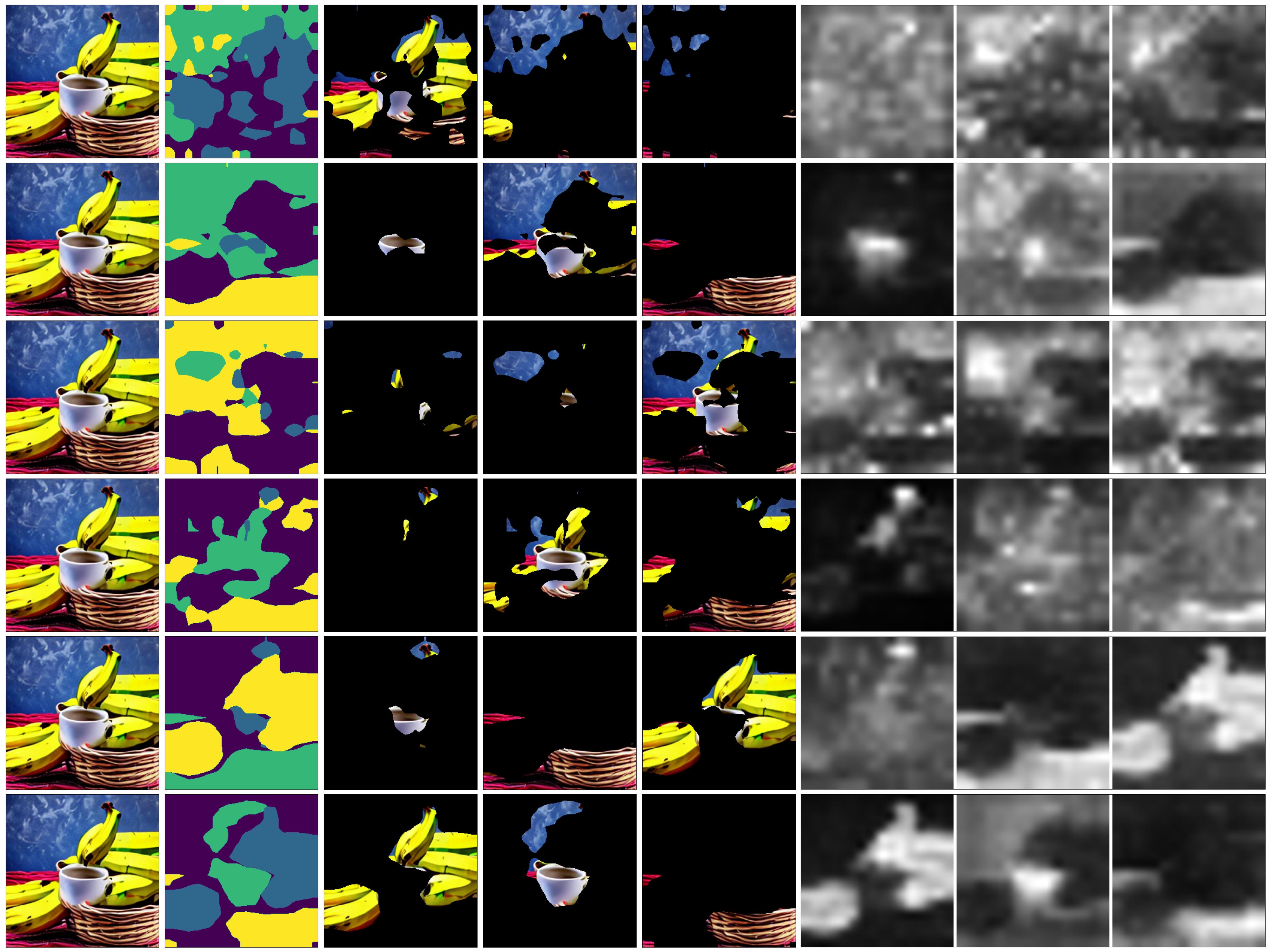} &
\includegraphics[trim={0 0 0 0},clip,width=0.555\textwidth]{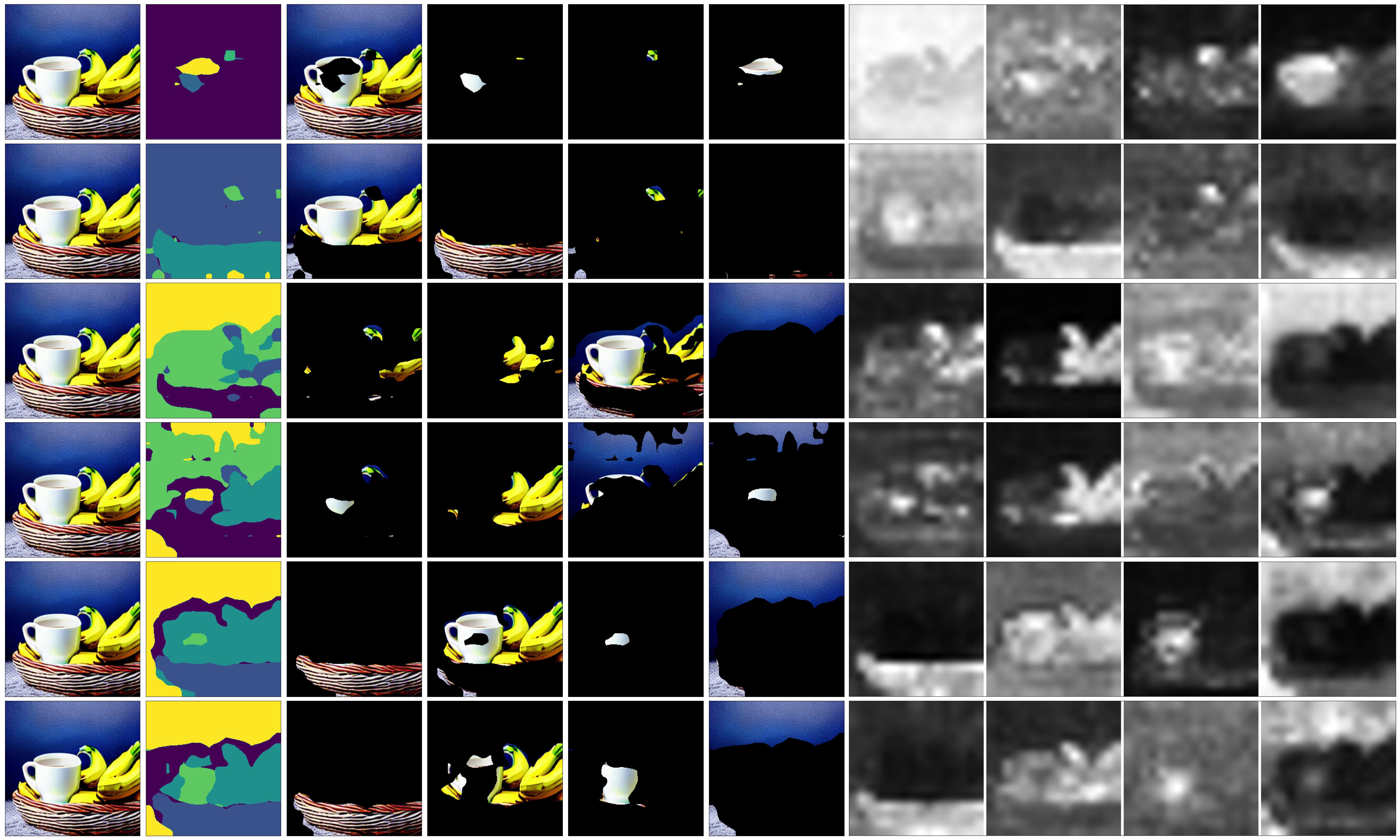} \\
\end{tabular}

\begin{tabular}{c c}
\hspace*{-.3cm}
\includegraphics[trim={0 0 0 0},clip,width=0.445\textwidth]{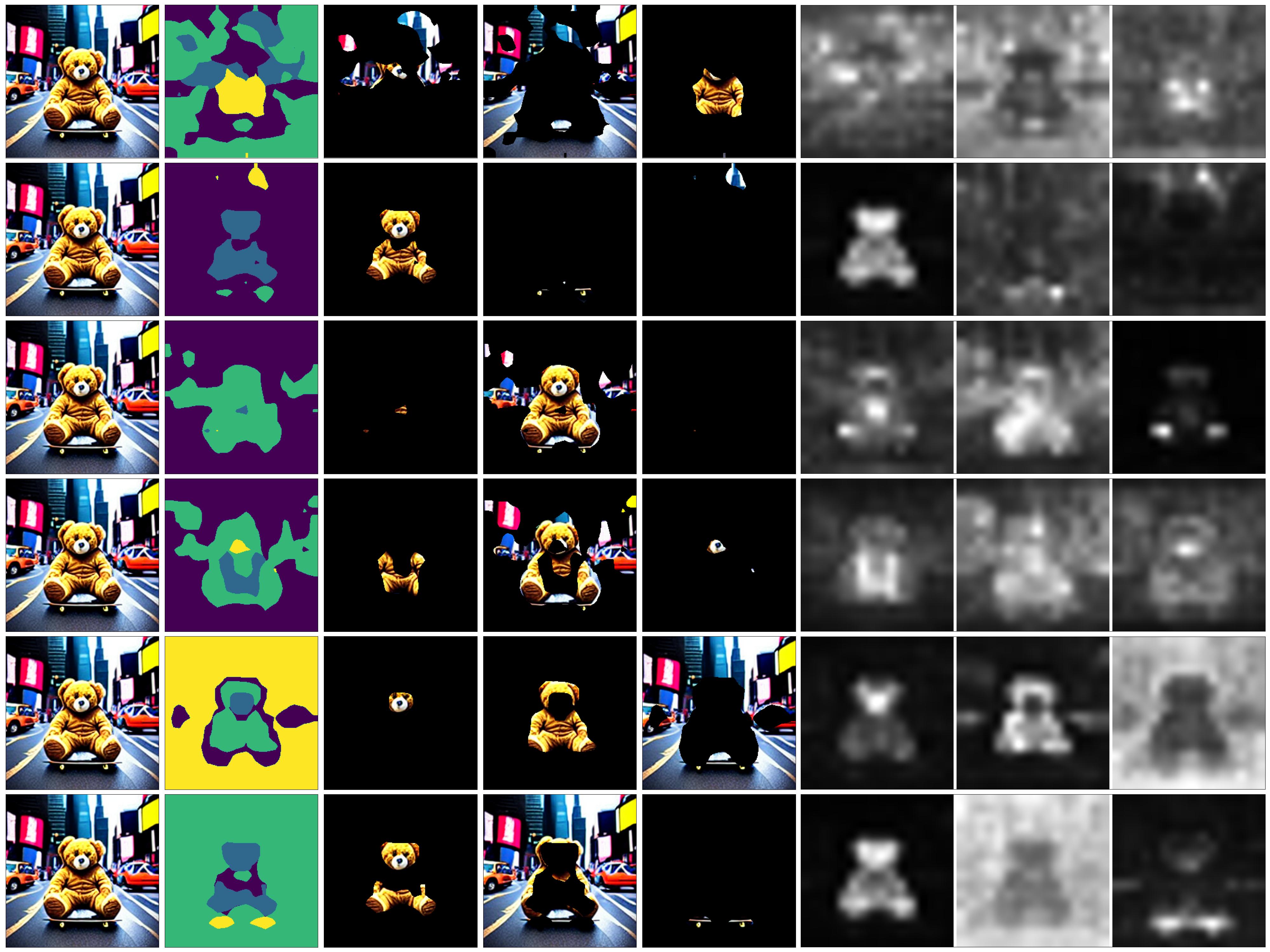} &
\includegraphics[trim={0 0 0 0},clip,width=0.555\textwidth]{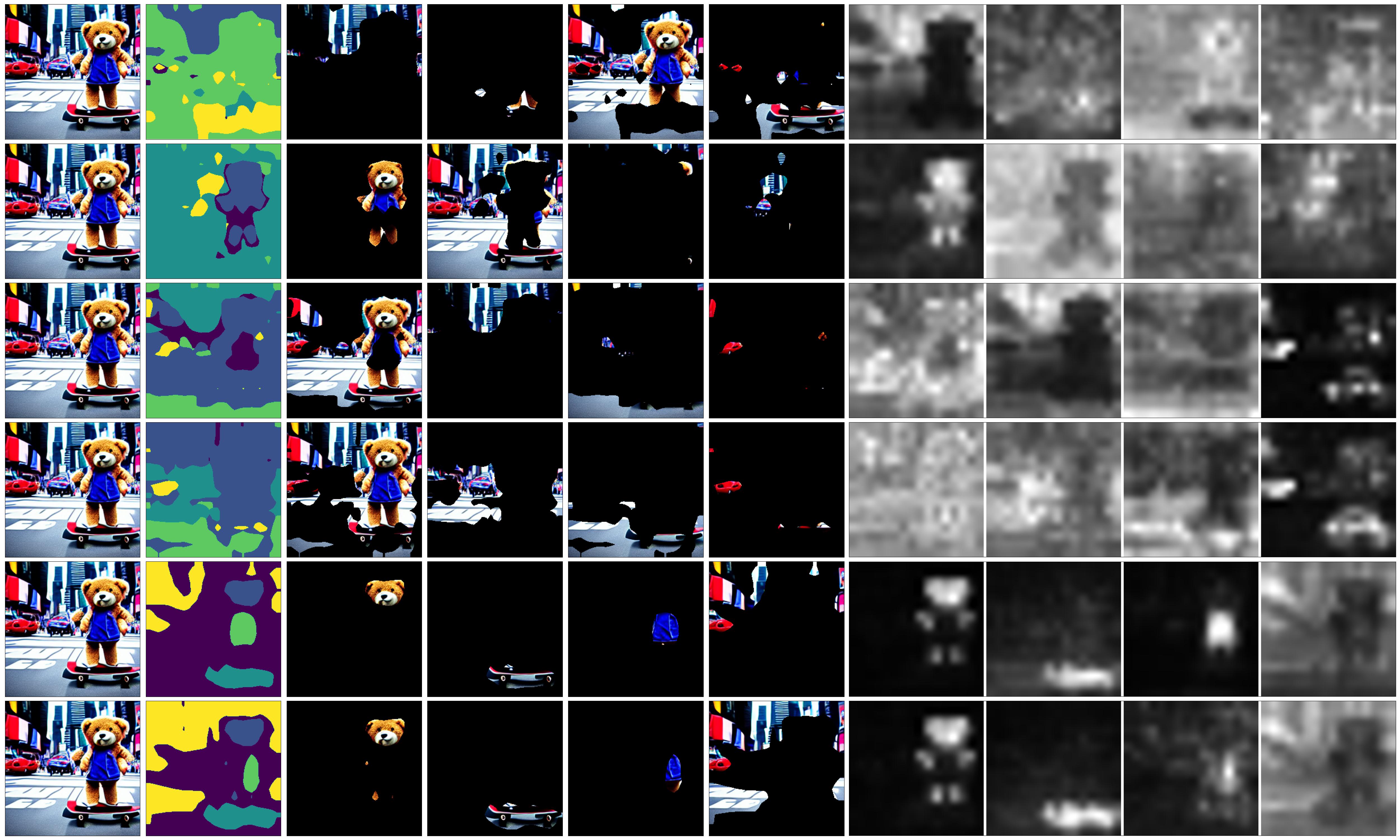} \\
\end{tabular}

\begin{tabular}{c c}
\hspace*{-.3cm}
\includegraphics[trim={0 0 0 0},clip,width=0.445\textwidth]{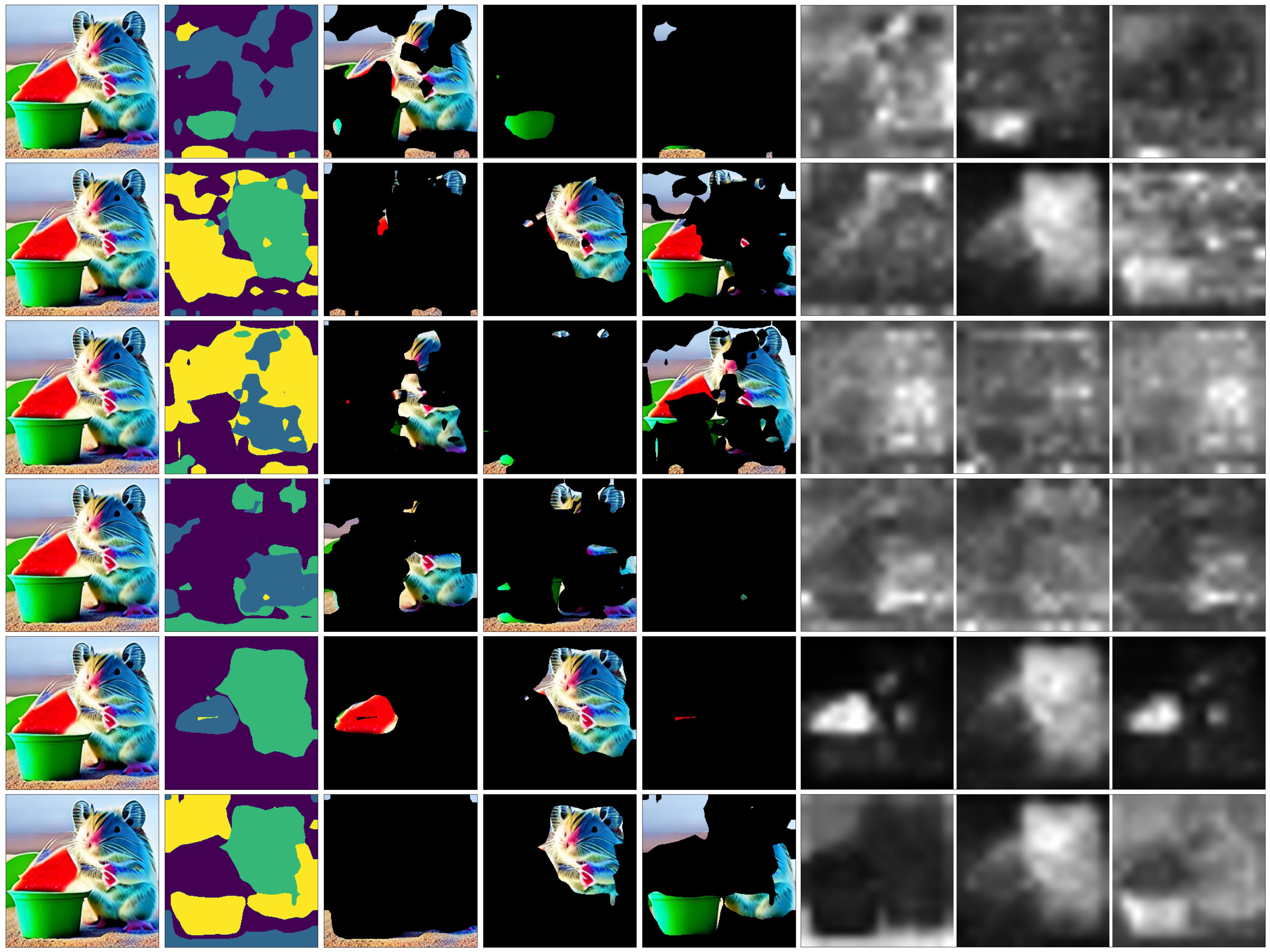} &
\includegraphics[trim={0 0 0 0},clip,width=0.555\textwidth]{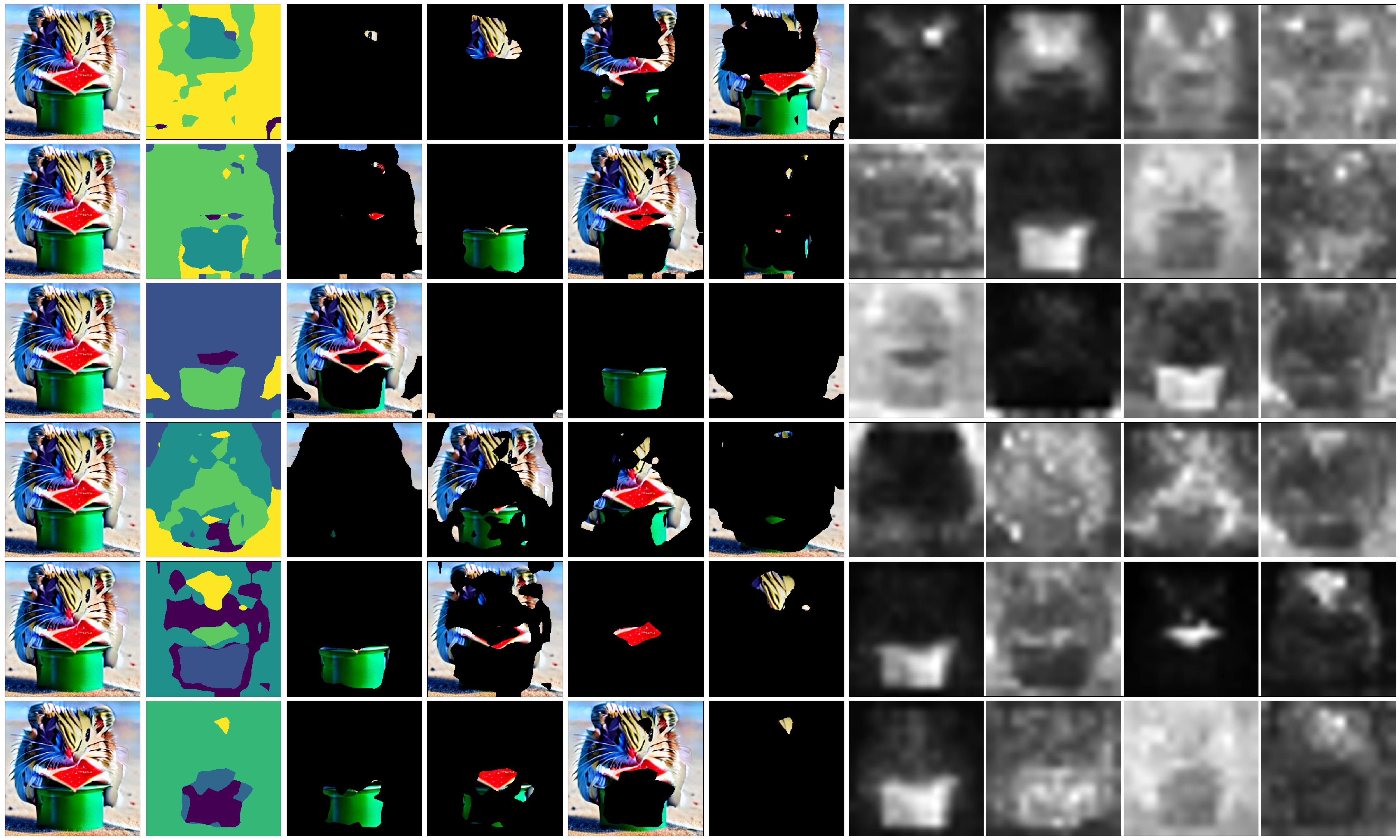} \\
\end{tabular}

\caption{Visualisation of concepts (3 to 5 concepts). 
We compare all baseline methods across each row: 
1) MCPL-all, 
2) MCPL-one, 
3) MCPL-all+\textit{PromptCL}+\textit{Bind adj.}, 
4) MCPL-all+\textit{AttnMask}+\textit{PromptCL}+\textit{Bind adj.}, 
5) MCPL-one+\textit{PromptCL}+\textit{Bind adj.}, 
6) MCPL-one+\textit{AttnMask}+\textit{PromptCL}+\textit{Bind adj.}.
Our findings indicate an increased challenge in learning multiple concepts as the number of objects in an image rises, particularly with a mask-free, language-driven approach like ours. Despite this, we \textbf{consistently observe improved accuracy in prompt-concept correlation when incorporating all proposed regularization terms (toward the bottom row)}.
}
\label{fig:concepts_345_seg_all}
\end{figure}

We present in this section the full results of assessing our proposed regularisation terms in \Secref{sec: regularisation}. The results presented in \Figref{fig:improve_MCPL} indicate that plain \textit{MCPL} may not accurately capture semantic correlations between prompts and objects. While adding incorporating the proposed regularisation terms enhances concept disentanglement. 
We assess the efficacy of these terms in disentangling learned concepts by visualising attention and segmentation masks, as shown in \Figref{fig:improve_MCPL}. \Figref{fig:visualise_learned_masked} and \Figref{fig:concepts_345_seg_all} present the same visualisation of the scaled quantitative experiments involving 2 to 5 concepts.

\newpage

\newpage
\subsection{Ablation study comparing MCPL-diverse versus MCPL-one in learning per-image different concept tasks}
\label{sec: ablation_diverse_vs_one}

\begin{figure}[H]
    \centering
    \begin{minipage}[t]{0.44\textwidth}
        \centering
        \includegraphics[width=\linewidth]{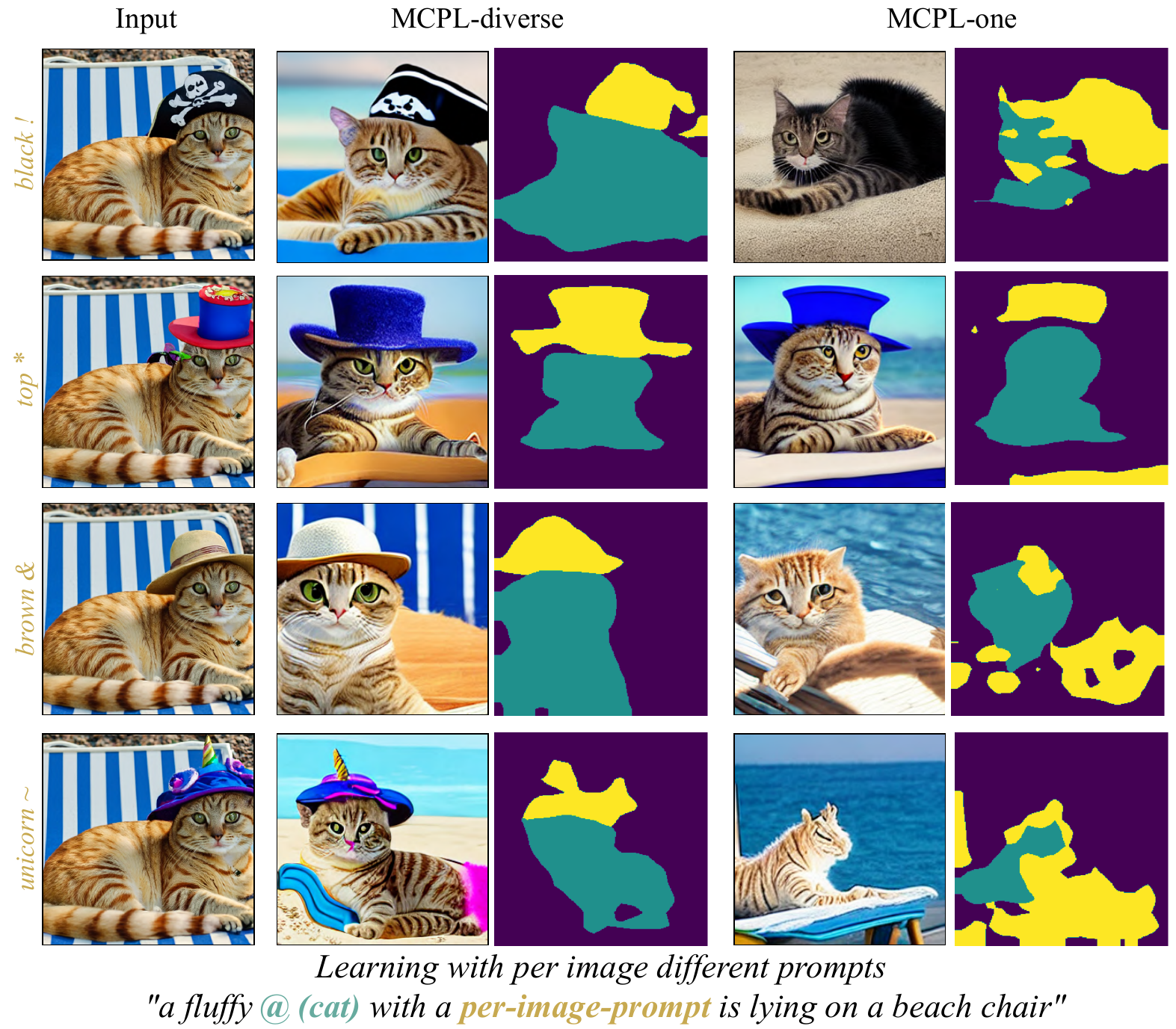}
    \end{minipage}\hfill 
    \begin{minipage}[t]{0.56\textwidth}
        \centering
        \includegraphics[width=\linewidth]{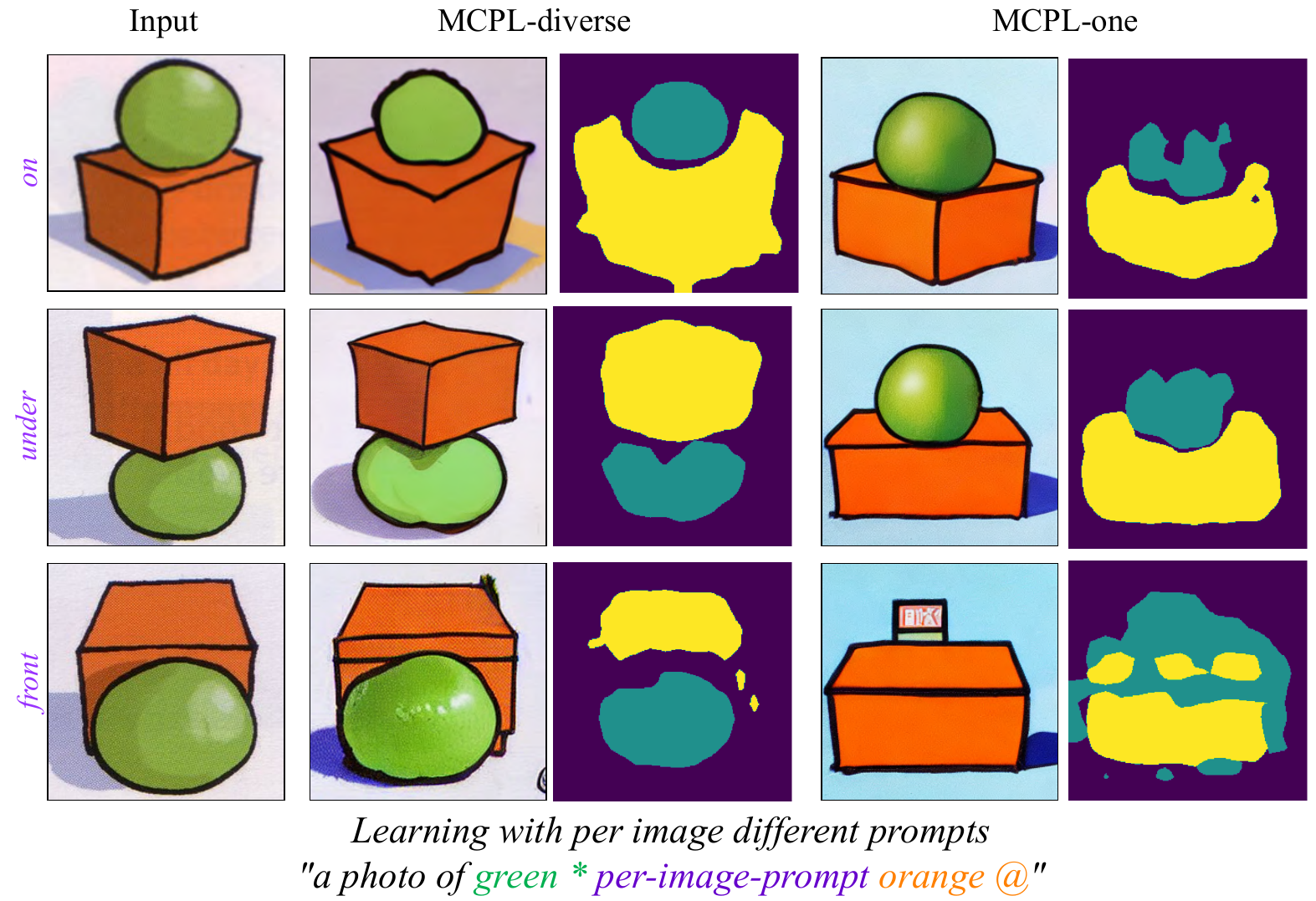}
    \end{minipage}
    \caption{Visual comparison of MCPL-diverse versus MCPL-one in learning per-image different concept tasks: Left - cat with different hat example. Right - ball and box relationships example. \textbf{As MCPL-diverse are specially designed for such tasks, it outperforms MCPL-one, which fails to capture per image different relationships.}}
    \label{fig:ablation_diverse_vs_one}
\end{figure}

The \textit{MCPL-diverse} training strategy has demonstrated potential in learning tasks with varying concepts per image. Therefore, we performed further experiments as detailed in \Figref{fig:ablation_diverse_vs_one} confirming its efficacy.

\subsection{Ablation study on the effect of Adjective words.}
\label{sec: ablation_adjective}

\begin{figure}[H]
    \centering
    \includegraphics[width=.8\linewidth]{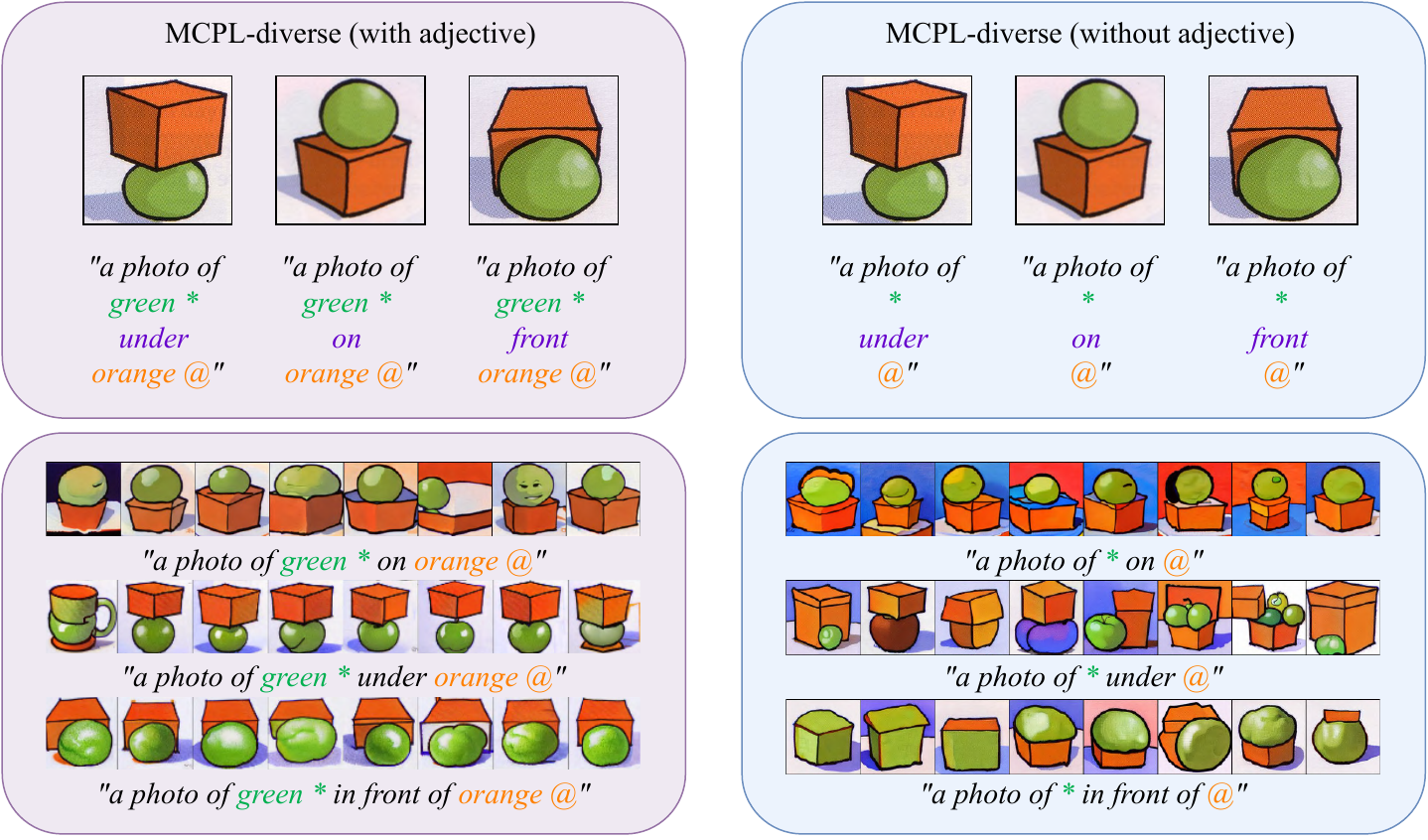}
    \caption{Visual comparison of MCPL-diverse with adjective word versus without adjective word. \textbf{Adjective words are crucial in linking each prompt to the correct region; without them, the model may struggle for regional guidance and we observe reduced performance.}}
    \label{fig:ablation_adj}
\end{figure}

Our language-driven approach benefits from the proposed adjective binding mechanism. To better understand its role, we performed experiments removing adjective words, as shown in \Figref{fig:ablation_adj}, which confirmed its significance.

\subsection{Visual comparison with BAS in composing complex scenes.}
\label{sec: bas_compare}

\begin{figure}[H]
    \centering
    \includegraphics[width=.9\linewidth]{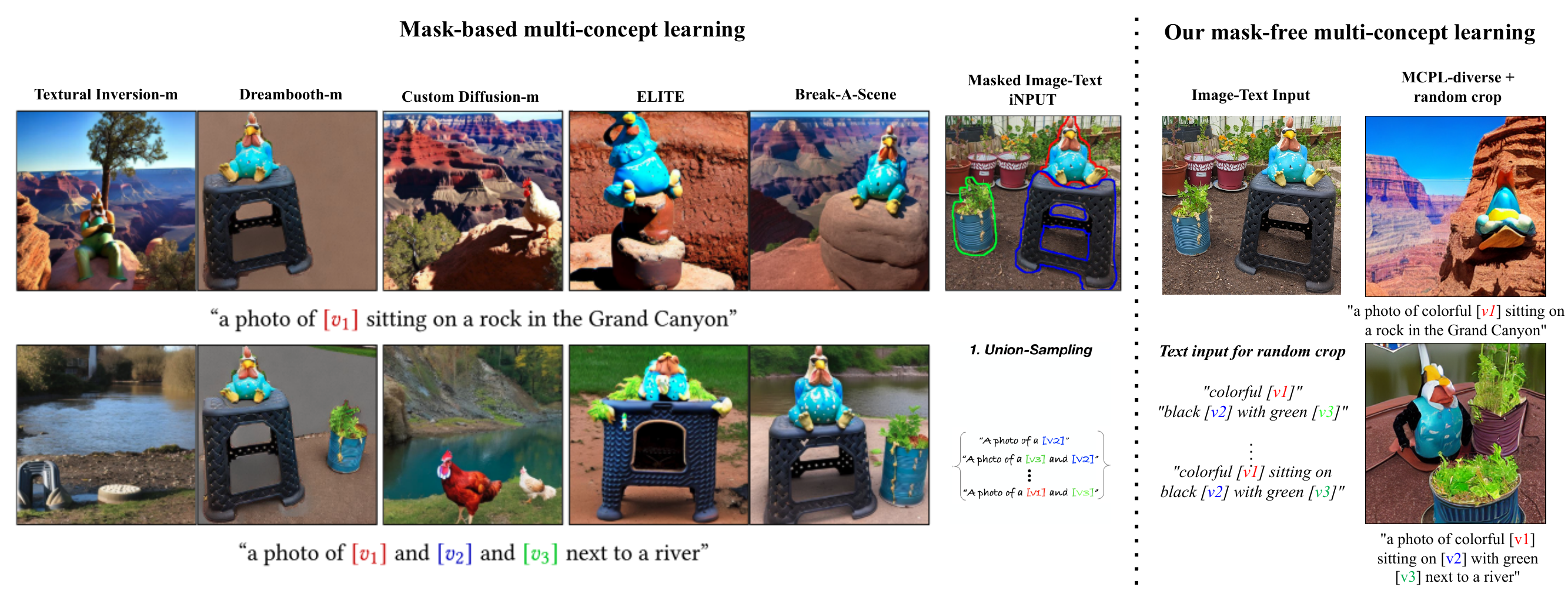}
    \caption{
    A qualitative comparison between 
    our method (\textit{MCPL-diverse}) and mask-based approaches: BAS, Textual Inversion \citep{gal2022textual} (masked), DreamBooth \citep{ruiz2022dreambooth} (masked), Custom Diffusion \citep{kumari2023multi} (masked) and ELITE \citep{wei2023elite}. 
    \textit{We stress our focus is not on optimising Diffusion Model (DM) parameters for enhanced composing performance}, unlike other compared methods. \textbf{Our approach delivers commendable results by learning from textual tokens (less than 0.1 MB) instead of updating the Diffusion Model of 4.9 GB as done in BAS, and it does so without depending on visual annotations such as masks.} \textit{For tasks prioritising composition, our method can serve as an initial mask proposal to identify pertinent tokens, subsequently integrating finetuning-based techniques for refinement.}
    Images modified from BAS \citep{avrahami2023break}.}
    \label{fig:ours_vs_bas_chicken}
\end{figure}

\begin{figure}[H]
    \centering
    \includegraphics[width=.9\linewidth]{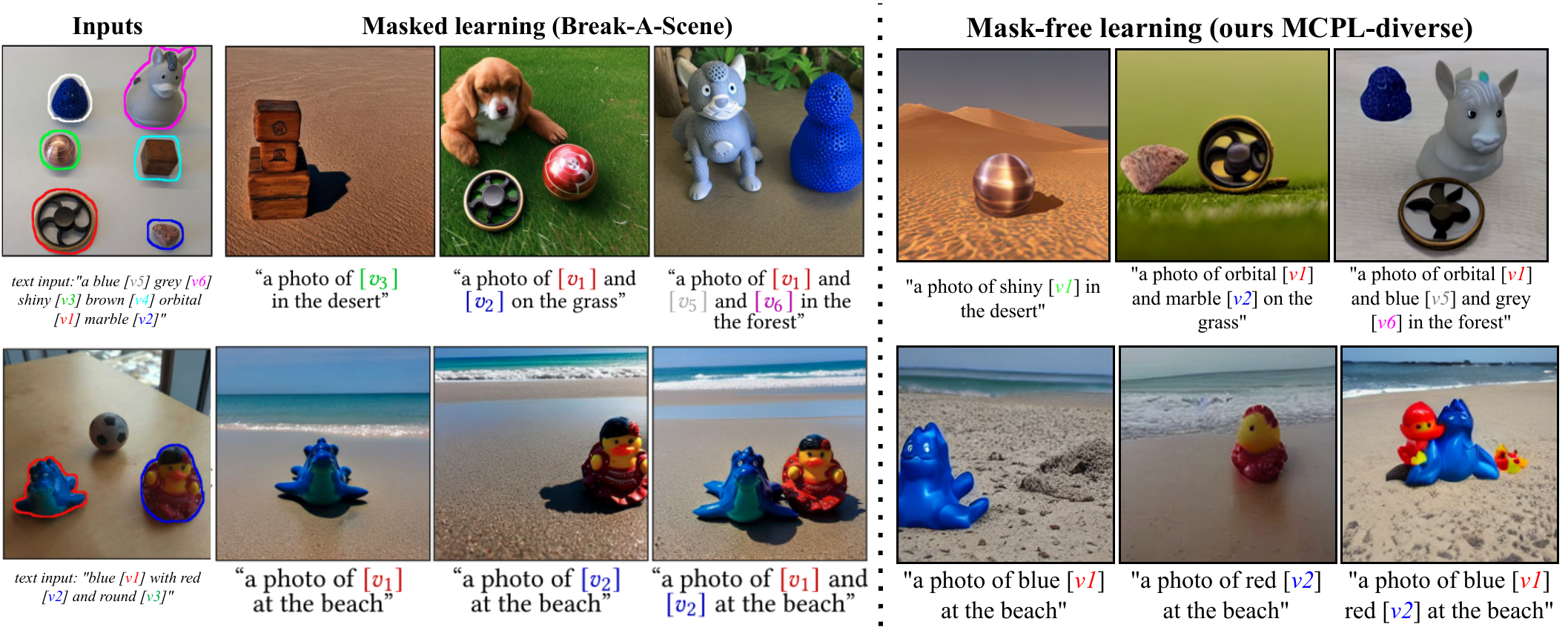}
    \caption{
    A qualitative comparison between BAS and our method (MCPL-one and MCPL-diverse). 
    \textbf{Top}, learning six concepts from a single image and then composing a subset, is particularly challenging, which BAS has acknowledged. \textbf{Similar to BAS, our method also faces challenges with a high number of concepts, but it shows promising and competitive results.} 
    \textbf{Bottom} learns three concepts from a single image. In this example, BAS performed better. Our MCPL-diverse, which neither uses mask inputs nor updates model parameters, showed decent results and was closer to BAS. 
    Images modified from BAS \citep{avrahami2023break}.
    }
    \label{fig:bas_compare}
\end{figure}

In tasks learning more than two concepts from a single image, we compare MCPL with Break-A-Scene (BAS). \textit{Unlike BAS, MCPL neither uses segmentation masks as input nor updates model parameters}. To level the playing field, we adopted BAS's 'union sampling' training strategy, which randomly selects subsets of multi-concepts in each iteration. We manually prepared a set of cropped images of each individual concept and randomly selected subsets to combine. This approach, termed \textit{'random crop,'} serves as our equivalent training strategy, see \Figref{fig:ours_vs_bas_chicken} for an illustration. Given that each cropped image has a different number of concepts, we utilised our \textit{MCPL-diverse}, designed to learn varying concepts per image. In \Figref{fig:ours_vs_bas_chicken} and \Figref{fig:bas_compare} we showcase examples of such tasks against a set of competitive baselines.

\twocolumn

\subsection{MCPL Algorithm and Variations}
\label{sec: algorithm}

We provide the formal definitions of all MCPL training strategies: MCPL (generic form) in Algorithm \ref{alg:alg_mcpl}, MCPL-all in Algorithm \ref{alg:alg_mcpl_all}, MCPL-one in Algorithm \ref{alg:alg_mcpl_one} and MCPL-diverse in Algorithm \ref{alg:alg_mcpl_diverse}.

\begin{algorithm}[ht]
\footnotesize
\caption{\textcolor{brown}{MCPL-all}}
\label{alg:alg_mcpl_all}
\begin{algorithmic}[1]
\STATE \textbf{Input:} example image(s) $x$, pre-trained $\{c_\theta, \eps_\theta\}$.
\STATE \textbf{Output:} a list of embeddings $\mathcal{V}=[v^*,\ldots,v^\&]$ corresponds to multiple new prompts $\mathcal{P}=[p^*,\ldots,{p}^\&]$,
\STATE \textcolor{brown}{which includes $[v_n^*,\ldots,v_n^\&]$ corresponds to noun words of target concepts $[n^*,\ldots,{n}^\&]$, $[v_a^*,\ldots,v_a^\&]$ corresponds to associated adjective words $[a^*,\ldots,{a}^\&]$ and $[v_t^1,\ldots,v_t^\&]$ corresponds to the rest texts in the string $\mathcal{T}=[t^*,\ldots,{t}^\&]$ (exclude random neutral texts).}
\STATE initialise $[v^*,\ldots,v^\&] = [c_\theta(p^*),\ldots,c_\theta(p^\&)]$
\STATE \textcolor{blue}{\texttt{\# optimising $\{v^*,\ldots,v^\&\}$ with $L_{DM}$}}
\FOR{$step = 1$ to $S$}
    \scriptsize
    \STATE \textbf{Encode} example image(s) $z = \mathcal{E}(x)$ and randomly sample neutral texts $y$ to make string $[y, p^*,\ldots,p^\&]$
    \STATE \makebox[0pt][l]{\textbf{Compute} $\mathcal{V_y}=[v^y,v^*,\ldots,v^\&] = [c_\theta(p^y),c_\theta(p^*),\ldots,c_\theta(p^\&)]$}
    \FOR{$t = T$ down to $1$}
        \STATE $\mathcal{V} := \argmin_{\mathcal{V}} E_{z,\mathcal{V},\eps,t} \Vert \eps-\eps_\theta(z_t,\mathcal{V_y}) \Vert^{2}$
    \ENDFOR
\ENDFOR
\STATE \textbf{Return} $(\mathcal{P}, \mathcal{V})$
\end{algorithmic}
\end{algorithm}

\begin{algorithm}[ht]
\footnotesize
\caption{\textcolor{orange}{MCPL-one}}
\label{alg:alg_mcpl_one}
\begin{algorithmic}[1]
\STATE \textbf{Input:} example image(s) $x$, pre-trained $\{c_\theta, \eps_\theta\}$.
\STATE \textbf{Output:} a list of embeddings including $\mathcal{V}=[v^*,\ldots,v^\&]$ corresponds to multiple new prompts $\mathcal{P}=[p^*,\ldots,{p}^\&]$,
\STATE \textcolor{orange}{which includes $[v_n^*,\ldots,v_n^\&]$ corresponds to noun words of target concepts $[n^*,\ldots,{n}^\&]$.}
\STATE initialise $[v^*,\ldots,v^\&] = [c_\theta(p^*),\ldots,c_\theta(p^\&)]$
\STATE \textcolor{blue}{\texttt{\# optimising $\{v^*,\ldots,v^\&\}$ with $L_{DM}$}}
\FOR{$step = 1$ to $S$}
    \scriptsize
    \STATE \textbf{Encode} example image(s) $z = \mathcal{E}(x)$ and randomly sample neutral texts $y$ to make string $[y, p^*,\ldots,p^\&]$
    \STATE \makebox[0pt][l]{\textbf{Compute} $\mathcal{V_y}=[v^y,v^*,\ldots,v^\&] = [c_\theta(p^y),c_\theta(p^*),\ldots,c_\theta(p^\&)]$}
    \FOR{$t = T$ down to $1$}
        \STATE $\mathcal{V} := \argmin_{\mathcal{V}} E_{z,\mathcal{V},\eps,t} \Vert \eps-\eps_\theta(z_t,\mathcal{V_y}) \Vert^{2}$
    \ENDFOR
\ENDFOR
\STATE \textbf{Return} $(\mathcal{P}, \mathcal{V})$
\end{algorithmic}
\end{algorithm}

\begin{algorithm}[ht]
\footnotesize
\caption{\textcolor{violet}{MCPL-diverse (generic form)}}
\label{alg:alg_mcpl_diverse}
\begin{algorithmic}[1]
\STATE \textbf{Input:} \textcolor{violet}{a set of $D$ example images $\mathcal{X}_d, d \in D$, each described by prompts $\mathcal{P}_d=[p_d^*,\ldots,{p}_d^\&]$}, pre-trained $\{c_\theta, \eps_\theta\}$.
\STATE \textbf{Output:} a list of embeddings $\mathcal{V}=[v^*,\ldots,v^\&]$ corresponds to multiple new prompts $\mathcal{P}=[p^*,\ldots,{p}^\&]$,
\STATE \textcolor{violet}{which includes $[v_d^*,\ldots,v_d^\&]$ corresponds to multiple new prompts $[p_d^*,\ldots,{p}_d^\&]$ of each $\mathcal{X}_d, d \in D$.}
\STATE initialise $[v^*,\ldots,v^\&] = [c_\theta(p^*),\ldots,c_\theta(p^\&)]$
\STATE \textcolor{blue}{\texttt{\# optimising $\{v^*,\ldots,v^\&\}$ with $L_{DM}$}}
\FOR{$step = 1$ to $S$}
    \scriptsize
    \STATE \textbf{Encode} example image(s) $z = \mathcal{E}(x)$ and randomly sample neutral texts $y$ to make string $[y, p^*,\ldots,p^\&]$
    \STATE \makebox[0pt][l]{\textbf{Compute} $\mathcal{V_y}=[v^y,v^*,\ldots,v^\&] = [c_\theta(p^y),c_\theta(p^*),\ldots,c_\theta(p^\&)]$}
    \FOR{$t = T$ down to $1$}
        \STATE $\mathcal{V} := \argmin_{\mathcal{V}} E_{z,\mathcal{V},\eps,t} \Vert \eps-\eps_\theta(z_t,\mathcal{V_y}) \Vert^{2}$
    \ENDFOR
\ENDFOR
\STATE \textbf{Return} $(\mathcal{P}, \mathcal{V})$
\end{algorithmic}
\end{algorithm}

\subsection{Implementation details.} 
\label{sec: implementation_details}
We use the same prompts collected during the data preparation, substituting nouns as learnable prompts, which are merged with CLIP prompts from \Secref{sec: preliminaries}. This process creates phrases such as \textit{``A photo of brown * on a rolling @ at times square"}.
Unless otherwise noted, we retain the original hyper-parameter choices of LDM \citep{rombach2022high}. Our experiments were conducted using a single V100 GPU with a batch size of 4. The base learning rate was set to $0.005$. Following LDM, we further scale the base learning rate by the number of GPUs and the batch size, for an effective rate of $0.02$. On calculating $L_{PromptCL}$, we apply the temperature and scaling term $(\tau, \gamma)$ of $(0.2, 0.0005)$ when \textit{AttnMask} is not applied, and  $(0.3, 0.00075)$ when \textit{AttnMask} is applied. All results were produced using $6100$ optimisation steps. We find that these parameters work well for most cases. 
The experiments were executed on a single V100 GPU, with each run taking approximately one hour, resulting in a total computational cost of around 3500 GPU hours (or 150 days on a single GPU). 
We employed various metrics to evaluate the method.

\subsection{Dataset preparation.}
\label{sec:dataset}
For the in-distribution natural images dataset, we first generate variations of two-concept images using local text-driven editing, as proposed by \cite{patashnik2023localizing}. This minimizes the influence of irrelevant elements like background. This approach also produces per-text local masks based on attention maps, assisting us in getting our best approximation for the ``ground truth" of disentangled embeddings. We generate five sets of natural images containing 10 object-level concepts. We generate each image using simple prompts, comprising one adjective and one noun for every relevant concept.
For three to five concept images, we use break-a-scene \citep{avrahami2023break} to generate the more complex composed images. We generate nine sets containing 9 more object-level concepts. We then use separate pre-trained segmentation models—MaskFormer \citep{cheng2021per} to create masked objects, refer to Appendix \ref{sec: bas_exp_setup} for details of this process. 

\begin{figure*}[t]
\centering
\setlength{\abovecaptionskip}{0.5pt}
\setlength{\belowcaptionskip}{0.5pt}
\setlength{\tabcolsep}{0.5pt}
\renewcommand{\arraystretch}{0.5} 

\begin{tabular}{c c}
\includegraphics[trim={0 0 0 0},clip,width=0.49\textwidth]{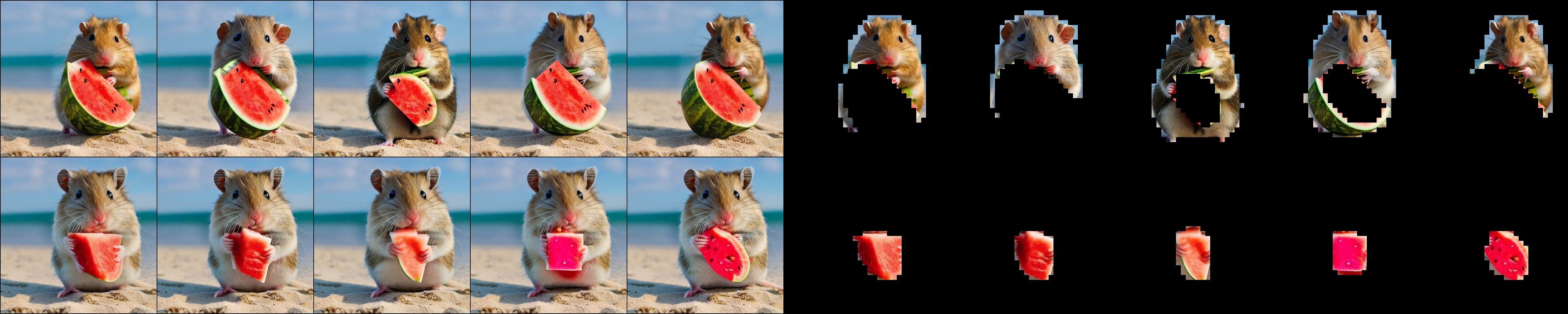} &
\includegraphics[trim={0 0 0 0},clip,width=0.49\textwidth]{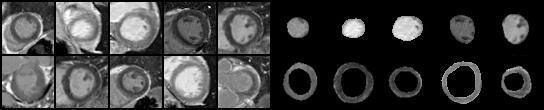} \\
{\scriptsize hamster (top) - watermelon (bottom)} & {\scriptsize cavity (top) - myocardium (bottom)} \\

\includegraphics[trim={0 0 0 0},clip,width=0.49\textwidth]{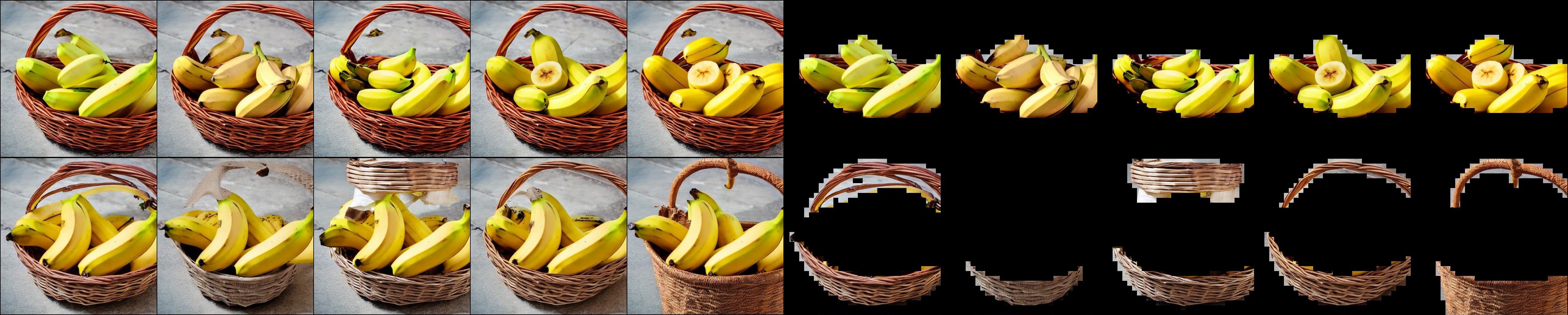} &
\includegraphics[trim={0 0 0 0},clip,width=0.49\textwidth]{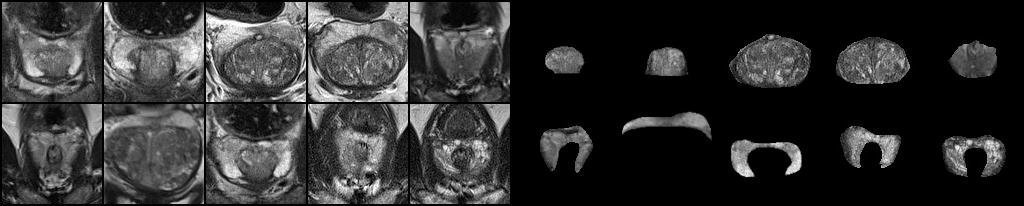} \\
{\scriptsize bananas (top) - basket (bottom)} & {\scriptsize transition (top) - peripheral (bottom)} \\

\includegraphics[trim={0 0 0 0},clip,width=0.49\textwidth]{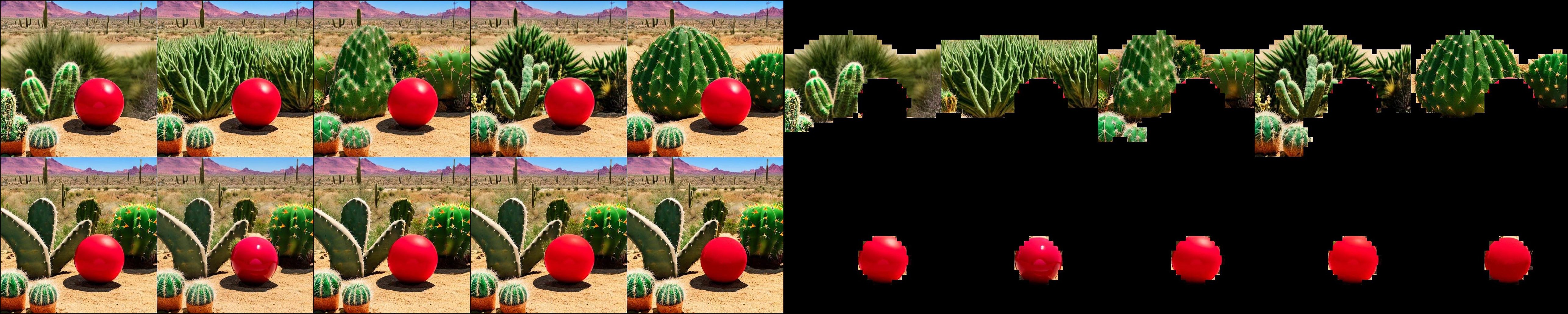} &
\includegraphics[trim={0 0 0 0},clip,width=0.49\textwidth]{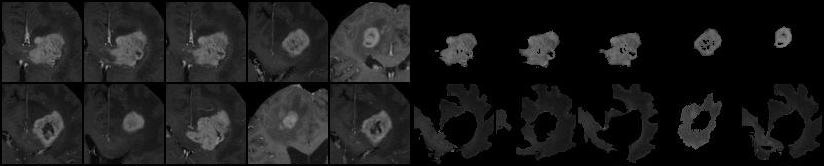} \\
{\scriptsize cactus (top) - ball (bottom)} & {\scriptsize tumour (top) - edema (bottom)} \\

\end{tabular}

\caption{\textbf{Evaluation dataset (two concepts).} We prepared five sets of in-distribution natural images and three sets of out-of-distribution biomedical images, each containing two concepts resulting in a total of 16 concepts.}
\label{fig:quantitative_dataset_highlight}
\end{figure*}

\begin{figure*}[t]
    \centering
    \includegraphics[width=1\linewidth]{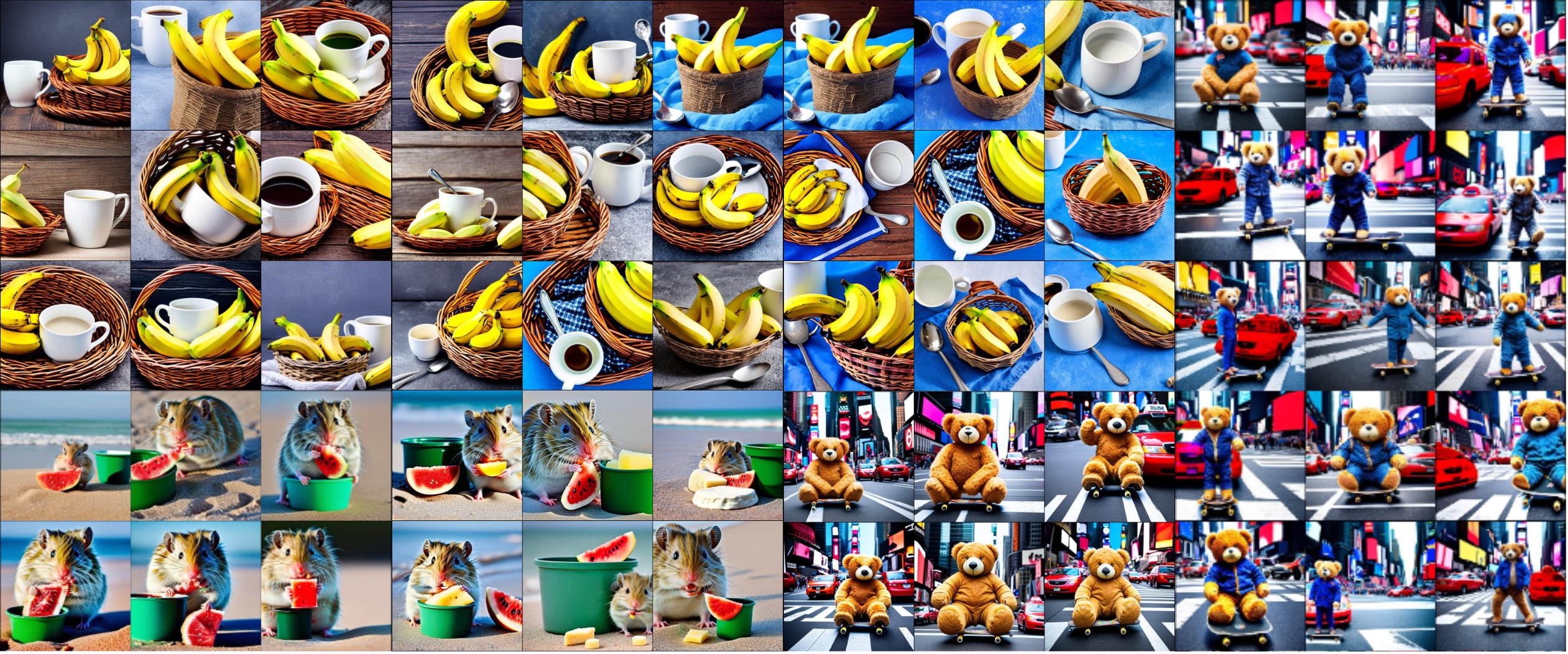}
    \caption{\textbf{Evaluation dataset (three to five concepts).} We generate nine sets containing 9 more object-level concepts.}
    \label{fig:concepts_345_dataset}
\end{figure*}

For the out-of-distribution bio-medical image dataset, we assemble three sets of radiological images featuring 6 organ/lesion concepts. These images are sourced from three public MRI segmentation datasets: heart myocardial infarction \citep{lalande2020emidec}, prostate segmentation \citep{antonelli2022medical}, and Brain Tumor Segmentation (BraTS) \citep{menze2014multimodal}. Each dataset includes per-concept masks.
For biomedical images, we request a human or a machine, such as GPT-4, to similarly describe each image using one adjective and one noun for each pertinent concept.
For both natural and biomedical datasets, we collected 40 images for each concept. \Figref{fig:quantitative_dataset_highlight} and \Figref{fig:concepts_345_dataset} gives some examples of the prepared datasets.  

\begin{table}[H]
\centering
\begin{tabular}{p{0.45\textwidth}}
\hline
\multicolumn{1}{c}{Two-concepts (natural images)} \\
\hline
\begin{itemize}
\small
    \item ``a brown \{bear/ tokens1\} on a rolling \{skateboard/ tokens2\} at times square"
    \item ``a fluffy \{hamster/ tokens1\} eating red \{watermelon/ tokens2\} on the beach"
    \item ``a green \{cactus/ tokens1\} and a red \{ball/ tokens2\} in the desert"
    \item ``a brown \{basket/ tokens1\} with yellow \{bananas/ tokens2\}"
    \item ``a white \{chair\ tokens1\} with a black \{dog/ tokens2\} on it"
\end{itemize}
\\
\hline
\end{tabular}
\end{table}

\begin{table}[H]
\centering
\begin{tabular}{p{0.45\textwidth}}
\hline
\multicolumn{1}{c}{Two-concepts (medical images)} \\
\hline
\begin{itemize}
\small
    \item ``a \{scan/ tokens1\} with brighter \{cavity/ tokens2\} encircled by grey \{myocardium/ tokens3\}"
    \item ``a \{scan/ tokens1\} with round \{transition/ tokens2\} encircled by circle \{peripheral/ tokens3\}"
    \item ``a \{scan/ tokens1\} with round \{tumour/ tokens2\} encircled by circle \{edema/ tokens3\}"
\end{itemize}
\\
\hline
\end{tabular}
\end{table}

\begin{table}[H]
\centering
\begin{tabular}{p{0.45\textwidth}}
\hline
\multicolumn{1}{c}{Three-concepts (natural images)} \\
\hline
\begin{itemize}
    \item ``a brown \{bear/ tokens1\} on a rolling \{skateboard/ tokens2\} beside a red \{taxi/ tokens3\} at times square"
    \item ``a green \{pot/ tokens1\} and a \{hamster/ tokens1\} eating red \{watermelon/ tokens2\} on the beach"
    \item ``a \{basket/ tokens1\} with yellow \{bananas/ tokens2\} beside a white \{mug/ tokens3\}"
\end{itemize}
\\
\hline
\end{tabular}
\end{table}

\begin{table}[H]
\centering
\begin{tabular}{p{0.45\textwidth}}
\hline
\multicolumn{1}{c}{Four-concepts (natural images)} \\
\hline
\begin{itemize}
    \item ``a brown \{bear/ tokens1\} on a rolling \{skateboard/ tokens2\} wearing a blue \{jacket/ tokens3\} beside a red \{taxi/ tokens4\} at times square"
    \item ``a green \{pot/ tokens1\} and a \{hamster/ tokens1\} eating red \{watermelon/ tokens2\} beside a yellow \{cheese/ tokens4\} on the beach"
    \item ``a \{basket/ tokens1\} with yellow \{bananas/ tokens2\} beside a white \{mug/ tokens3\} and silver \{spoon/ tokens4\}"
\end{itemize}
\\
\hline
\end{tabular}
\end{table}

\begin{table}[H]
\centering
\begin{tabular}{p{0.45\textwidth}}
\hline
\multicolumn{1}{c}{Five-concepts (natural images)} \\
\hline
\begin{itemize}
    \item ``a brown \{bear/ tokens1\} on a rolling \{skateboard/ tokens2\} wearing a blue \{jacket/ tokens3\} beside a red \{taxi/ tokens4\} over grey \{road/ tokens5\} at times square"
    \item ``a green \{pot/ tokens1\} and a \{hamster/ tokens1\} eating red \{watermelon/ tokens2\} beside a yellow \{cheese/ tokens4\} on the sandy \{beach/ tokens5\}"
    \item ``a \{basket/ tokens1\} with yellow \{bananas/ tokens2\} beside a white \{mug/ tokens3\} and silver \{spoon/ tokens4\} over blue \{tablecloth/ tokens5\}"
\end{itemize}
\\
\hline
\end{tabular}
\end{table}

\subsection{Break-A-Scene experiments setup}
\label{sec: bas_exp_setup}
Break-A-Scene (BAS) \citep{avrahami2023break} learns multiple concepts from images paired with object-level masks. It augments input images with masks to highlight target concepts and updates both textual embeddings and model weights accordingly. BAS introduces 'union sampling', a training strategy that randomly selects subsets of multi-concepts in each iteration to enhance the combination of multiple concepts in generated images, see \Figref{fig:ours_vs_bas_chicken} for an illustration. During inference, BAS employs a pre-trained segmentation model to obtain masked objects, facilitating localised editing.

To fit BAS \citep{avrahami2023break} into our evaluation protocol, we first learned object-level concepts and then generated masked objects for evaluation, including the following steps:
\begin{enumerate}
    \item BAS Learning: For each concept pair, we randomly selected 20 images with ground truth segmentations from our dataset for BAS learning, resulting in 20 BAS embeddings per concept.
    \item BAS Generation: We then generated 20 images for each concept pair, producing a total of 100 BAS-generated natural images and 60 medical images.
    \item Segmentation: For masked object production with BAS, we used different pre-trained segmentation models. MaskFormer \citep{cheng2021per} was effective for natural images, but segmenting medical images posed challenges due to their out-of-distribution characteristics.
    \item Quantitative Evaluation: With the obtained masked objects (20 per concept), we applied the embedding similarity evaluation protocol from Section \ref{sec: quantitative} to assess the preservation of semantic and textual details per concept in four embedding spaces.
\end{enumerate}

\begin{figure}[h]
\vspace{-3mm}
  \centering
  \begin{minipage}{.5\linewidth}
    \centering
    \includegraphics[width=1\linewidth]{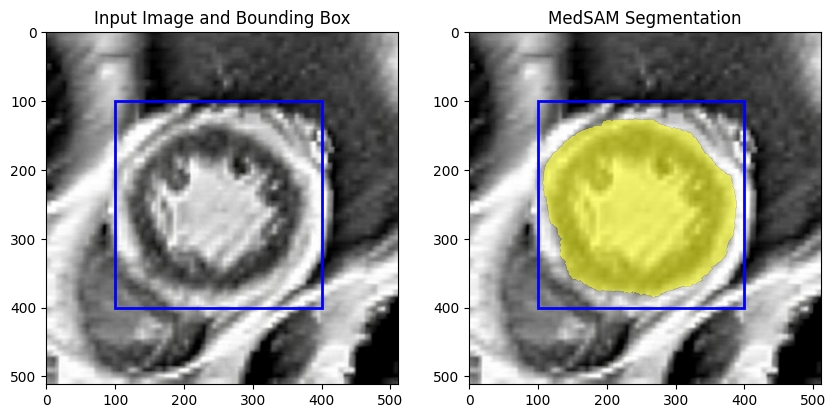}
  \end{minipage}%
  \begin{minipage}{.5\linewidth}
    \centering
    \includegraphics[width=1\linewidth]{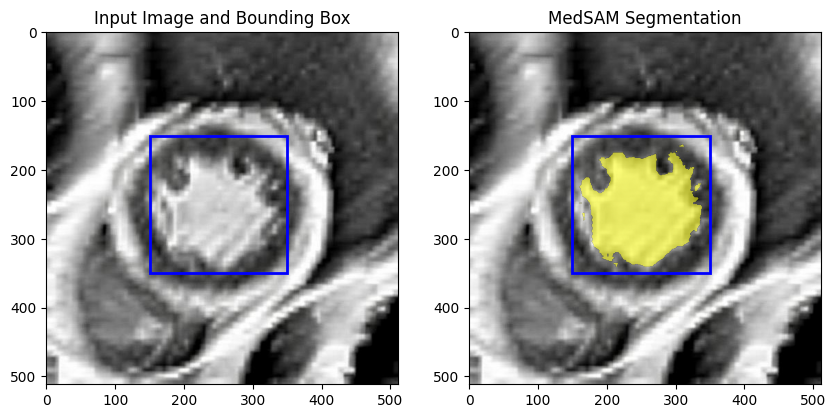}
  \end{minipage}
  \vspace{-3mm}
  \caption{
  This demonstration shows how MedSAM is used to segment medical images generated by BAS. On the left, MedSAM segmentation with a large bounding box prompt can identify the combined area of the cavity-myocardium classes, but it does not distinguish between the two. On the right, using a smaller bounding box prompt, MedSAM successfully segments the central cavity class. We calculate the difference to get the segmentation of the missing myocardium class (outer ring-like pattern).
  }
    \label{fig:medsam_example}
\end{figure}

For segmenting medical images, given the diversity of classes in our dataset, we utilised MedSAM \citep{ma2023segment}, a state-of-the-art foundation model adapted from SAM \citep{kirillov2023segment} for the medical domain. MedSAM requires a bounding box for input, making it a multi-step, human-in-the-loop process. We initially assessed segmentation quality from several (up to five) bounding box proposals, as exemplified in \Figref{fig:medsam_example}. MedSAM, despite having a bounding box, cannot fully automate segmentation for all classes. 

Thus, we employed an additional post-processing step to discern the segmentation of both classes by calculating the difference between the two segmentations.

\begin{figure}[h]
    \centering
    \includegraphics[width=.9\linewidth]{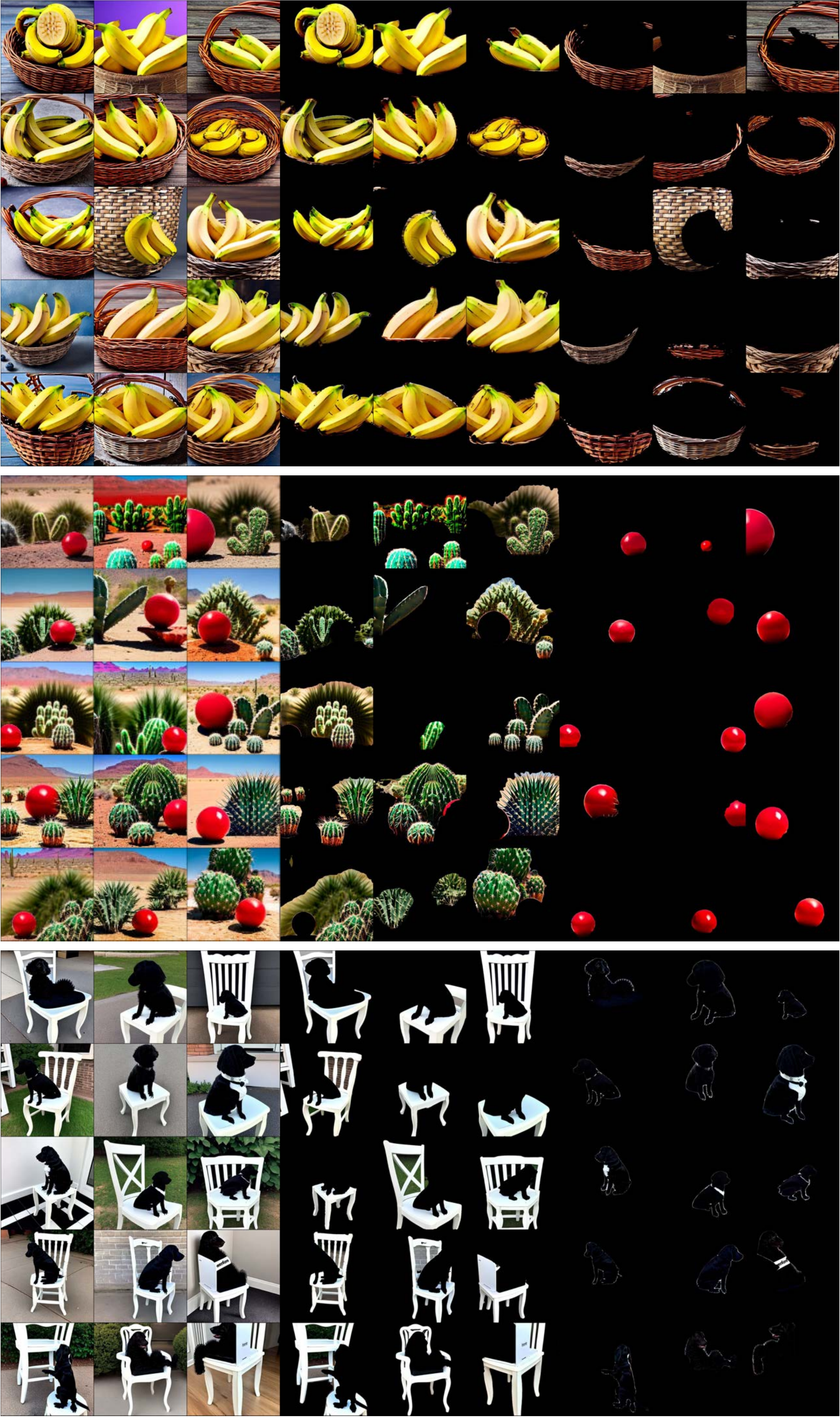}
    \caption{Visualisation of the Break-A-Scene results of generated and masked natural images.}
    \label{fig:break_a_scene_all_natural}
\end{figure}

\begin{figure}[h]
    \centering
    \includegraphics[width=\linewidth]{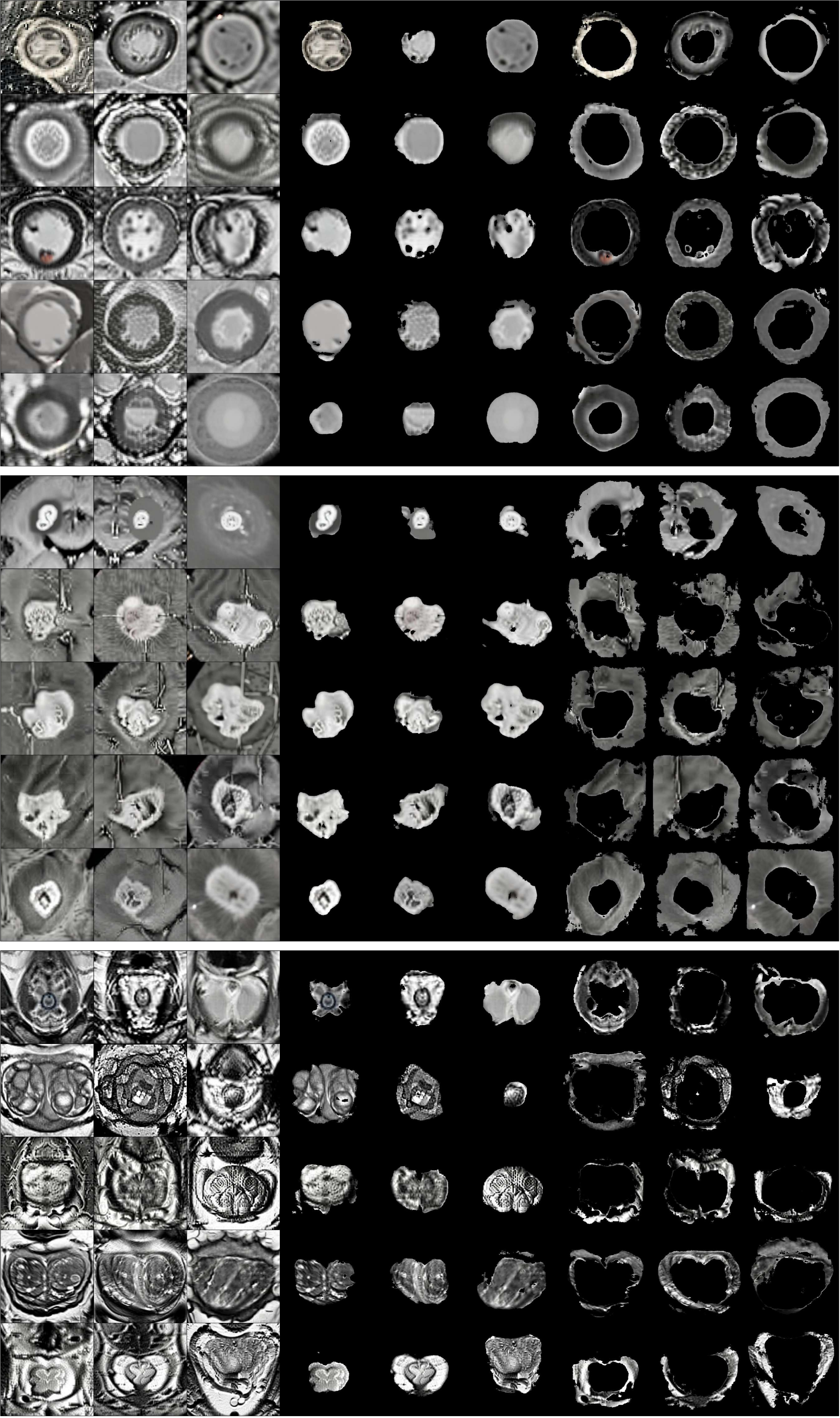}
    \caption{Visualisation of the Break-A-Scene results of generated and masked medical images.}
    \label{fig:break_a_scene_all_medical}
\end{figure}


\end{document}